\documentclass[10pt,journal,compsoc]{IEEEtran}

\ifCLASSOPTIONcompsoc
  \usepackage[nocompress]{cite}
\else
  \usepackage{cite}
\fi

\ifCLASSINFOpdf
\usepackage[pdftex]{graphicx}
\DeclareGraphicsExtensions{.pdf,.jpeg,.png}
\DeclareGraphicsExtensions{.eps}
\usepackage{epstopdf}
\else
\fi

\ifCLASSOPTIONcompsoc
\usepackage[caption=false,font=footnotesize,labelfont=sf,textfont=sf]{subfig}
\else
\usepackage[caption=false,font=footnotesize]{subfig}
\fi
\usepackage{booktabs}
\usepackage{amssymb}

\usepackage{times}
\usepackage{epsfig}
\usepackage{graphicx}
\usepackage{amsmath}
\usepackage{cite}
\usepackage{enumerate}
\usepackage{cases}
\usepackage{multirow}
\usepackage{verbatim}
\usepackage{amssymb}
\usepackage{CJK}
\usepackage{algorithm}
\usepackage{algorithmicx}
\usepackage{algpseudocode}

\usepackage{color}
\usepackage{bm}
\usepackage{booktabs}
\usepackage{colortbl}
\usepackage{float}
\usepackage{ragged2e}  
\usepackage{caption}
\usepackage{csquotes}
\usepackage[breaklinks=true,bookmarks=false]{hyperref}

\newcommand{\etal}{\textit{et al}.}
\newcommand{\ie}{\textit{i}.\textit{e}.}
\newcommand{\eg}{\textit{e}.\textit{g}.}

\begin{document}

\title{Learning to Enhance Low-Light Image \\
	 via Zero-Reference Deep Curve Estimation}

\author{Chongyi Li, Chunle Guo, and Chen Change Loy,~\IEEEmembership{Senior Member,~IEEE}\\

\thanks{C. Li and C. C. Loy are with S-Lab, Nanyang Technological University (NTU), Singapore  (e-mail: chongyi.li@ntu.edu.sg and ccloy@ntu.edu.sg).}
\thanks{C. Guo is with the College of Computer Science, Nankai University, Tianjin, China  (e-mail: guochunle@nankai.edu.cn).}
\thanks{C. Li and C. Guo contribute equally.}
\thanks{C. C. Loy is the corresponding author.}
}

\markboth{IEEE TRANSACTIONS ON PATTERN ANALYSIS AND MACHINE INTELLIGENCES}%
{Shell \MakeLowercase{\textit{et al.}}: Bare Demo of IEEEtran.cls for Computer Society Journals}

\IEEEtitleabstractindextext{%
\justify  
\begin{abstract}
\label{sec:Abstrat}
This paper presents a novel method, Zero-Reference Deep Curve Estimation (Zero-DCE), which formulates light enhancement as a task of image-specific curve estimation with a deep network.  Our method trains a lightweight deep network, DCE-Net, to estimate pixel-wise and high-order curves for dynamic range adjustment of a given image. The curve estimation is specially designed, considering pixel value range, monotonicity, and differentiability. Zero-DCE is appealing in its relaxed assumption on reference images, i.e., it does not require any paired or even unpaired data during training. This is achieved through a set of carefully formulated non-reference loss functions, which implicitly measure the enhancement quality and drive the learning of the network. Despite its simplicity, we show that it generalizes well to diverse lighting conditions. Our method is efficient as image enhancement can be achieved by an intuitive and simple nonlinear curve mapping.  
We further present an accelerated and light version of Zero-DCE, called Zero-DCE++, that takes advantage of a tiny network with just 10K parameters. Zero-DCE++ has a fast inference speed (1000/11 FPS on a single GPU/CPU for an image of size 1200$\times$900$\times$3) while keeping the enhancement performance of Zero-DCE.
Extensive experiments on various benchmarks demonstrate the advantages of our method over state-of-the-art methods qualitatively and quantitatively. Furthermore, the potential benefits of our method to face detection in the dark are discussed. 
The source code will be made publicly available at \url{https://li-chongyi.github.io/Proj_Zero-DCE++.html}.
\end{abstract}

}

\maketitle

\IEEEdisplaynontitleabstractindextext

\IEEEpeerreviewmaketitle


\IEEEraisesectionheading{\section{Introduction}
\label{sec:Introduction}}
\IEEEPARstart{M}{any} photos are often captured under suboptimal lighting conditions due to inevitable environmental and/or technical constraints. These include inadequate and unbalanced lighting conditions in the environment, incorrect placement of objects against extreme back light, and under-exposure during image capturing. Such low-light photos suffer from compromised aesthetic quality and unsatisfactory transmission of information. The former affects viewers' experience while the latter leads to wrong message being communicated, such as inaccurate object/face detection and recognition. In addition, although deep neural networks have shown impressive performance on image enhancement and restoration \cite{Pan18,Lai18,Gu19,LiTIP19,LiTMM19}, they inevitably lead to high memory footprint and long inference time due to massive parameter space. 
The low computational cost and fast inference speed of deep models are desired in practical applications, especially for resource-limited and real-time devices, such as mobile platforms.

In this study, we present a novel deep learning-based method, Zero-Reference Deep Curve Estimation (Zero-DCE), for low-light image enhancement. It can cope with diverse lighting conditions including nonuniform and poor lighting cases.
Instead of performing image-to-image mapping, we reformulate the task as an image-specific curve estimation problem.
In particular, the proposed method takes a low-light image as input and produces high-order curves as its output. These curves are then used for pixel-wise adjustment on the dynamic range of the input to obtain an enhanced image.
The curve estimation is carefully formulated so that it maintains the range of the enhanced image and preserves the contrast of neighboring pixels. Importantly, it is differentiable, and thus we can learn the adjustable parameters of the curves through a deep convolutional neural network.
The proposed network is lightweight and the designed curve can be iteratively applied to approximate higher-order curves for more robust and accurate dynamic range adjustment.

A unique advantage of our deep learning-based method is \textbf{zero-reference}, \ie, it does not require any paired or even unpaired data in the training process as in existing CNN-based~\cite{Wang2019,Chen2018,Xu2020CVPR} and GAN-based methods~\cite{Jiang2019,CycleGAN}.
This is made possible through a set of specially designed non-reference loss functions including spatial consistency loss, exposure control loss, color constancy loss, and illumination smoothness loss, all of which take into consideration multi-factor of light enhancement.
We show that even with zero-reference training, Zero-DCE can still perform competitively against other methods that require paired or unpaired data for training.
The proposed method is flexible. We provide options to balance the enhancement performance and the computational cost for the practical application of Zero-DCE and propose an accelerated and light version Zero-DCE++. This is achieved  by re-designing the network structure, reformulating the curve estimation, and controlling the  sizes of input image.

\begin{figure*}
	\begin{center}
		\begin{tabular}{c@{ }c@{ }c@{ }c@{ }c@{ }}
			\includegraphics[height=2.6cm,width=3.4cm]{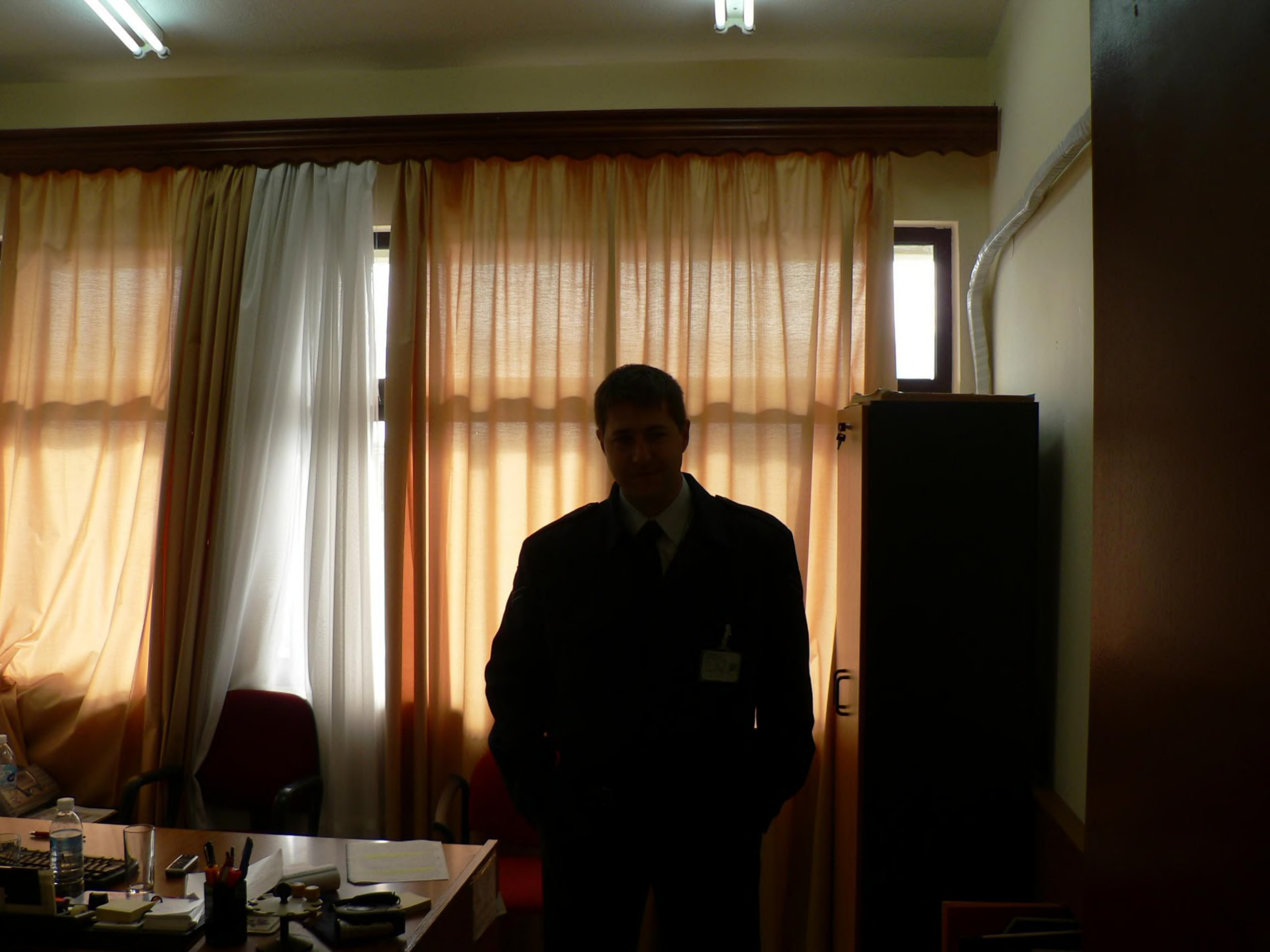}&
			\includegraphics[height=2.6cm,width=3.4cm]{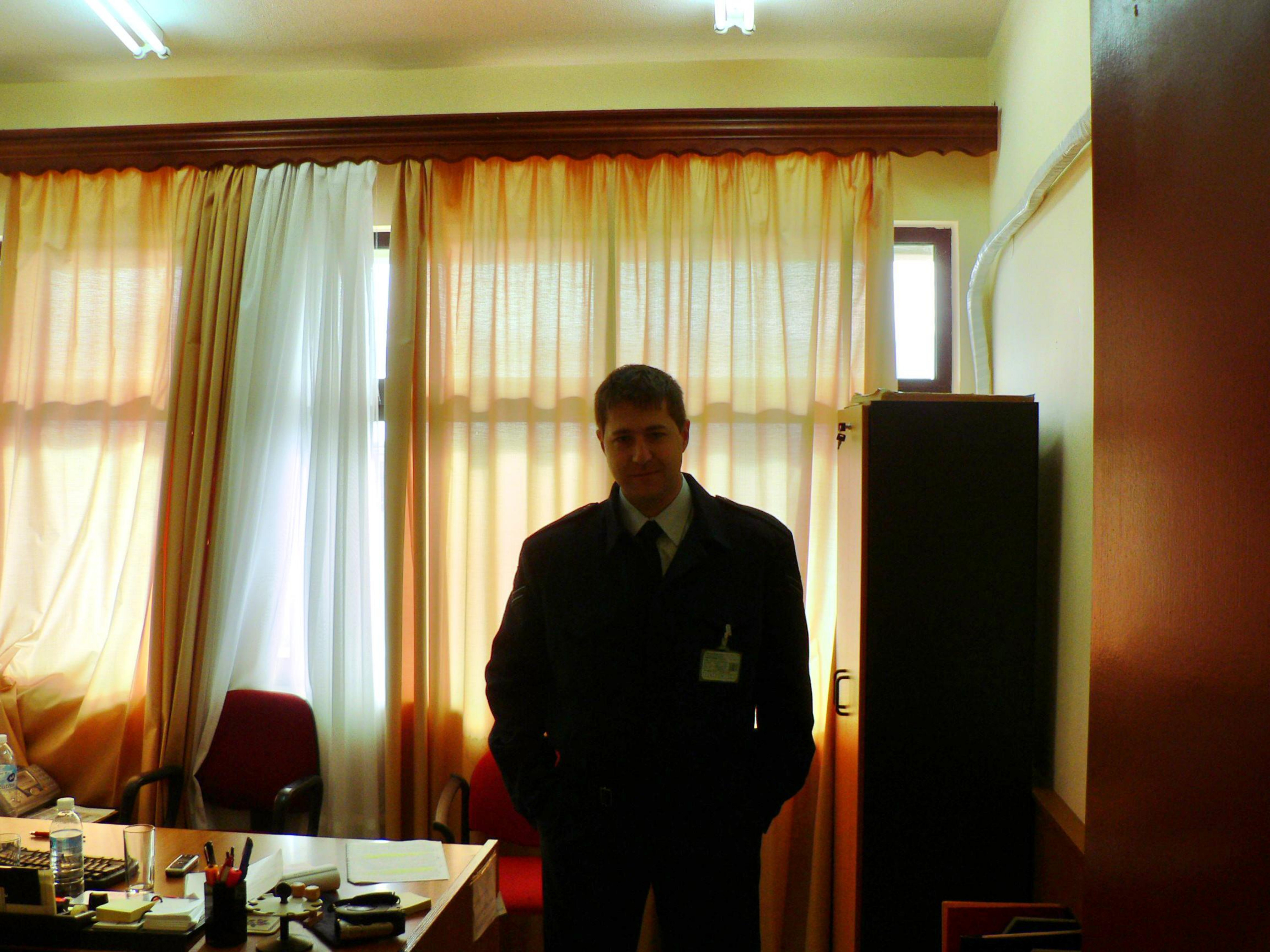}&
			\includegraphics[height=2.6cm,width=3.4cm]{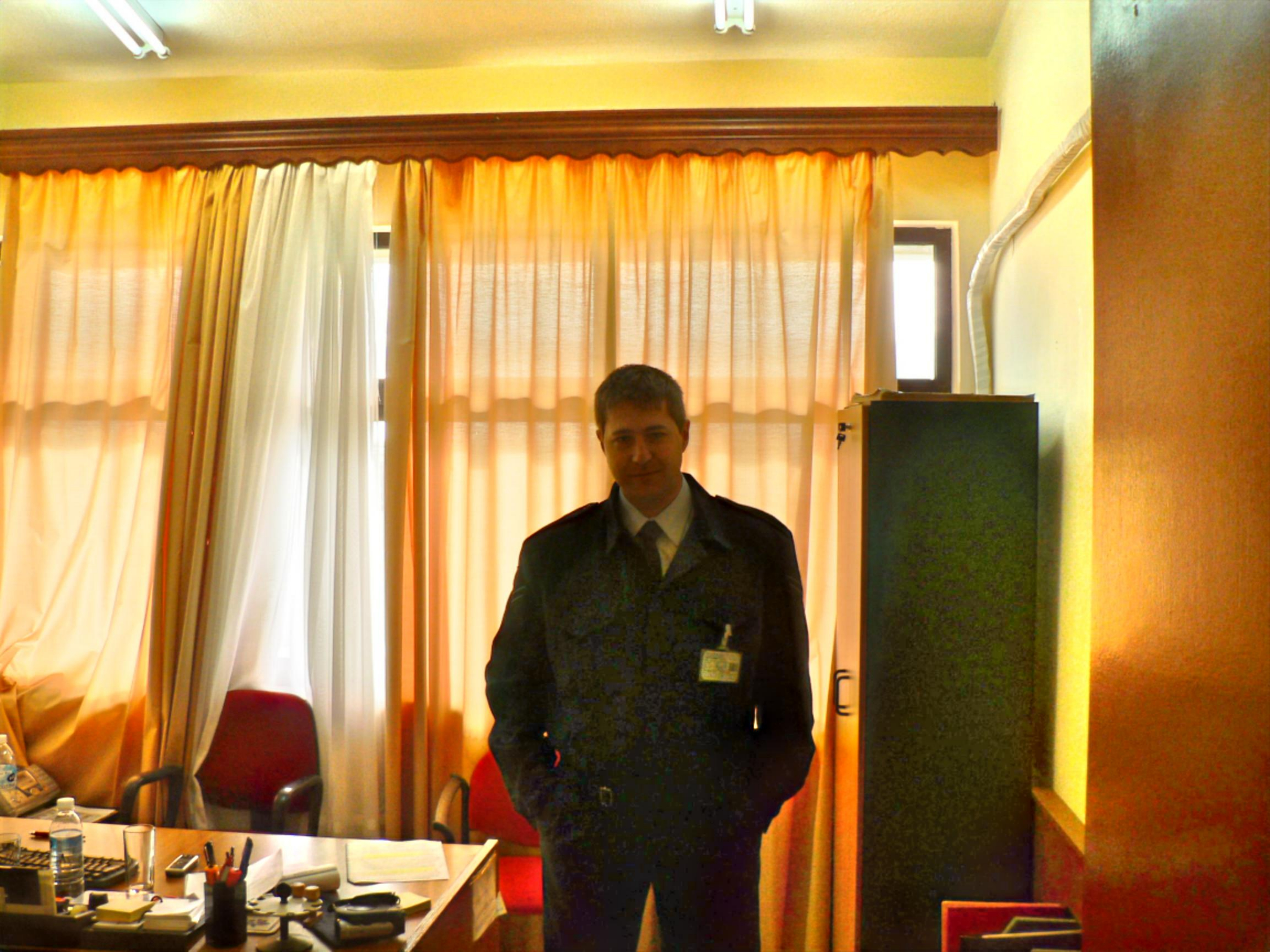}&
			\includegraphics[height=2.6cm,width=3.4cm]{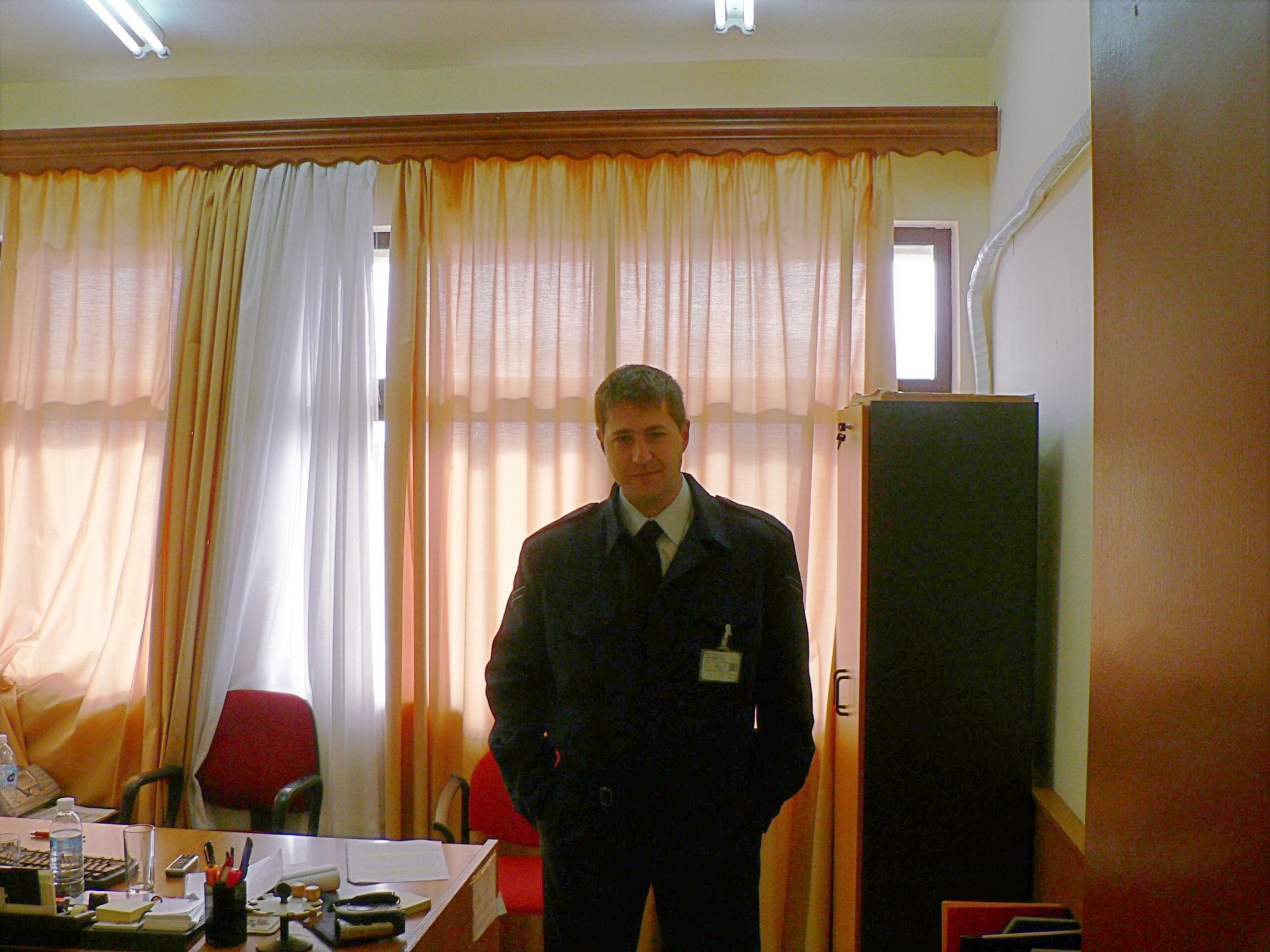}&
			\includegraphics[height=2.6cm,width=3.4cm]{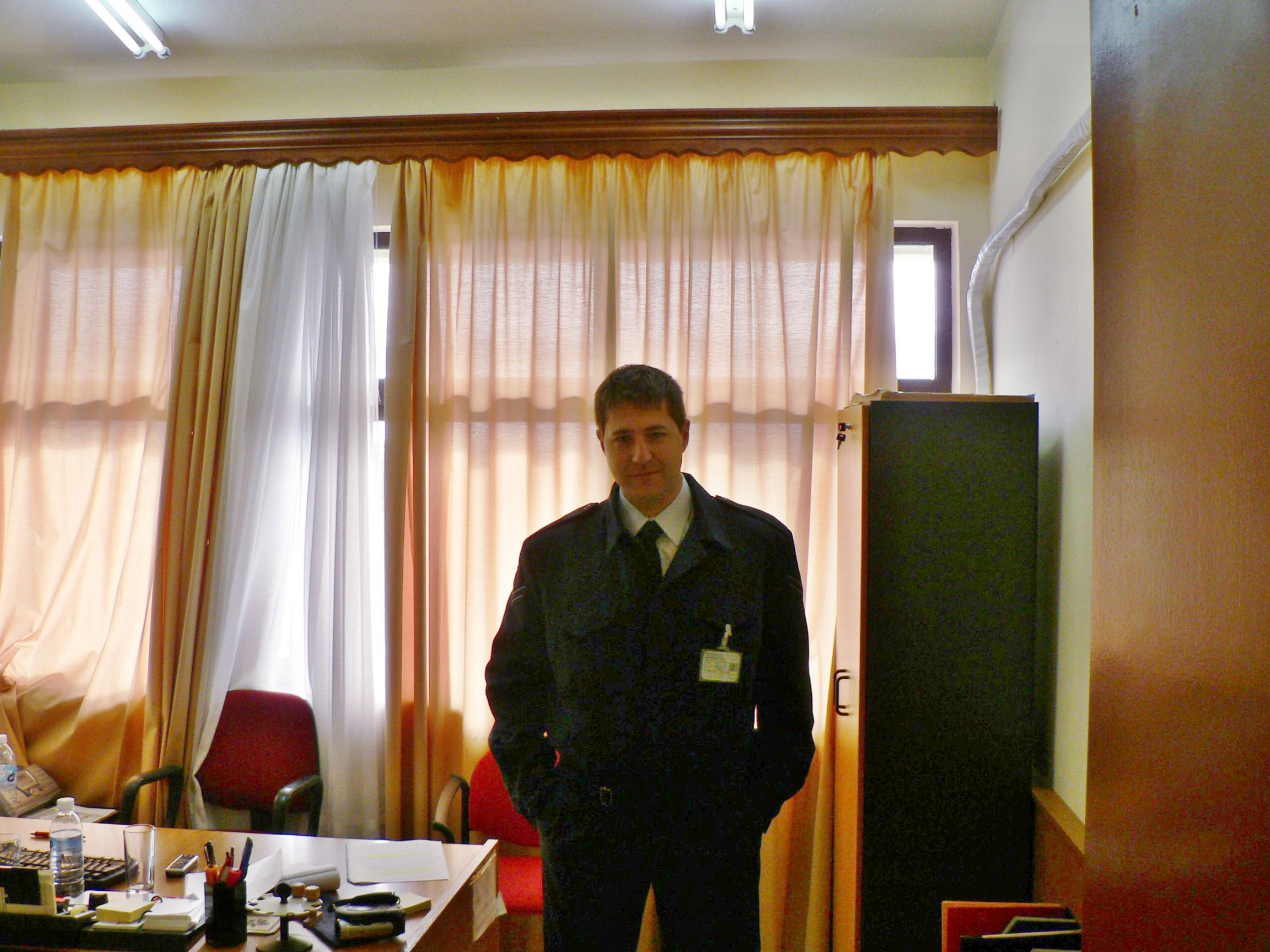} \\
			(a) input & (b) Wang \etal~\cite{Wang2019} & (c) EnlightenGAN~\cite{Jiang2019} & (d) Zero-DCE &	(e) Zero-DCE++\\
		\end{tabular}
	\end{center}
	\caption{Visual comparisons on a typical low-light image comprising nonuniform illumination. The proposed Zero-DCE and Zero-DCE++ achieve visually pleasing results in terms of brightness, color, contrast, and naturalness, while existing methods either fail to cope with the extreme back light or generate color artifacts. In contrast to other deep learning-based methods, our approach is trained without any reference image.}
	\label{fig:im_sample2}
\end{figure*}

An example of enhancing a low-light image comprising nonuniform illumination is shown in Figure~\ref{fig:im_sample2}.
Comparing to state-of-the-art methods, both Zero-DCE and Zero-DCE++ brighten up the image while preserving the inherent color and details. In contrast, both CNN-based method~\cite{Wang2019} and GAN-based method~\cite{Jiang2019} yield under-(the face) and over-(the cabinet) enhancement.
We show in this paper that our method obtains state-of-the-art performance both in qualitative and quantitative metrics. In addition, it is capable of improving high-level visual tasks, \eg, face detection, without inflicting high computational burden.

Our \textbf{contributions} are summarized as follows.
\begin{itemize}
	\item We propose the first low-light enhancement network that is independent of paired and unpaired training data, thus avoiding the risk of overfitting. As a result, our method generalizes well to various lighting conditions.
	\item We design an image-specific curve that is able to approximate pixel-wise and higher-order curves by iteratively applying itself. Such an image-specific curve can effectively perform mapping within a wide dynamic range.
	\item We show the potential of training a deep image enhancement model in the absence of reference images through task-specific non-reference loss functions that indirectly evaluate enhancement quality.
	\item The proposed Zero-DCE can be accelerated considerably while still keeping impressive enhancement performance. We provide multiple options to balance the enhancement performance and the cost of computational resources.
\end{itemize}

This work is an extension of our earlier conference version that has appeared in CVRP2020\cite{Guo2020CVPR}. In comparison to the conference version, we have introduced a significant amount of new materials. 
1) We investigate the relations between the enhancement performance and the network structure, curve estimation, and input sizes. According to the investigation, we re-design the network structure, reformulate the curve formation, and control  the sizes of input image, and thus present an accelerated and light version, called Zero-DCE++, which is more suitable for real-time enhancement on resource-limited devices. 
2) Comparing to our earlier work, without compromising  the enhancement performance, the trainable parameters (79K) and floating  point operations (FLOPs) (84.99G) for an input image of size 1200$\times$900$\times$3 of Zero-DCE are reduced to 10K and 0.115G on Zero-DCE++. This translates to two times in runtime speed up, from 500 FPS in Zero-DCE to 1000 FPS in Zero-DCE++, for processing an image of size 1200$\times$900$\times$3 on a single NVIDIA 2080Ti GPU. In addition, even only with Intel Core i9-10920X CPU@3.5GHz, the processing time of Zero-DCE also can be reduced from 10s to 0.09s on Zero-DCE++, a 111 times speed up on a single CPU setting. The training time is also reduced from 30 minutes to 20 minutes.
3) We perform more experiments, design analysis, and ablation studies to demonstrate the advantages of zero-reference learning for low-light image enhancement and show the effectiveness of our method over existing state-of-the-art methods.
4) We conduct a more comprehensive literature survey on low-light image enhancement and discuss the advantages and limitations of current methods.

\section{Related Work}
\label{sec:Related_Work}
Our work is a new attempt for low-light image enhancement by combining zero-reference learning with  deep curve estimation, which is rarely touched in the previous works. In what follows, we review the low-light image enhancement related works, including conventional methods and data-driven methods.

\noindent
\textbf{Conventional Methods.}
Histogram Equalization (HE)-based methods perform light enhancement through expanding the dynamic range of an image. Histogram distribution of images is adjusted at both global~\cite{Coltuc2006,Ibrahim2007} and local levels~\cite{Stark2000,Lee2013}. There are also various methods adopting the Retinex theory~\cite{Land1977} that typically decomposes an image into reflectance and illumination. The reflectance component is commonly assumed to be consistent under any lighting conditions; thus, light enhancement is formulated as an illumination estimation problem. Building on the Retinex theory, several methods have been proposed. Wang \etal~\cite{Wang2013} designed a naturalness- and information-preserving method when handling images of nonuniform illumination; Fu \etal~\cite{Fu2016} proposed a weighted variation model to simultaneously estimate the reflectance and the illumination of an image. The estimated reflectance is treated as the enhanced result; Guo \etal~\cite{Guo2017} first estimated a coarse illumination map by searching the maximum intensity of each pixel position, then refining the illumination map by a structure prior; Li \etal~\cite{Li2018} proposed a new Retinex model that takes noise into consideration. The illumination map was estimated by  solving an optimization problem.

Contrary to the conventional methods that fortuitously change the distribution of image histogram or that rely on potentially inaccurate physical models, the proposed method produces an enhanced result through image-specific curve mapping. Such a strategy enables light enhancement on images without creating unrealistic artifacts.
Yuan and Sun \cite{Yuan2012} proposed an automatic exposure correction method, where the S-shaped curve for a given image is estimated by a global optimization algorithm and each segmented region is pushed to its optimal zone by curve mapping. Different from \cite{Yuan2012}, our method is  purely data-driven and takes multiple light enhancement factors into consideration in the design of the non-reference loss functions, and thus enjoys better robustness, wider image dynamic range adjustment, and lower computational burden.

\begin{figure*}[t]
	\centering
	\centerline{\includegraphics[width=\linewidth]{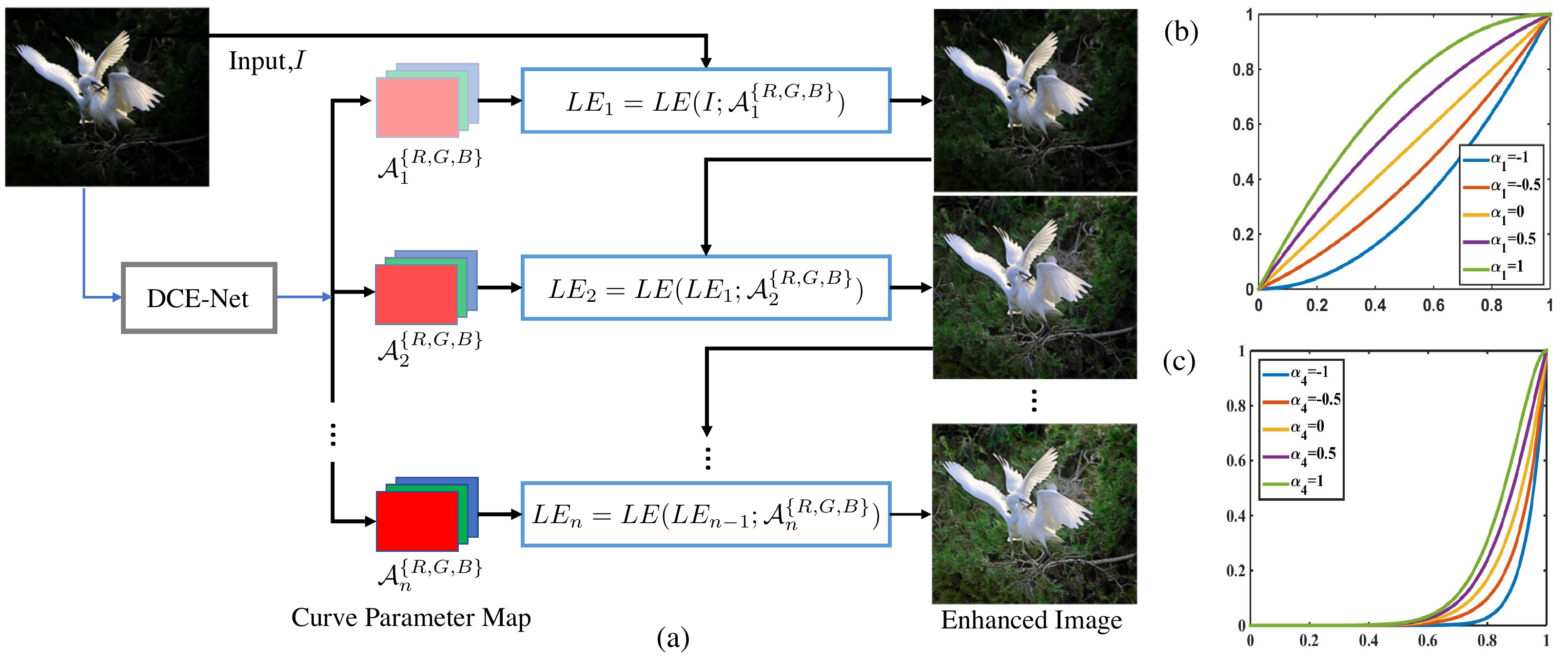}}
	\caption{(a) The framework of Zero-DCE. A DCE-Net is devised to estimate a set of best-fitting Light-Enhancement curves (LE-curves) that iteratively enhance a given input image (\ie, takes the enhanced image as the input of next iteration and the input is enhanced in a progressive manner). (b, c) LE-curves with different adjustment parameters $\alpha$ and numbers of iteration $n$. In (c), $\alpha_{1}$, $\alpha_{2}$, and $\alpha_{3}$ are equal to -1 while $n$ is equal to 4. In each subfigure, the horizontal axis represents the input pixel values while the vertical axis represents the output pixel values.}
	\label{fig:pipeline}
\end{figure*}

\noindent
\textbf{Data-Driven Methods.}
Data-driven methods are largely categorized into two branches, namely Convolutional Neural Network (CNN)-based and Generative Adversarial Network (GAN)-based methods. Most CNN-based solutions rely on paired data for supervised training, therefore they are resource-intensive. Often time, the paired data are exhaustively collected through automatic light degradation, changing the settings of cameras during data capturing, or synthesizing data via image retouching. For example, LL-Net~\cite{Lore2017} and MBLLEN~\cite{MBLLEN} were trained on data simulated on random Gamma correction; the LOL dataset~\cite{Chen2018} of paired low/normal light images was collected through altering the exposure time and ISO during image acquisition; the MIT-Adobe FiveK dataset~\cite{Adobe5K} comprises 5,000 raw images, each of which has five retouched images produced by trained experts. MIT-Adobe FiveK dataset was originally collected for image global retouching;  the SID~\cite{Chenchen2018} provides paired low/normal light raw  data;
a dataset of raw low-light videos with the corresponding normal light videos captured at video rate was collected in~\cite{Chenchen2019}.

Inspired by the Retinex model, recent deep models design the networks to estimate the reflectance and illumination of an input image by supervised learning with paired data. 
Ren \etal~\cite{Ren2019} proposed a deep hybrid network for low-light image enhancement, which consists of two  streams to learn the global content and the salient structures in a unified network.
Wang \etal~\cite{Wang2019} proposed an underexposed photo enhancement network by estimating the illumination map. This network was trained on paired data that were retouched by three experts.
Zhang \etal~\cite{Zhang2019ACM} built a network  for kindling the darkness of an image, called KinD, which decomposes images into two components. The illumination component is responsible for the light adjustment while the reflectance component is for degradation removal.
Retinex model-based deep models still suffer from the same limitations as the conventional Retinex-based methods, such as ideal assumption.

More recently, Xu \etal~\cite{Xu2020CVPR} proposed a frequency-based decomposition-and-enhancement model for low-light image enhancement. This model first recovers the image content in the low-frequency layer, then enhances high-frequency details  based on the recover image content. This model is trained on a low-light dataset of real noisy low-light and ground truth sRGB image pairs.

Understandably, light enhancement solutions based on paired data are impractical in many ways, considering the high cost involved in collecting sufficient paired data as well as the inclusion of factitious and unrealistic data in training the deep models. Such constraints are reflected in the poor generalization capability of CNN-based methods. Artifacts and color casts are commonly generated, when these methods are presented with real-world images of various light intensities.

Unsupervised GAN-based methods have the advantage of eliminating paired data for training.  An unsupervised GAN-based method, EnlightenGAN~\cite{Jiang2019}, learns to enhance low-light images using unpaired low/normal light data. The network was trained by taking  elaborately designed discriminators and loss functions into account. However, unsupervised GAN-based solutions usually require careful selection of unpaired training data.

To integrate the superiorities  of CNNs and GANs, Yang \etal~\cite{Yang2020CVPR} proposed a semi-supervised model for low-light image enhancement, which performs enhancement in two stages. In the first stage, a coarse-to-fine band representation is learned and different band signals are inferred jointly in a recursive process with paired data. In the second stage, the band representation is recomposed via adversarial learning. Although the semi-supervised learning framework can effectively improve the generalization capability  of the deep model, it still takes the risk of overfitting on paired training data and induces high memory footprint.

\begin{figure*}
	\begin{center}
		\begin{tabular}{c@{ }c@{ }c@{ }c@{ }c@{ }c}
			\includegraphics[width=.15\textwidth,height=3cm]{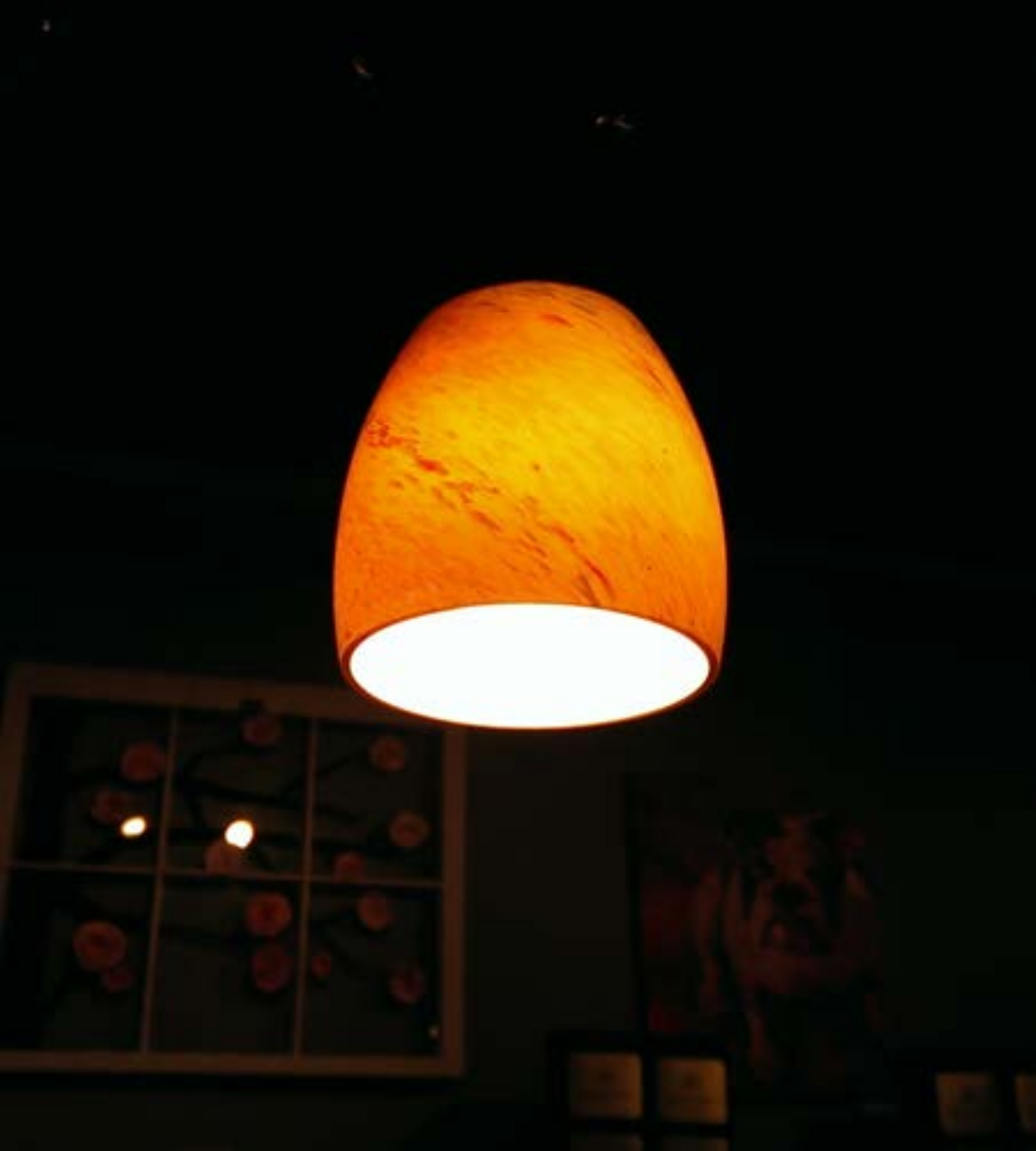}~~~~~~~&
			\includegraphics[width=.175\textwidth,height=3.05cm]{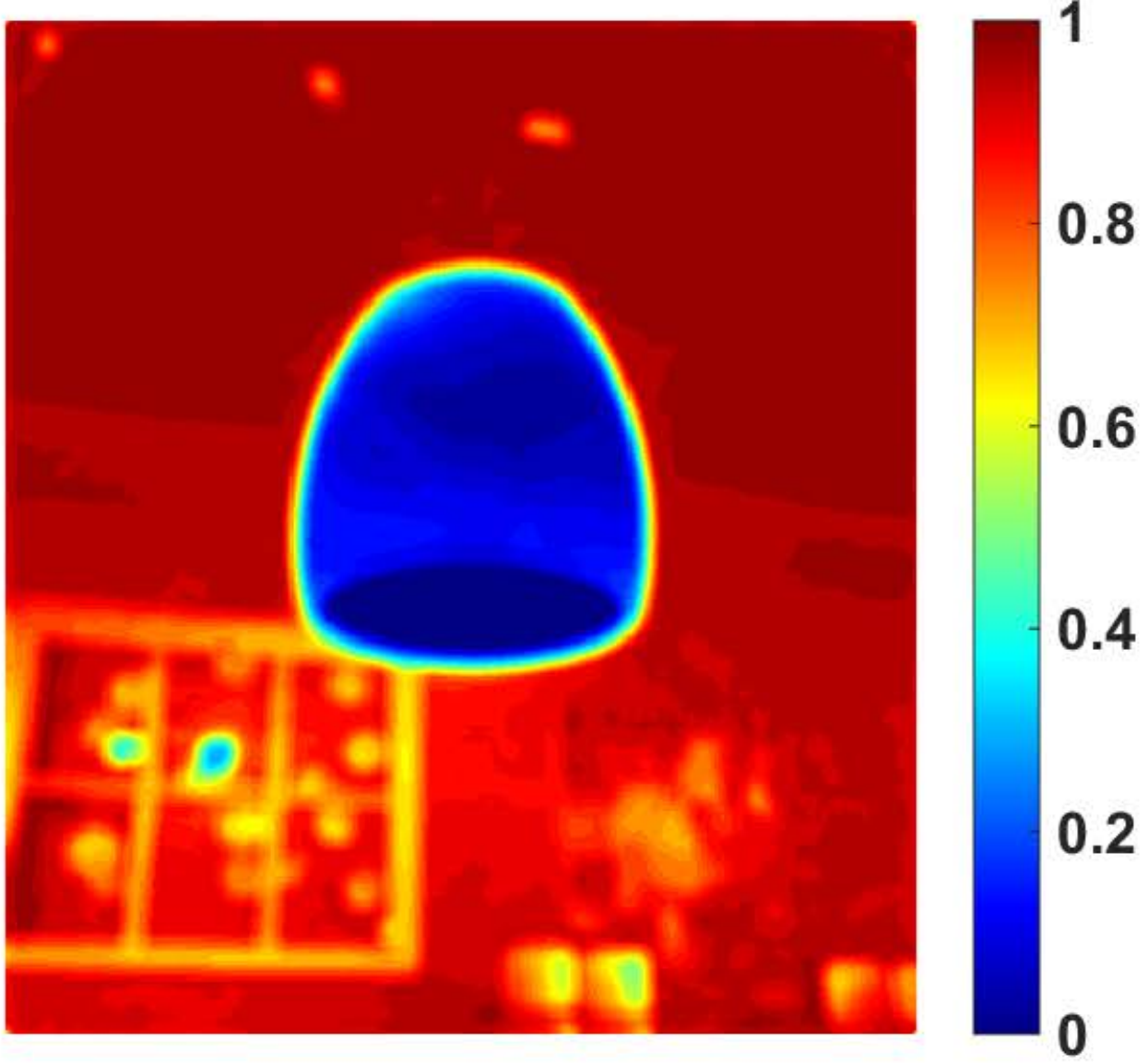}~~~~~~~&
			\includegraphics[width=.15\textwidth,height=3cm]{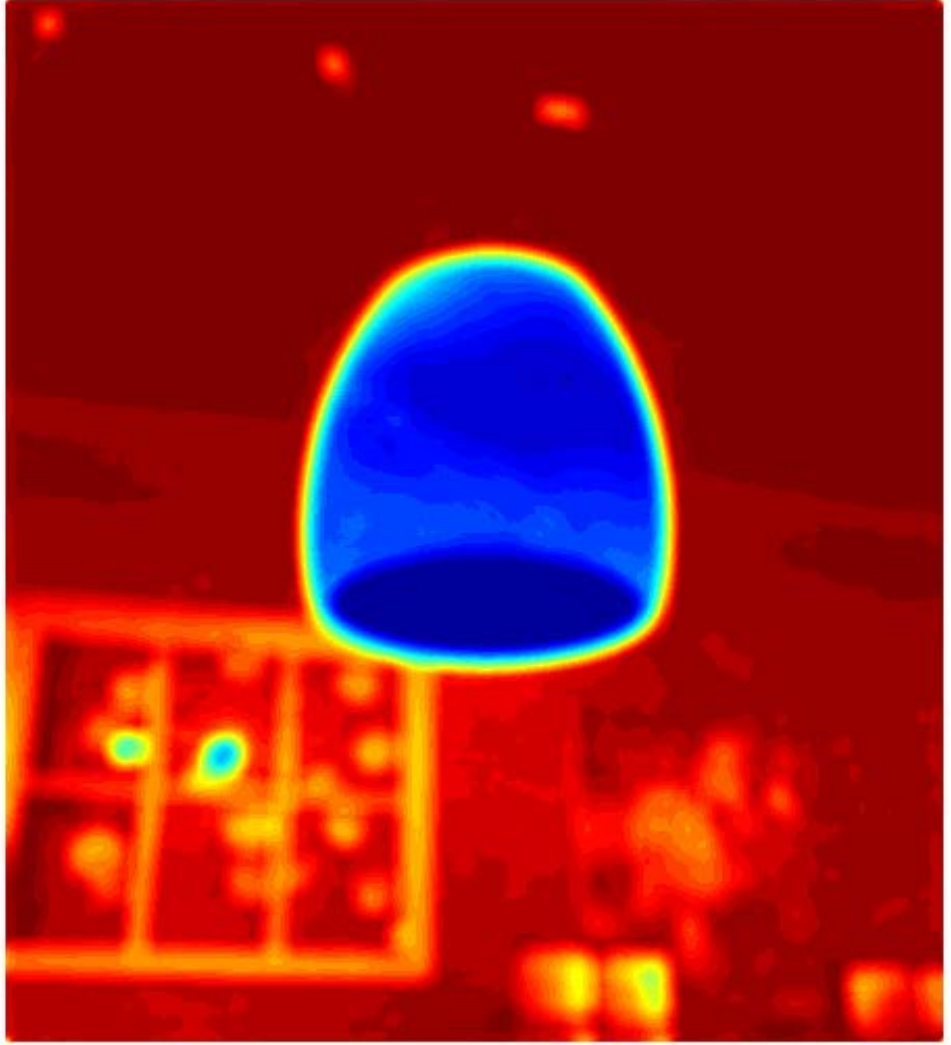}~~~~~~~&
			\includegraphics[width=.15\textwidth,height=3cm]{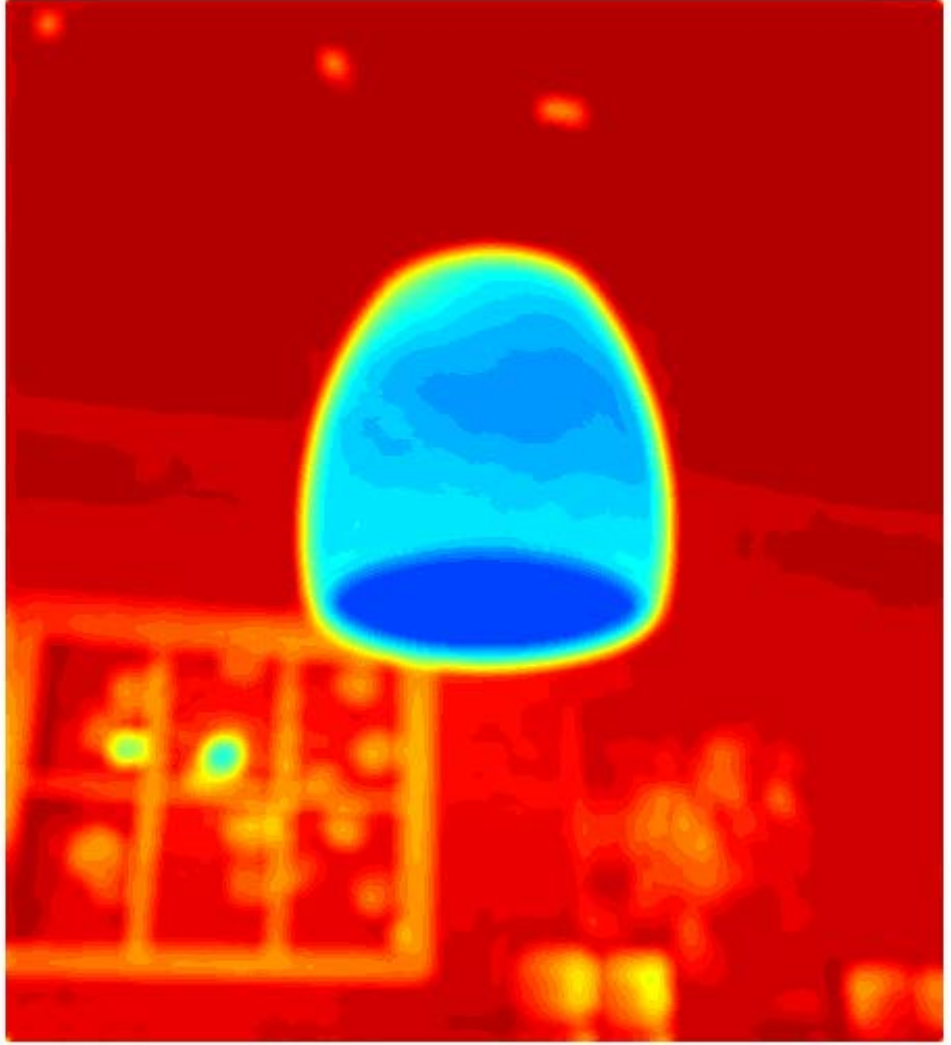}~~~~~~~&
			\includegraphics[width=.15\textwidth,height=3cm]{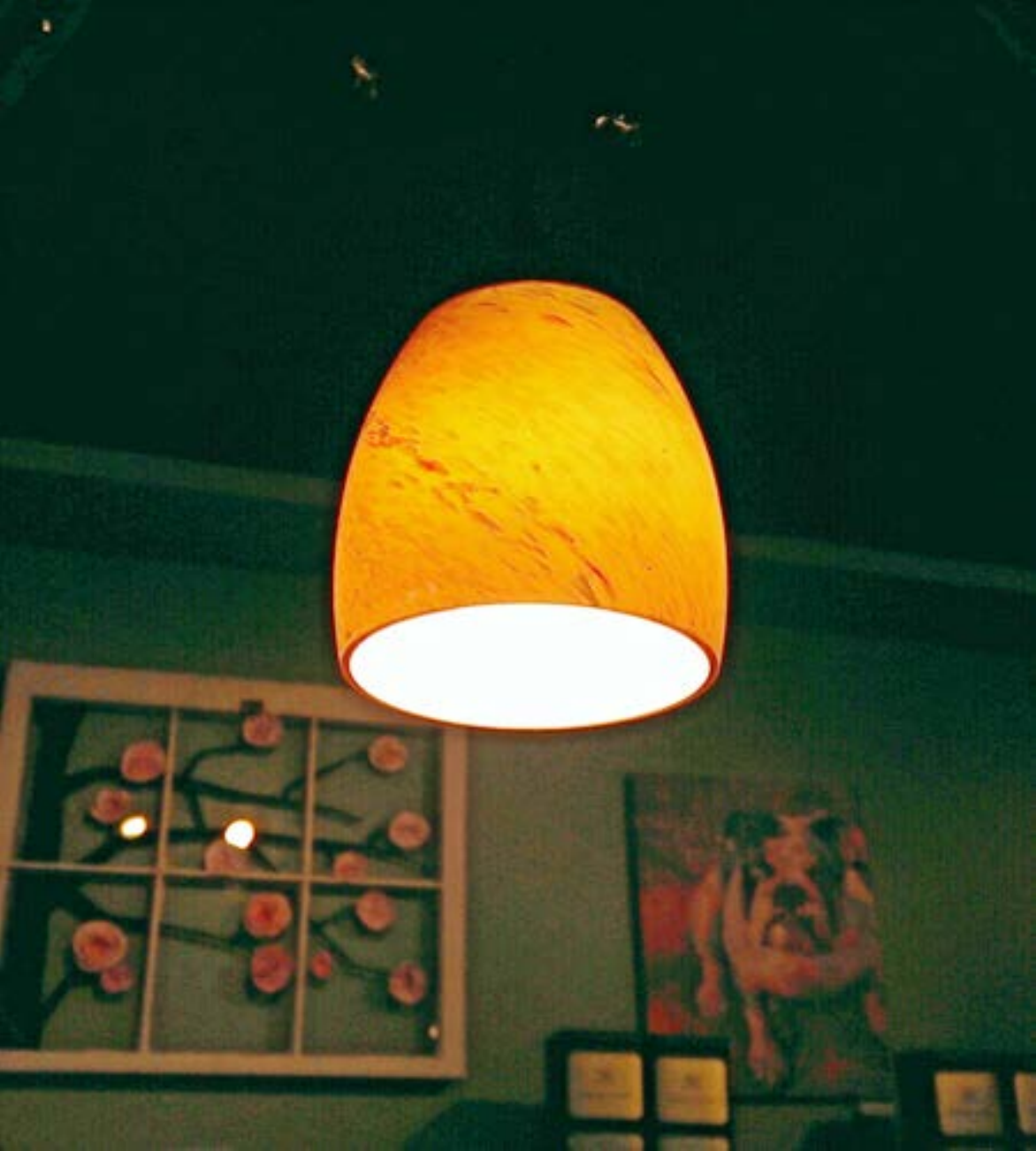}\\
			(a) input~~~~~~& (b) $\mathcal{A}_{n}^{R}$~~~~~~& (c) $\mathcal{A}_{n}^{G}$~~~~~~& (d) $\mathcal{A}_{n}^{B}$~~~~~~& (e) result \\
		\end{tabular}
	\end{center}
	\caption{An example of the pixel-wise curve parameter maps. For visualization, we average the curve parameter maps of all iterations ($n=8$) and normalize the values to the range of $[0,1]$. $\mathcal{A}_{n}^{R}$, $\mathcal{A}_{n}^{G}$, and $\mathcal{A}_{n}^{B}$ represent the averaged best-fitting curve parameter maps of R, G, and B channels, respectively. The maps in (b), (c), and (d) are represented by heatmaps.}
	\label{fig:map}
\end{figure*}

The proposed method is superior to existing data-driven methods in three aspects. First, it explores a new learning strategy, \ie, one requires \textit{zero reference}, hence eliminating the need for paired and unpaired data. Second, the network is trained by taking  carefully defined non-reference loss functions into account. This strategy allows output image quality to be implicitly evaluated, the results of which would be reiterated for network learning. Third, our method is highly efficient and cost-effective. The accelerated and light version Zero-DCE++ only contains 10K trainable parameters and 0.115G FLOPs, achieves 1000/11 FPS inference time on a single GPU/CPU, and needs 20 minutes for training. The efficiency of our method precedes current deep models \cite{MBLLEN,Chen2018,Wang2019,Jiang2019} by a large margin. These advantages benefit from our zero-reference learning framework, lightweight network structure, and effective non-reference loss functions.

\section{Methodology}
\label{sec:Method}
We show the framework of Zero-DCE in Figure~\ref{fig:pipeline}.
A Deep Curve Estimation Network (DCE-Net) is devised to estimate a set of best-fitting Light-Enhancement curves (LE-curves) given an input image.
The framework then maps all pixels of the input's RGB channels by applying the curves iteratively for obtaining the final enhanced image.
In what follows, we detail the key components, namely LE-curve, DCE-Net, and non-reference loss functions.

\subsection{Light-Enhancement Curve}

Inspired by curve adjustment used in photo editing software, we design a kind of curve that can map a low-light image to its enhanced version automatically, where the self-adaptive curve parameters are solely dependent on the input image.
There are three objectives in the design of such a curve: 
\begin{enumerate}
	\item each pixel value of the enhanced image should fall in the normalized range of [0,1] to avoid information loss induced by overflow truncation;
	\item  this curve should be monotonous to preserve the differences (contrast) of neighboring pixels; and 
	\item the form of this curve should be as simple as possible and differentiable in the process of gradient backpropagation.
\end{enumerate}

To achieve these three objectives, we design a quadratic curve, which can be expressed as:
\begin{equation}
\label{equ_1}
LE(I(\mathbf{x});\alpha)=I(\mathbf{x})+\alpha I(\mathbf{x})(1-I(\mathbf{x})),
\end{equation}
where $\mathbf{x}$ denotes pixel coordinates, $LE(I(\mathbf{x});\alpha)$ is the enhanced version of the given input $I(\mathbf{x})$,  the trainable curve parameter $\alpha\in[-1,1]$ adjusts the magnitude of LE-curve and also controls the exposure level. Each pixel of input is normalized to the range of $[0,1]$ and all operations are pixel-wise.
We separately apply the LE-curve to three RGB channels instead of solely on the luminance channel. The three-channel adjustment can better preserve the inherent color and reduce the risk of over-saturation. We report more details in the ablation study.

An illustration of LE-curves with different adjustment parameters $\alpha$ is shown in Figure~\ref{fig:pipeline}(b).
It is clear that the LE-curve complies with the three aforementioned objectives. Thus, each pixel value of enhanced images is in the range of [0,1]. 
In addition, the LE-curve enables us to increase or decrease the dynamic range of an input image. This capability is conducive to not only enhancing low-light regions but also removing over-exposure artifacts.
We choose a specific single-parameter form  for the quadratic because 1) the single-parameter form can reduce the computational cost and speed up our method and 2) the specially designed quadratic meets the three objectives of our designs and already achieves satisfactory enhancement performance.

\noindent
\textbf{Higher-Order Curve.}
The LE-curve defined in Equation~\eqref{equ_1} can be applied iteratively to enable more versatile adjustment to cope with challenging low-light conditions. Specifically,
\begin{equation}
\label{equ_2}
LE_{n}(\mathbf{x})=LE_{n-1}(\mathbf{x})+\alpha_{n}LE_{n-1}(\mathbf{x})(1-LE_{n-1}(\mathbf{x})),
\end{equation}
where $n$ is the number of iteration, which controls the curvature. In this paper, we set the value of $n$ to 8, which can deal with most cases satisfactorily. Equation~\eqref{equ_2} can be degraded to Equation~\eqref{equ_1} when $n$ is equal to 1. Figure~\ref{fig:pipeline}(c) provides an example showing high-order curves with different $\alpha$ and $n$. Such high-order curves offer more powerful adjustment capability (\ie, greater curvature) than the curves in Figure~\ref{fig:pipeline}(b).

\noindent
\textbf{Pixel-Wise Curve.}
In comparison to a single-order curve, a higher-order curve adjusts an image within a wider dynamic range. Nonetheless, it is still a global adjustment since $\alpha$ is used for all pixels.
A global mapping tends to over-/under- enhance local regions.
To address this problem, we formulate $\alpha$ as a pixel-wise parameter, \ie, each pixel of the given input image has a corresponding curve with the best-fitting $\alpha$ to adjust its dynamic range.
Hence, Equation~\eqref{equ_2} can be reformulated as:
\begin{equation}
\label{equ_3}
LE_{n}(\mathbf{x})=LE_{n-1}(\mathbf{x})+\mathcal{A}_{n}(\mathbf{x})LE_{n-1}(\mathbf{x})(1-LE_{n-1}(\mathbf{x})),
\end{equation}
where $\mathcal{A}$ is a parameter map with the same size as the given image.
Here, pixels in a local region are assumed to having the same intensity (also the same adjustment curves), and thus the neighboring pixels in the output result still preserve the monotonous relations.
In this way, the pixel-wise higher-order curves also comply with the three aforementioned objectives.
As a result, each pixel value of enhanced images is still in the range of [0,1]. 

We present an example of the estimated curve parameter maps in Figure~\ref{fig:map}.
As shown, the best-fitting parameter maps of different color channels have similar adjustment tendency but different values, indicating the relevance and difference among the three channels of a low-light image.
The curve parameter map accurately indicates the brightness of different regions (\eg, the two glitters on the wall).
With the fitting maps, the enhanced version image can be directly obtained by pixel-wise
curve mapping. As shown in Figure~\ref{fig:map}(e), the enhanced version reveals the content in dark regions and preserves the bright regions.

\begin{figure}[t]
	\centering
	\centerline{\includegraphics[width=1\linewidth]{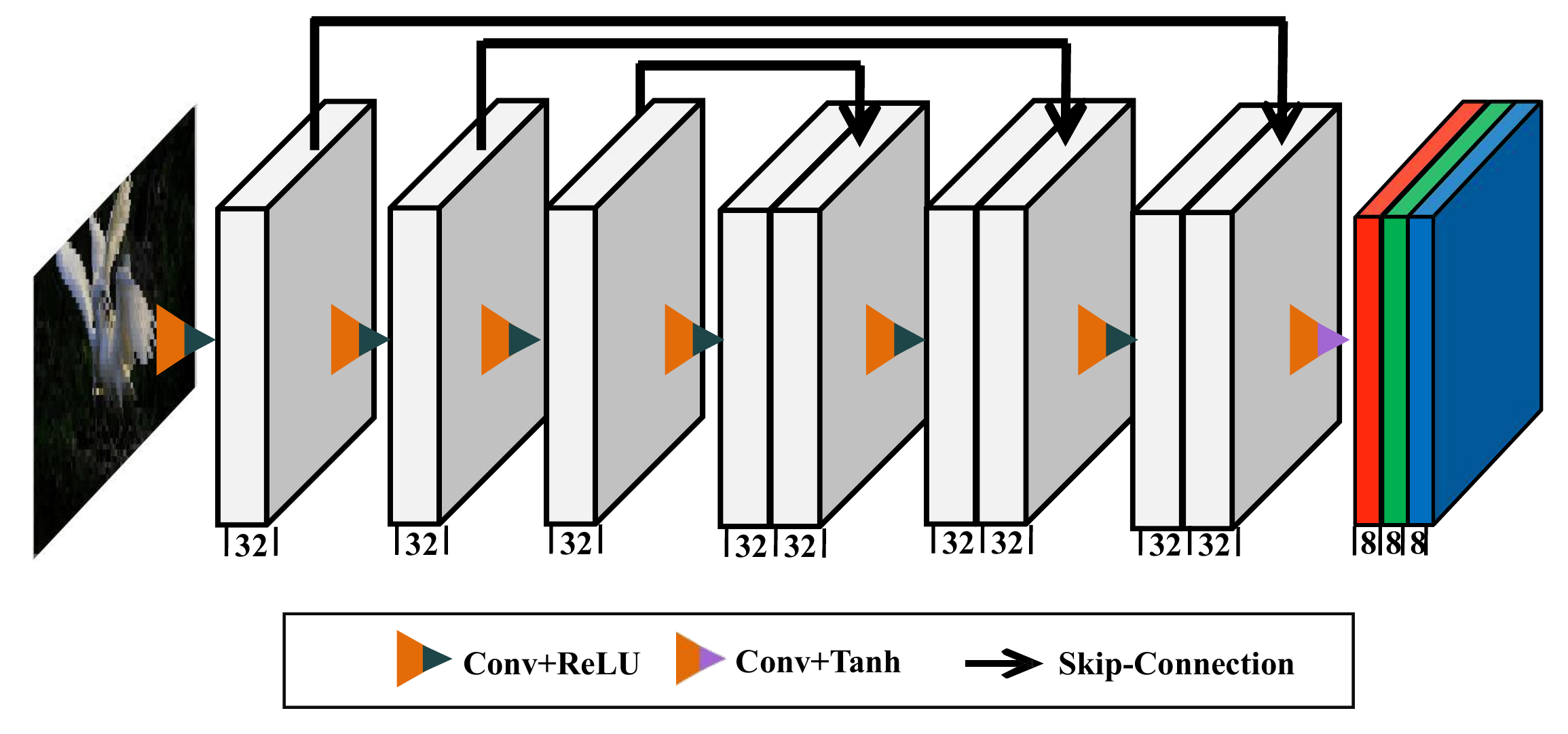}}
	\caption{The architecture of DCE-Net.}
	\label{fig:network}
\end{figure}

\subsection{DCE-Net}
To learn the mapping between an input image and its best-fitting curve parameter maps, we propose a Deep Curve Estimation Network (DCE-Net). In Figure~\ref{fig:network}, we present the detailed network architecture and parameter settings of DCE-Net. 

The input to the DCE-Net is a low-light image while the outputs are a set of pixel-wise curve parameter maps for corresponding higher-order curves.
Instead of employing fully connected layers that require fixed input sizes,
we employ a plain CNN of seven convolutional layers with symmetrical skip concatenation. 
In the first six convolutional layers, each convolutional layer consists of 32 convolutional kernels of size 3$\times$3 and stride 1 followed by the ReLU activation function. The last convolutional layer consists of 24 convolutional kernels of size 3$\times$3 and stride 1 followed by the Tanh activation function, which produces 24 curve parameter maps for eight iterations, where each iteration generates three curve parameter maps for the three channels (\ie, RGB channels).
We discard the down-sampling and batch normalization layers that break the relations of neighboring pixels. It is noteworthy that DCE-Net only has 79K trainable parameters and 85G FLOPs for an input image of size 1200$\times$900$\times$3, which is already smaller than existing low-light image enhancement deep models, such as RetinexNet~\cite{Chen2018}: 555K/587G,  EnlightenGAN~\cite{Jiang2019}: 8M/273G), and  MBLLEN~\cite{MBLLEN}: 450K/301G.

\subsection{Non-Reference Loss Functions}

To enable zero-reference learning in DCE-Net, we propose a set of differentiable non-reference losses that allow us to evaluate the quality of enhanced images.
The following four types of losses are adopted to train our DCE-Net.

\begin{figure}[!t]
	\begin{center}
		\begin{tabular}{c@{ }c@{ }}
			\includegraphics[width=0.45\linewidth]{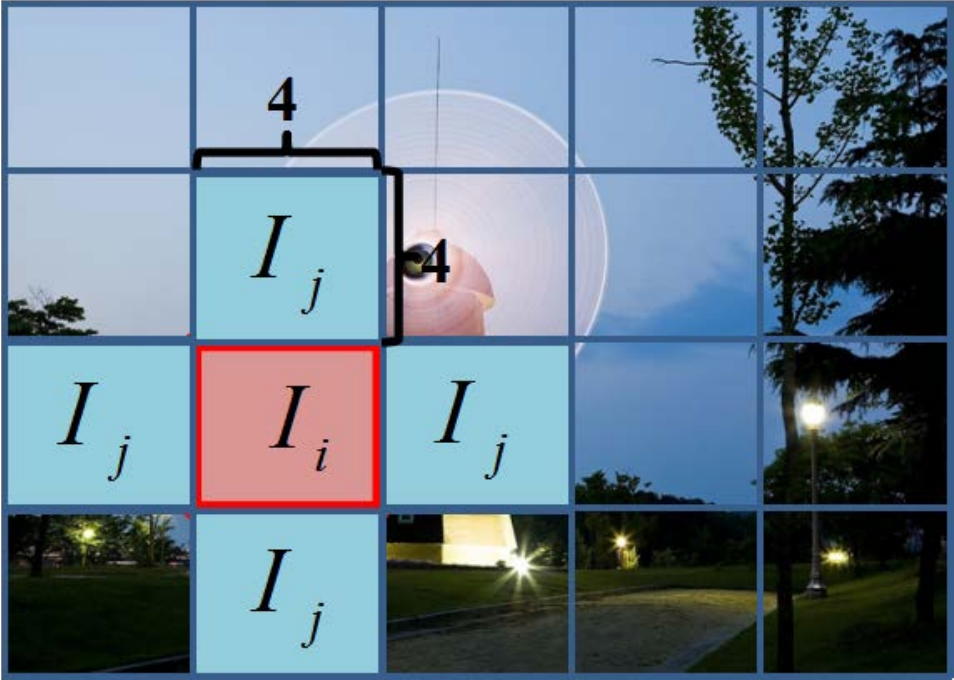}&
			\includegraphics[width=0.45\linewidth]{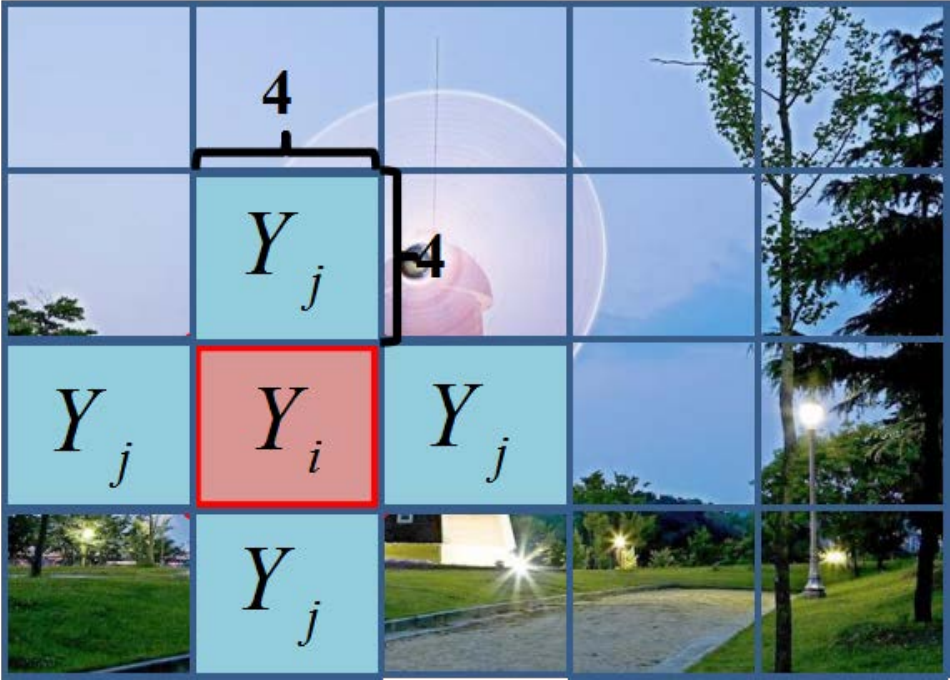}\\
			(a) low-light& (b) enhanced \\
		\end{tabular}
	\end{center}
	\caption{An illustration of the spatial consistency loss.}
\label{fig:spatial_loss}
\end{figure}

\noindent
\textbf{Spatial Consistency Loss.}
The spatial consistency loss $L_{spa}$ encourages spatial coherence of the enhanced image through preserving the difference of neighboring regions between the input image and its enhanced version:
\begin{equation}
\label{equ_4}
L_{spa}=\frac{1}{K}\sum\limits_{i=1}^K\sum\limits_{j\in\Omega(i)}(|(Y_{i}-Y_{j})|-|(I_{i}-I_{j})|)^2,
\end{equation}
where $K$ is the number of local region, and $\Omega$($i$) is the four neighboring regions (top, down, left, right) centered at the region $i$. We denote $Y$ and $I$ as the average intensity value of the local region in the enhanced version and input image, respectively.
We empirically set the size of the local region to 4$\times$4. This loss is stable given other region sizes. We illustrate the process of computing the spatial consistency loss in Figure \ref{fig:spatial_loss}.

\noindent
\textbf{Exposure Control Loss.}
To restrain under-/over-exposed regions, we design an exposure control loss $L_{exp}$ to control the exposure level.
The exposure control loss measures the distance between the average intensity value of a local region to the well-exposedness level $E$.
We follow existing practices~\cite{Exposure2007,Exposure2009} to set $E$ as the gray level in the RGB color space. We empirically set $E$ to 0.6 in our experiments.
The loss $L_{exp}$ can be expressed as:
\begin{equation}
\label{equ_5}
L_{exp}=\frac{1}{M}\sum\limits_{k=1}^M|Y_{k}-E|,
\end{equation}
where $M$ represents the number of non-overlapping local regions of size 16$\times$16, the average intensity value of a local region in the enhanced image is represented as $Y$.

\noindent
\textbf{Color Constancy Loss.}
Following the Gray-World color constancy hypothesis~\cite{Buchsbaum1980} that color in each sensor channel averages to gray over the entire image, we design a color constancy loss to correct the potential color deviations in the enhanced image and also build the relations among the three adjusted channels.
The color constancy loss $L_{col}$ can be expressed as:
\begin{equation}
\label{equ_6}
L_{col}=\sum\limits_{\forall(p,q)\in\varepsilon}(J^{p}-J^{q})^2, \varepsilon=\{(R,G),(R,B),(G,B)\},
\end{equation}
where $J^{p}$ denotes the average intensity value of $p$ channel in the enhanced image,  a pair of channels is represented as ($p$,$q$).

\noindent
\textbf{Illumination Smoothness Loss.} To preserve the monotonicity relations between neighboring pixels, we add an illumination smoothness loss to each curve parameter map $\mathcal{A}$. The illumination smoothness loss $L_{tv_\mathcal{A}}$ is defined as:
\begin{equation}
\label{equ_7}
L_{tv_\mathcal{A}}=\frac{1}{N}\sum\limits_{n=1}^N\sum\limits_{c\in\xi}(|\nabla_{x}\mathcal{A}_{n}^{c}|+|\nabla_{y}\mathcal{A}_{n}^{c}|)^2,  \xi=\{R,G,B\},
\end{equation}
where $N$ is the number of iteration,  the horizontal and vertical gradient operations are represented as $\nabla_{x}$ and $\nabla_{y}$, respectively.

\noindent
\textbf{Total Loss.}
The total loss can be expressed as:
\begin{equation}
\label{equ_9}
L_{total}=L_{spa}+L_{exp}+W_{col}L_{col}+W_{tv_\mathcal{A}}L_{tv_\mathcal{A}},
\end{equation}
where the weights $W_{col}$ and $W_{tv_\mathcal{A}}$  are used for balancing the scales of different losses.

\begin{figure*}[!htb]
	\begin{center}
		\begin{tabular}{c@{ }c@{ }c@{ }c@{ }c@{ }c@{ }}
			\includegraphics[width=.15\textwidth,height=2.5cm]{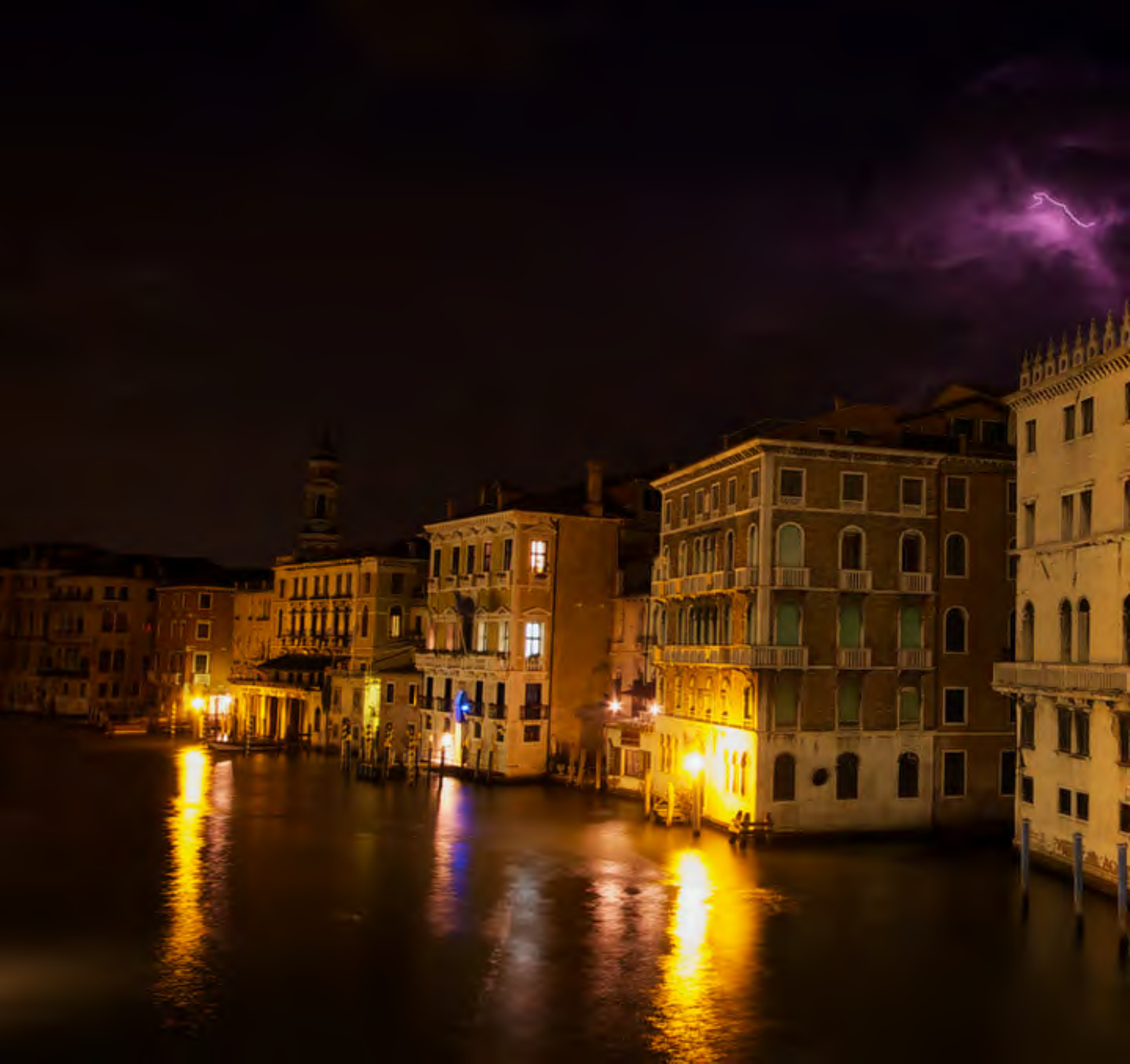}&
			\includegraphics[width=.15\textwidth,height=2.5cm]{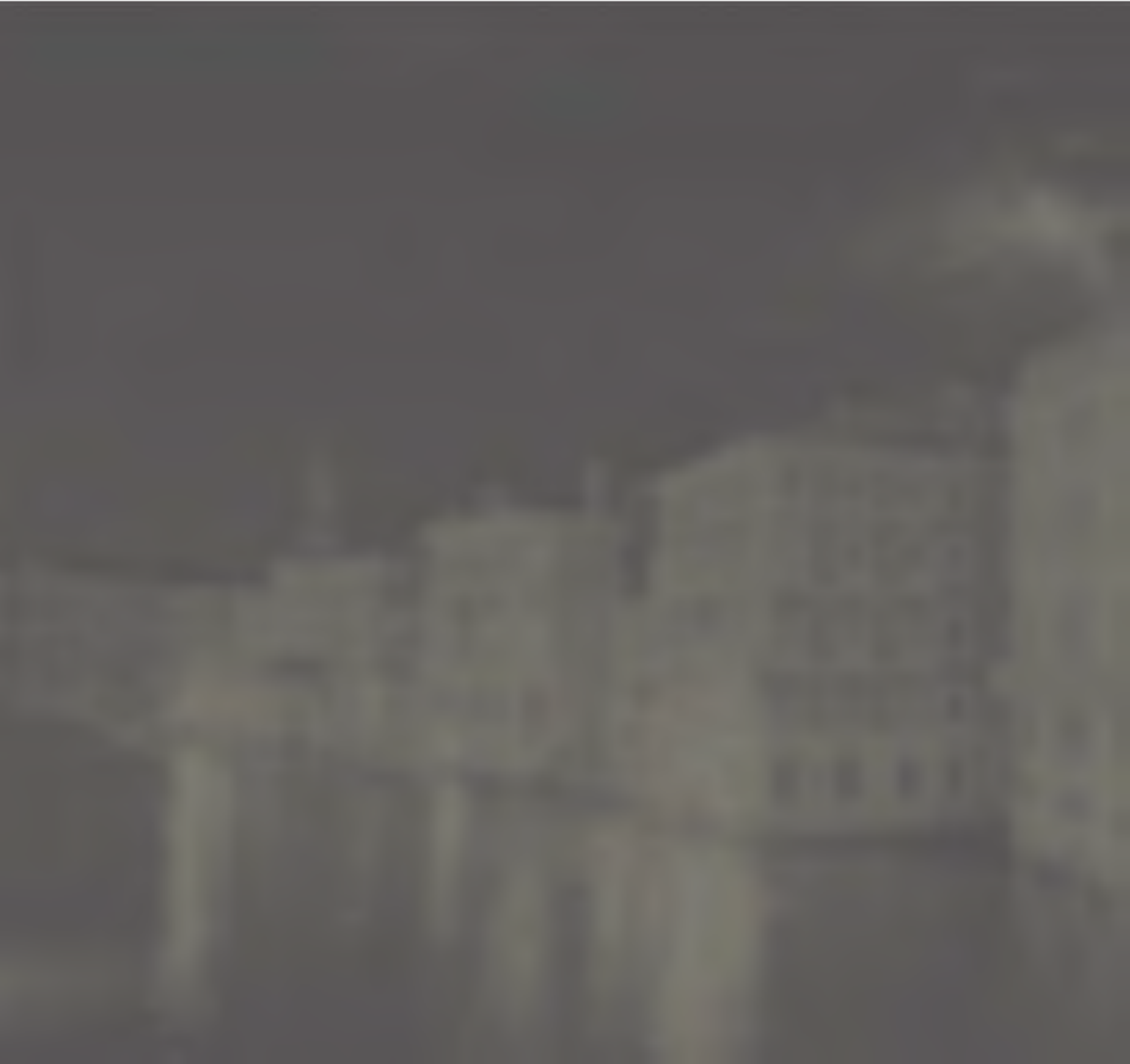}&
			\includegraphics[width=.15\textwidth,height=2.5cm]{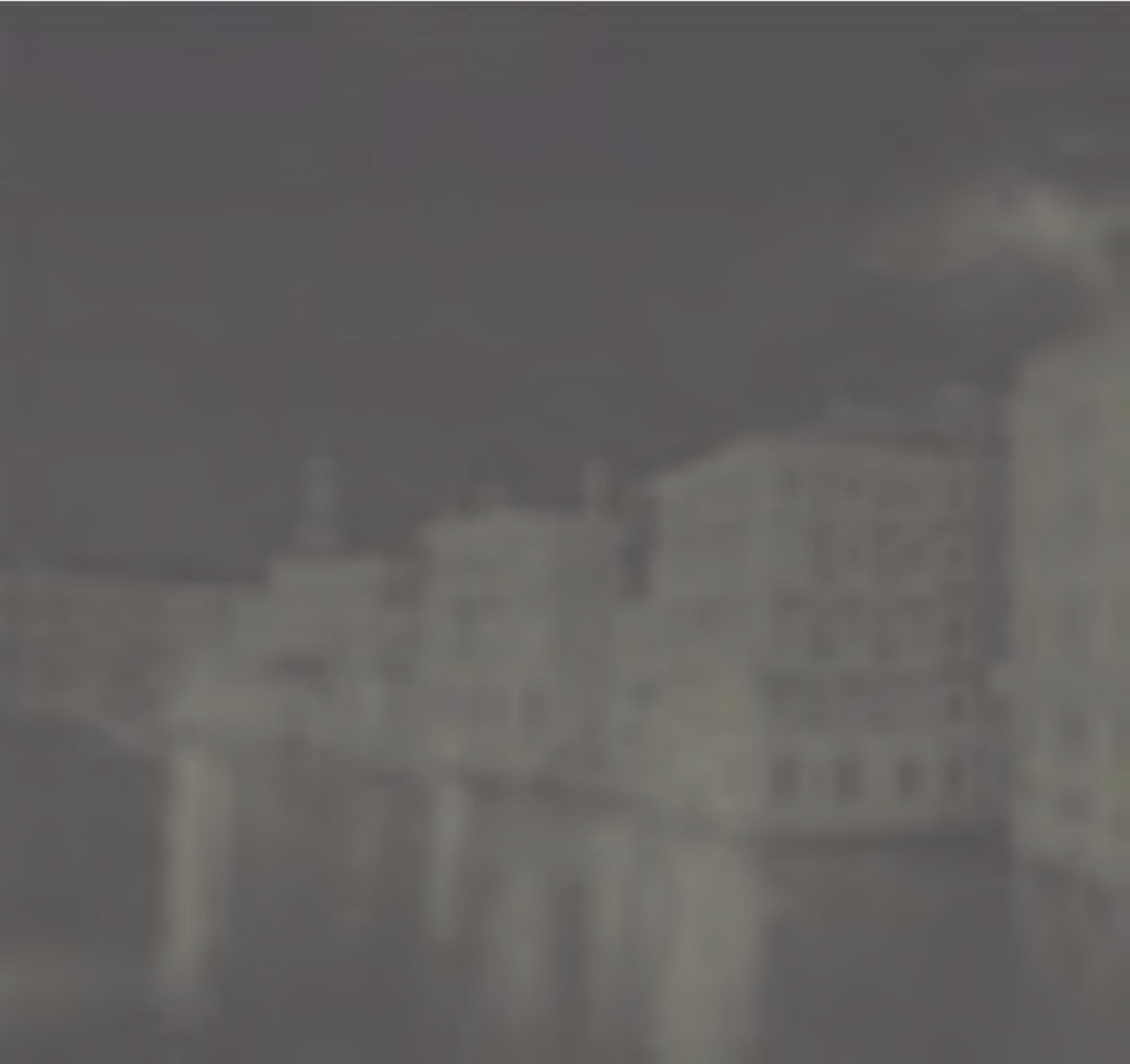}&
			\includegraphics[width=.15\textwidth,height=2.5cm]{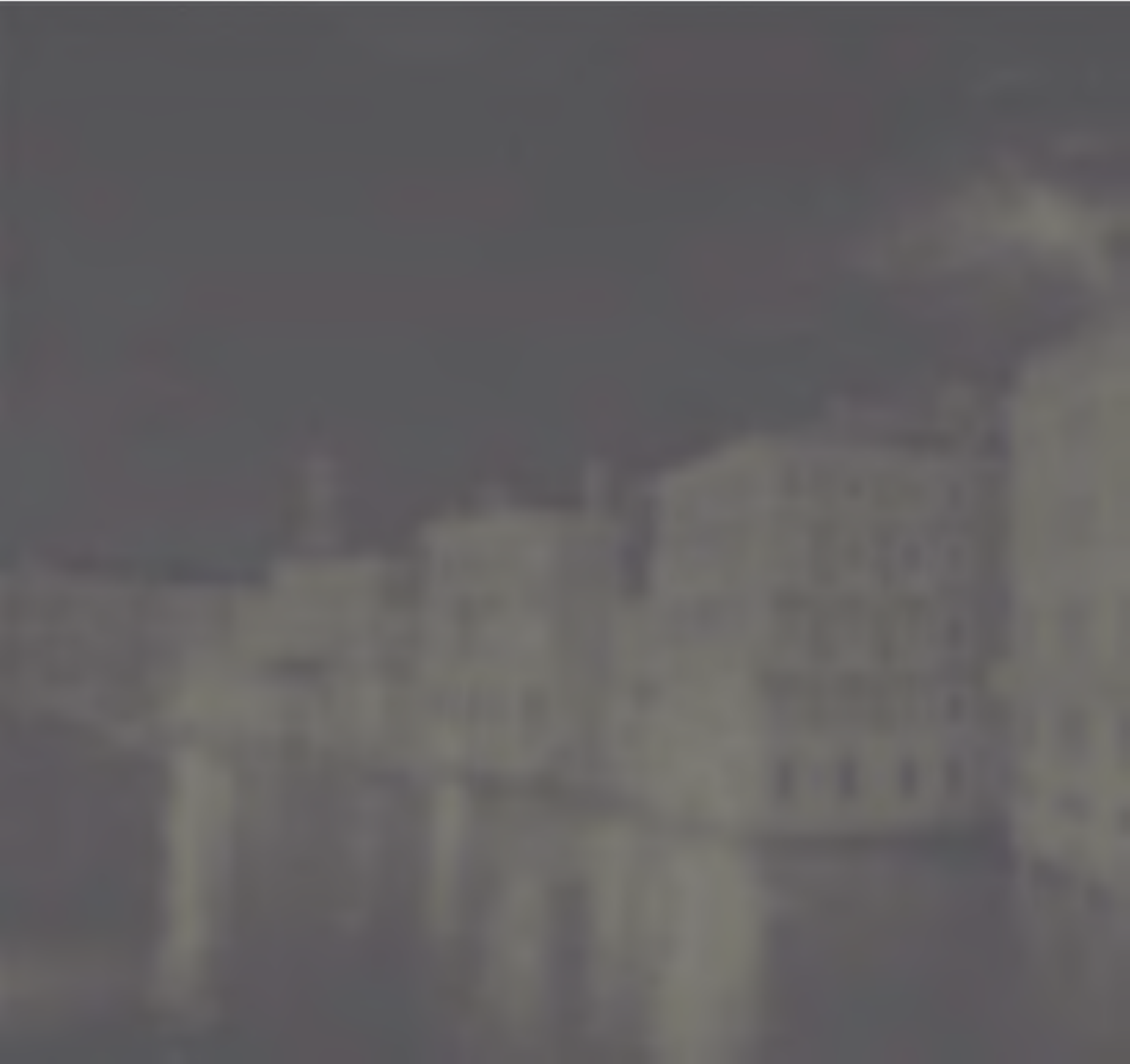}&
			\includegraphics[width=.17\textwidth,height=2.55cm]{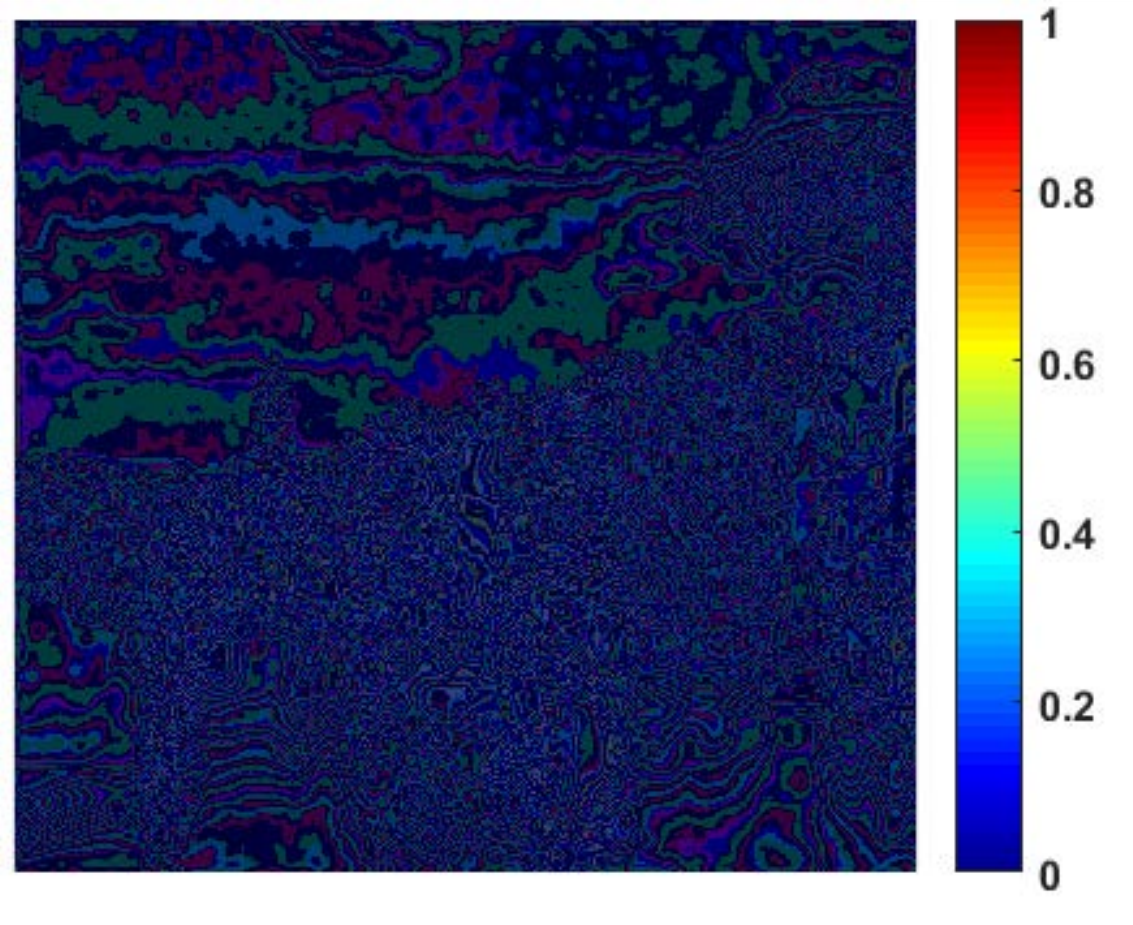}&
			\includegraphics[width=.17\textwidth,height=2.55cm]{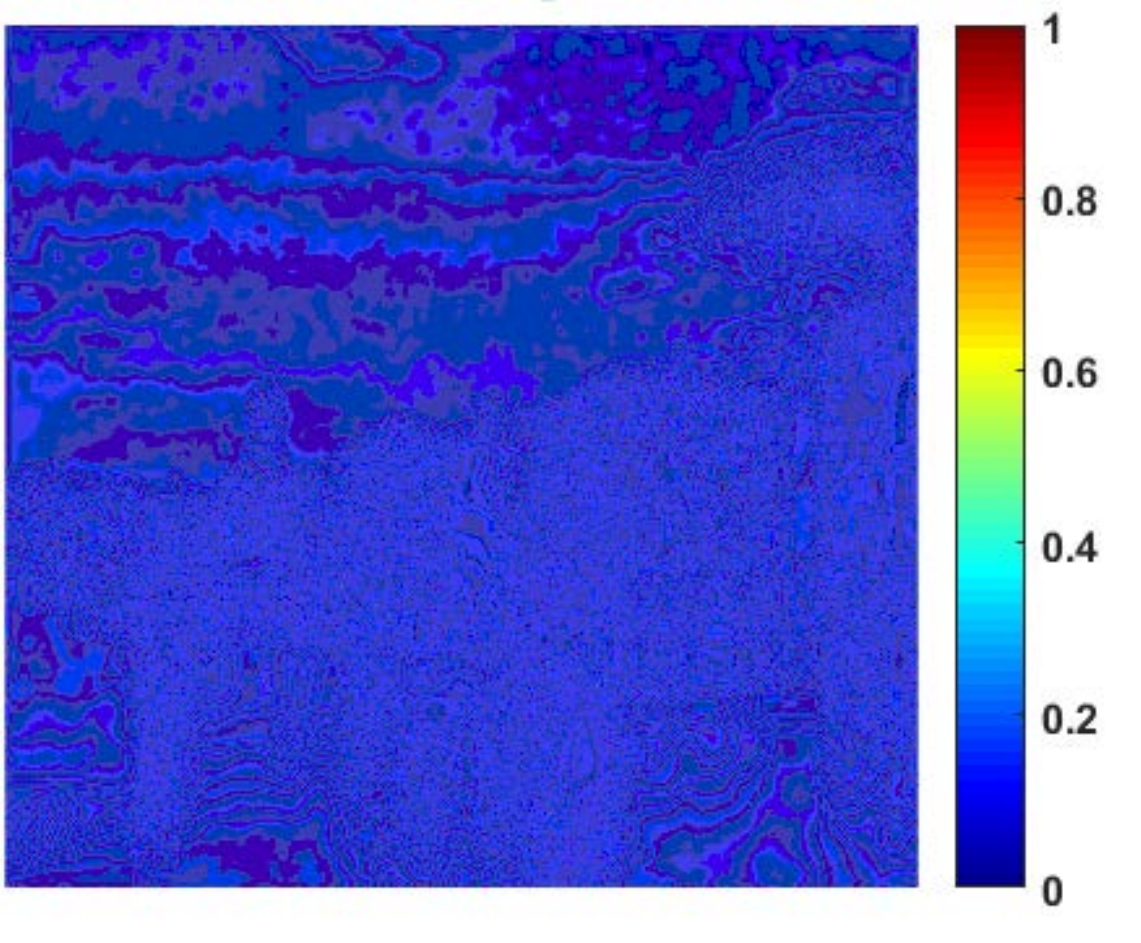}\\
			(a) input& (b) iteration 1 & (c)  iteration 2 &(d)  iteration 8& (e) difference1\_2& (f) difference1\_8\\
		\end{tabular}
	\end{center}
	\caption{The estimated curve parameter maps in different iteration stages. Subfigure (b), (c), and (d) depict the estimated curve parameter maps in iterations 1, 2, and 8, respectively. Subfigure (e) and (f) show the difference maps between iteration 1 and iteration 2 as well as between iteration 1 and iteration 8, respectively. For visualization, we normalize the curve parameter maps and amplify the intensity of the difference maps by 30 times.}
	\label{fig:disparity}
\end{figure*}

\section{Zero-DCE++}
\label{Zero-DCE++}
Though Zero-DCE is already faster and smaller than existing deep learning-based models  \cite{MBLLEN,Chen2018,Jiang2019}, reduced computational cost and faster inference speed are still desired for practical applications, especially when dealing with large images captured by modern mobile devices. We propose an accelerated and light version of Zero-DCE, called Zero-DCE++, to achieve the above-mentioned characteristics.  

To this end, we carefully investigate the relations between enhancement performance and network structure, curve estimation, and input sizes.
We observed that 1) the convolutional layers used in DCE-Net can be replaced with the more efficient depthwise separable convolutions \cite{depthwiseCon} that are commonly used in computer vision tasks~\cite{ChenECCV18,HuCVPR18,LiuCVPR19} for reducing network parameters without compromising the performance much;
2) the estimated curve parameters in different iteration stages (a total of eight iterations in  Zero-DCE) are similar in most cases. In Figure \ref{fig:disparity}, we show an example of the estimated curve parameter maps in different iteration stages and their difference maps. As observed, the curve parameter maps are similar and the values in the difference maps are small. Such results manifest that the curve parameter map can be reused in different iteration stages to handle most cases, thus we can reduce the estimated curve parameter maps from 24 to 3; and 
3) our method is not sensitive to the sizes of input image. 
Consequently, we can use the downsampled input as the input of curve parameter estimation network and then upsample the estimated curve parameter maps back to the original resolutions for image enhancement. The low-resolution input can significantly reduce the computational cost. Based on these observations, we modify the Zero-DCE from three aspects.

First, we re-design the DEC-Net by replacing the convolutional layers with depthwise separable convolutions for reducing the network parameters. Each depthwise separable convolutional layer consists of a depthwise convolution with kernels of size 3$\times$3 and stride 1 and a pointwise convolution with kernels of size 1$\times$1 and stride 1.

Second, we reformulate the curve estimation and only estimate 3 curve parameter maps, then reuse them in different iteration stages instead of estimating 24 parameter maps across eight iterations. Thus, Equation \eqref{equ_3} can be reformulated as
\begin{equation}
\label{equ_9}
LE_{n}(\mathbf{x})=LE_{n-1}(\mathbf{x})+\mathcal{A}(\mathbf{x})LE_{n-1}(\mathbf{x})(1-LE_{n-1}(\mathbf{x})),
\end{equation}
where the same curve parameter map  $\mathcal{A}$ is used to adjust the curves in different iteration stages. Although we reuse the curve parameter maps, it still retains the high-order property thanks to the iteration process.

Third, we can use the downsampled image as the input of our network to estimate the curve parameter maps. By default, we downsample an input by a factor of 12 in Zero-DCE++ to balance the enhancement performance and computational cost. Even with the extreme downsampling factors, our method maintains a good  performance. The reasons are briefly  explained as follows. Firstly, although we adopt the downsampled input to estimate the curve parameters, we resize the small curve parameter maps back to the same size as the original input image based on the assumption that the pixels in a local region have the same intensity (also the same adjustment curves). The  mapping from the input image to the enhanced image is conducted on the original resolution. Secondly, the proposed spatial consistency loss encourages the results to preserve the content of the input image. Thirdly, the adopted losses in our framework are region-wise but not pixel-wise.  We present more discussions and results in the ablation study.

These modifications offer Zero-DCE++ the advantages of having a tiny network (10K trainable parameters, 0.115G FLOPs for an image of size 1200$\times$900$\times$3), real-time inference speed (1000/11 FPS on a single GPU/CPU for an image of size 1200$\times$900$\times$3), and fast training (20 minutes).


\section{Experiments}
\label{sec:Experimental_Results}

\begin{figure*}[!htb]
	\begin{center}
		\begin{tabular}{c@{ }c@{ }c@{ }c@{ }c@{ }c@{ }}
			\includegraphics[width=.16\textwidth,height=3.3cm]{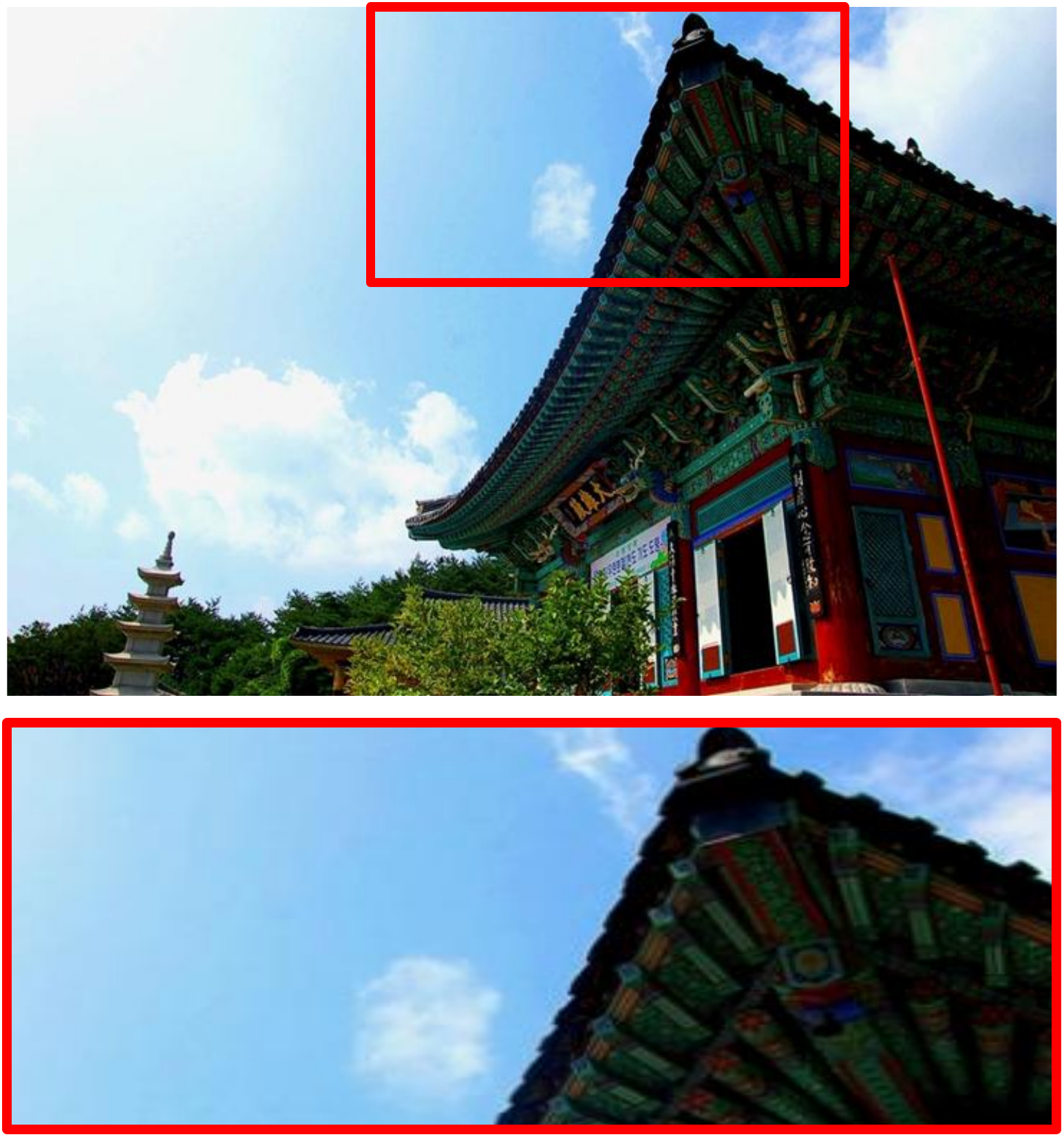}&
			\includegraphics[width=.16\textwidth,height=3.3cm]{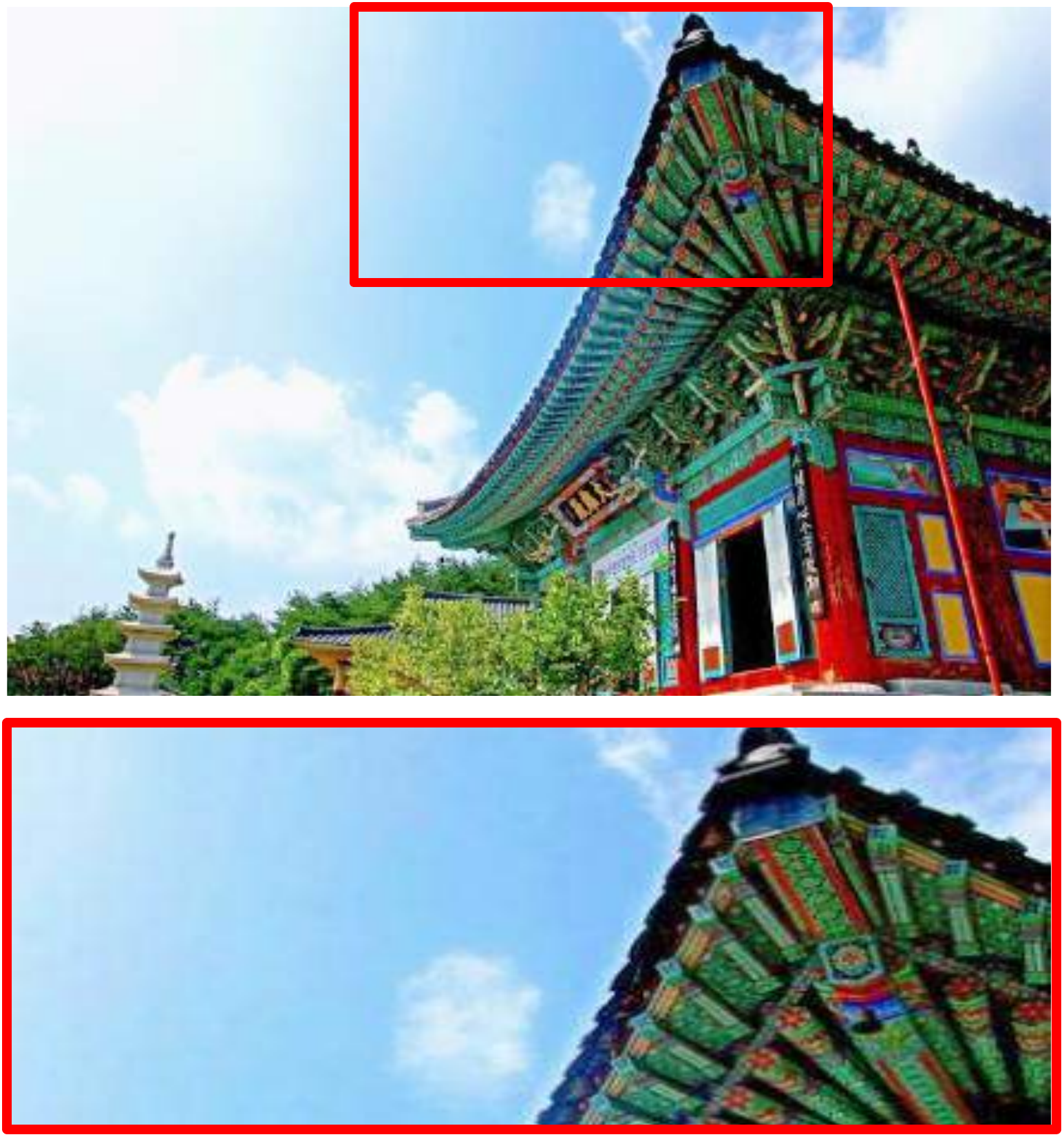}&
			\includegraphics[width=.16\textwidth,height=3.3cm]{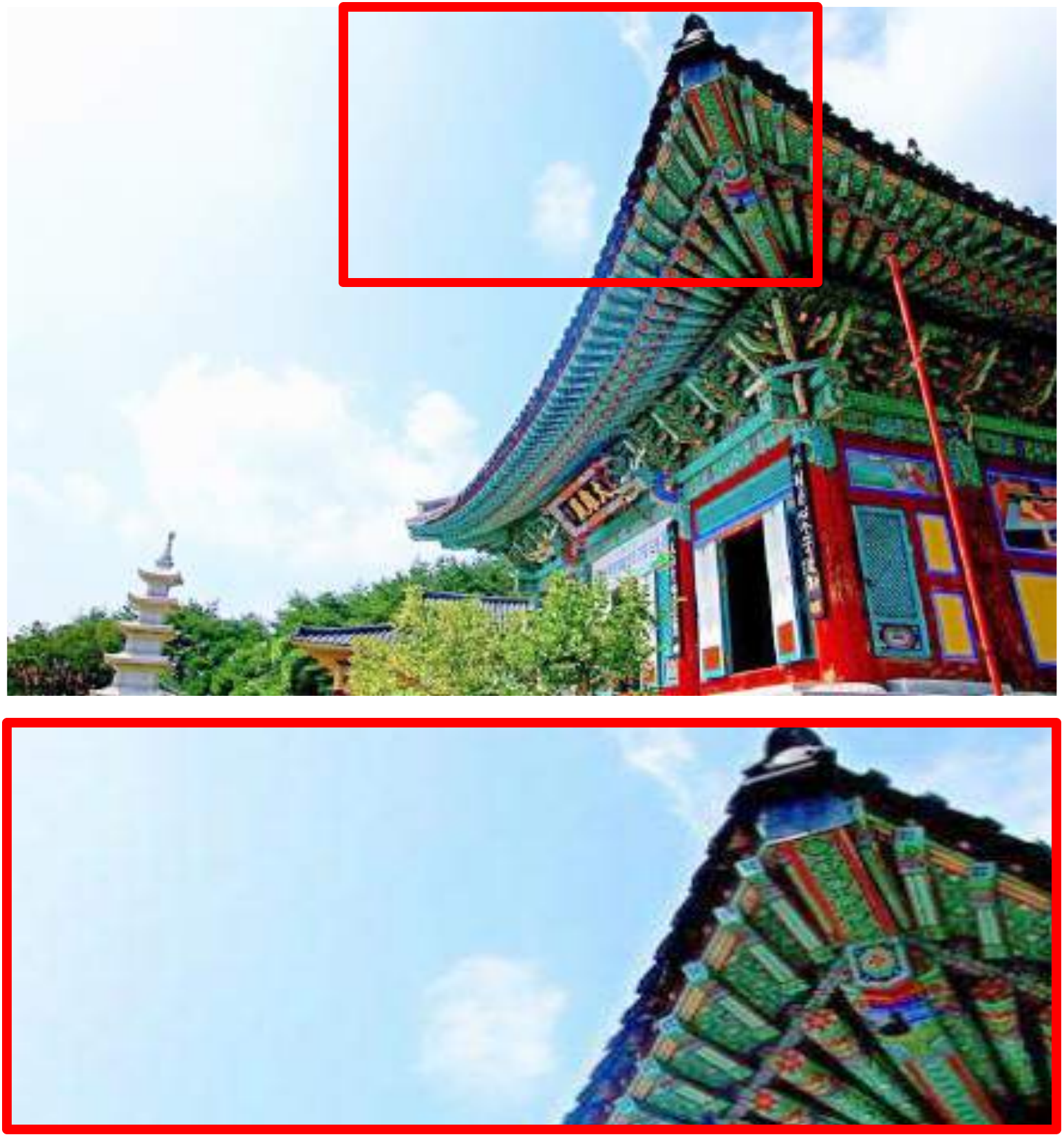}&
			\includegraphics[width=.16\textwidth,height=3.3cm]{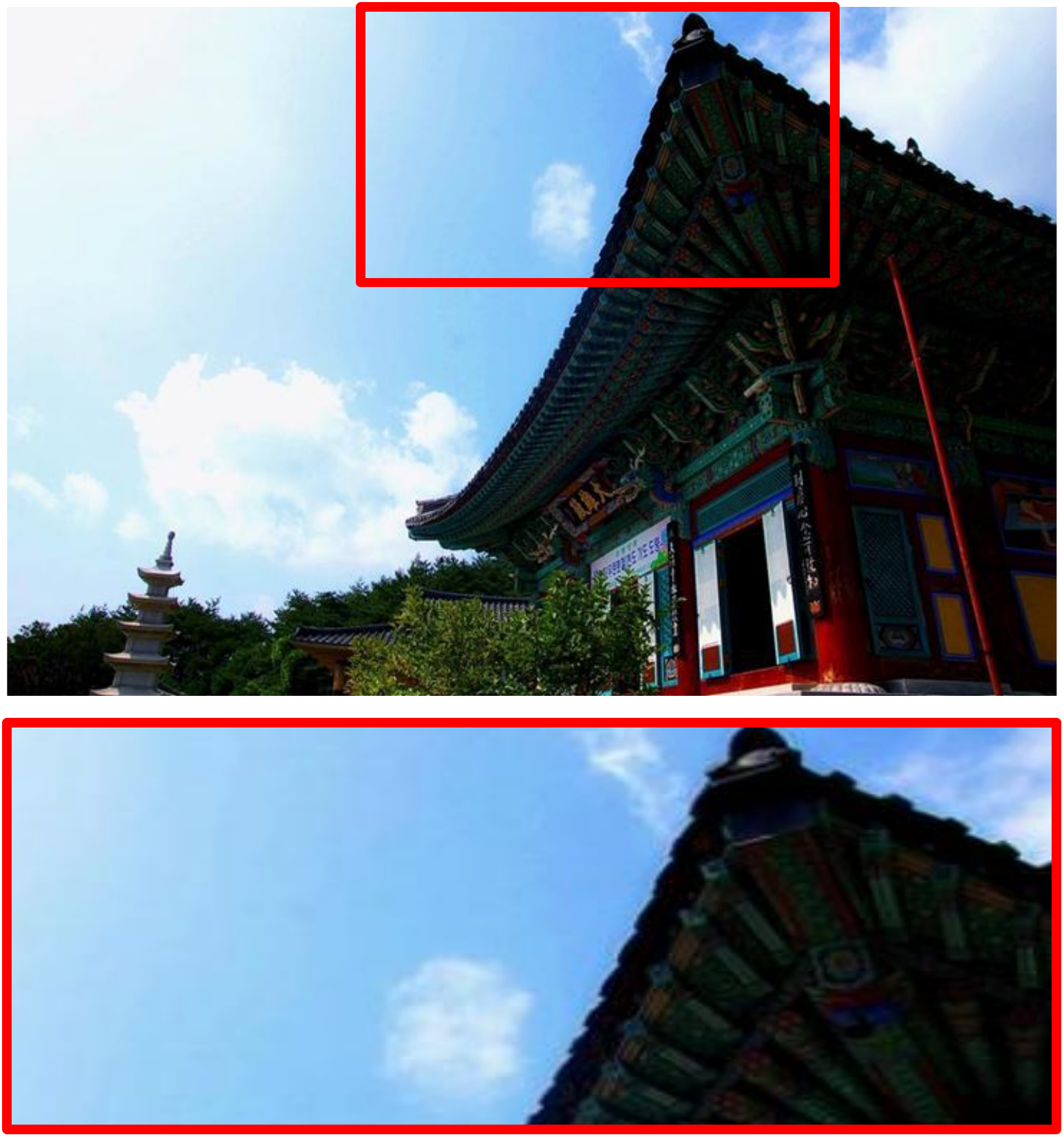}&
			\includegraphics[width=.16\textwidth,height=3.3cm]{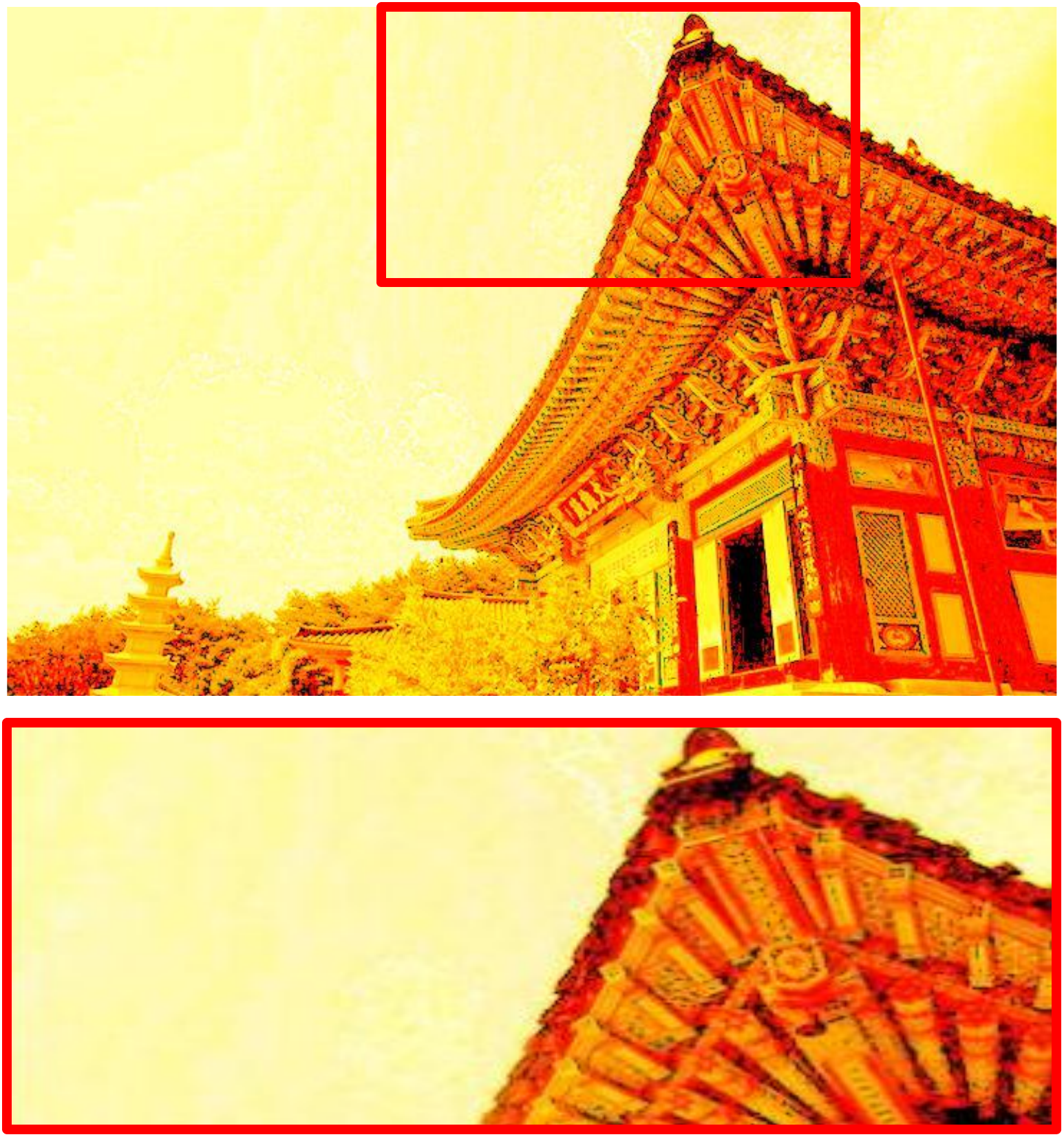}&
			\includegraphics[width=.16\textwidth,height=3.3cm]{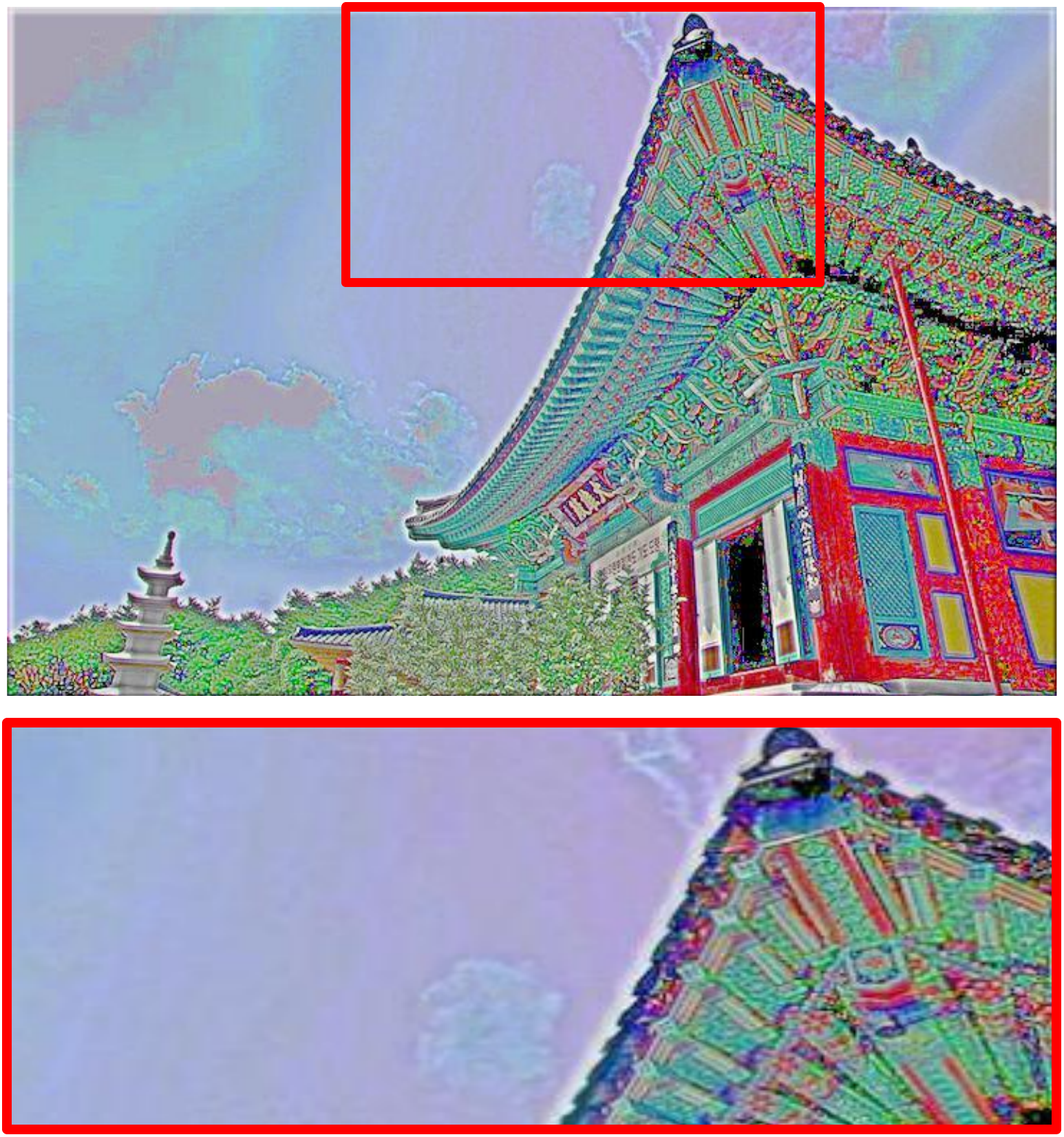}\\
			(a) input& (b) Zero-DCE& (c) w/o $L_{spa}$ &(d) w/o $L_{exp}$& (e) w/o $L_{col}$& (f) w/o $L_{tv_\mathcal{A}}$\\
		\end{tabular}
	\end{center}
	\caption{Ablation study of the contribution of each loss (spatial consistency loss $L_{spa}$, exposure control loss $L_{exp}$, color constancy loss $L_{col}$, illumination smoothness loss $L_{tv_\mathcal{A}}$). Red boxes indicate the obvious differences and amplified details.}
	\label{fig:loss}
\end{figure*}

\subsection{Implementation Details}
\label{Implementation}
CNN-based models usually use self-captured paired data for network training \cite{Chen2018,Chenchen2018} while GAN-based models elaborately select unpaired data \cite{Jiang2019,Chencvpr2018}. To bring the capability of wide dynamic range adjustment into full play, we incorporate both low-light and over-exposed images into our training set. To this end, we employ 360 multi-exposure sequences from the Part1 of SICE dataset~\cite{Cai2018} to train our model. The dataset is also used as a part of the training data in EnlightenGAN \cite{Jiang2019}.
We randomly split 3,022 images of different exposure levels in the Part1 subset~\cite{Cai2018} into two parts (2,422 images for training and the rest for validation).
We resize the training and testing images to the size of 512$\times$512$\times$3.

We implement our framework with MindSpore on an NVIDIA $2080$Ti GPU.
A batch size of $8$ is applied.
The filter weights of each layer are initialized with standard zero mean and 0.02 standard deviation Gaussian function.
Bias is initialized as a constant.
We use ADAM optimizer with default parameters and fixed learning rate $1e^{-4}$ for our network optimization.
The weights $W_{col}$ and $W_{tv_\mathcal{A}}$ are set to 0.5, and 20, respectively, to balance the scale of losses.
Zero-DCE and Zero-DCE++ adopt the same training dataset and configurations during training.

\subsection{Experimental Settings}
\label{settings}
We compare our method with several state-of-the-art methods: three conventional methods (SRIE~\cite{Fu2016}, LIME~\cite{Guo2017}, Li \etal~\cite{Li2018}), four CNN-based methods (Wang \etal~\cite{Wang2019}, RetinexNet~\cite{Chen2018}, LightenNet~\cite{LightenNet}, MBLLEN~\cite{MBLLEN}), and one GAN-based method (EnlightenGAN~\cite{Jiang2019}).
The results are reproduced using publicly available source codes with recommended parameters.

We perform qualitative and quantitative experiments on standard image sets used by previous works including NPE~\cite{Wang2013} (84 images), LIME~\cite{Guo2017} (10 images), MEF~\cite{Ma2015} (17 images), DICM~\cite{Lee2012} (64 images), and VV\footnote[3]{\url{https://sites.google.com/site/vonikakis/datasets}}  (24 images).
Besides, we quantitatively validate our method on the Part2 subset of SICE dataset~\cite{Cai2018}, which consists of 229 multi-exposure sequences and the corresponding reference image for each multi-exposure sequence.
For a fair comparison, we only use the low-light images of Part2 subset~\cite{Cai2018} for testing, since baselines cannot handle over-exposed images well.
Specifically, we choose the first three (resp. four) low-light images if there are seven (resp. nine) images in a multi-exposure sequence and resize all images to a size of 1200$\times$900$\times$3.
Finally, we obtain 767 paired low/normal light images, denoted as Part2 testing set.

The low/normal light image dataset mentioned in~\cite{2019arXiv190404474Y} was discarded because the training datasets of RetinexNet~\cite{Chen2018} and EnlightenGAN~\cite{Jiang2019} consist of some images from this dataset. We did not use the MIT-Adobe FiveK dataset \cite{Adobe5K} as it is not primarily designed for underexposed photos enhancement and still contains some low-light images in the ground truth set. Note that this paper only focuses on low-light image enhancement on RGB images, thus we did not include the methods that require raw data as inputs and were  designed for general photo enhancement.

\subsection{Ablation Study}

We perform ablation studies to demonstrate the effectiveness of each component of Zero-DCE. Additionally, the comparisons between Zero-DCE and Zero-DCE++ are carried out to analyze the advantages and disadvantages of the accelerated and light version at the end of this section.

\noindent
\textbf{Contribution of Each Loss.}
We present the results of Zero-DCE trained by various combinations of losses in Figure~\ref{fig:loss}.
The result without spatial consistency loss $L_{spa}$ has relatively lower contrast (\eg, the cloud regions) than the full result. This shows the importance of $L_{spa}$ in preserving the difference of neighboring regions between the input and the enhanced image.
Removing the exposure control loss $L_{exp}$ fails to recover the low-light region.
Severe color casts emerge when the color constancy loss $L_{col}$ is discarded. This variant ignores the relations among three channels when curve mapping is applied.
Finally, removing the illumination smoothness loss $L_{tv_\mathcal{A}}$ hampers the correlations between neighboring regions leading to obvious artifacts.
Such results demonstrate that each loss used in our zero-reference learning framework plays a significant role in achieving the final visually pleasing results.

\begin{figure}[t]
	\begin{center}
		\begin{tabular}{c@{ }c@{ }c@{ }c@{ }c}
			\includegraphics[width=.11\textwidth,height=2.6cm]{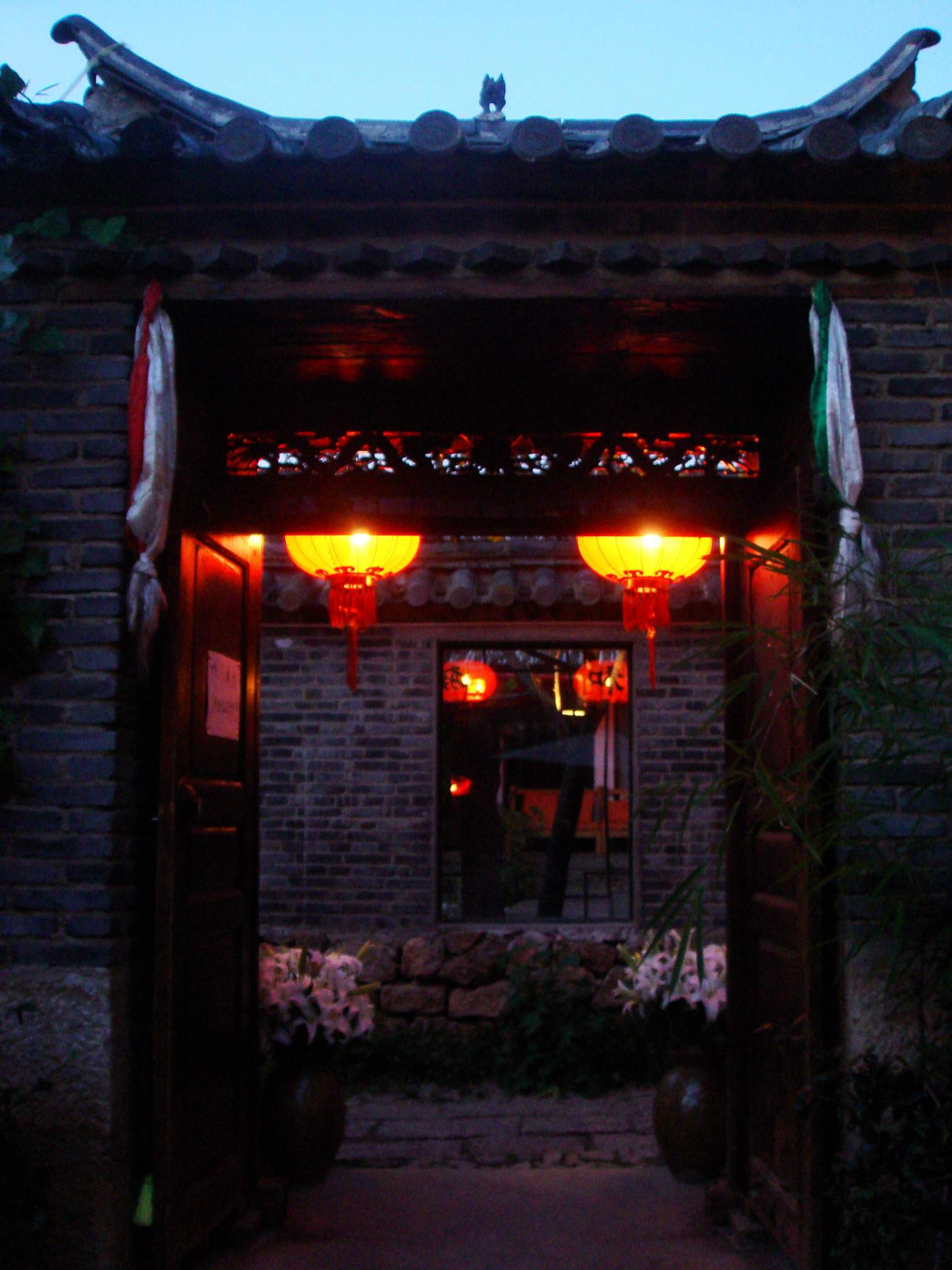}&
			\includegraphics[width=.11\textwidth,height=2.6cm]{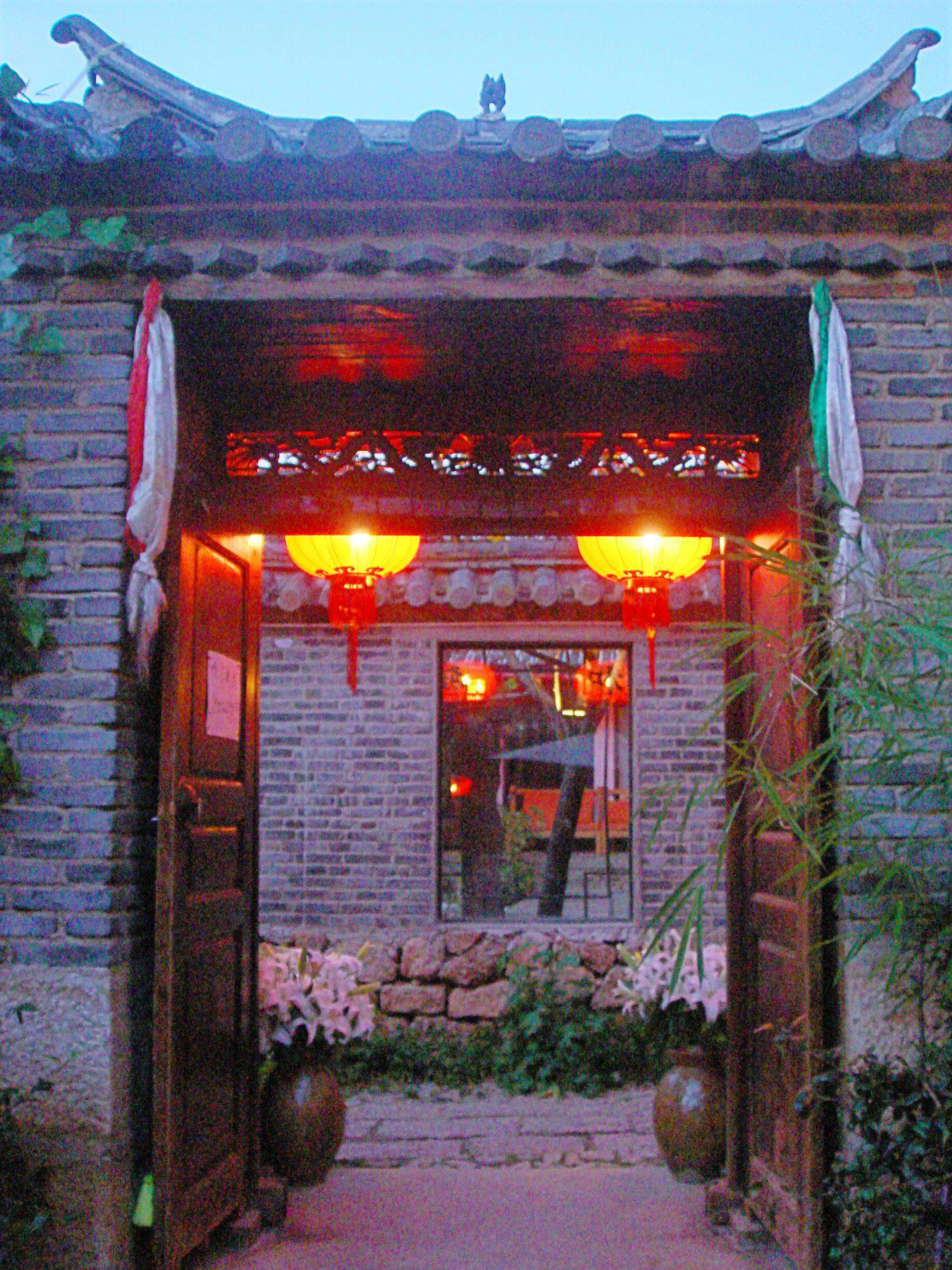}&
			\includegraphics[width=.11\textwidth,height=2.6cm]{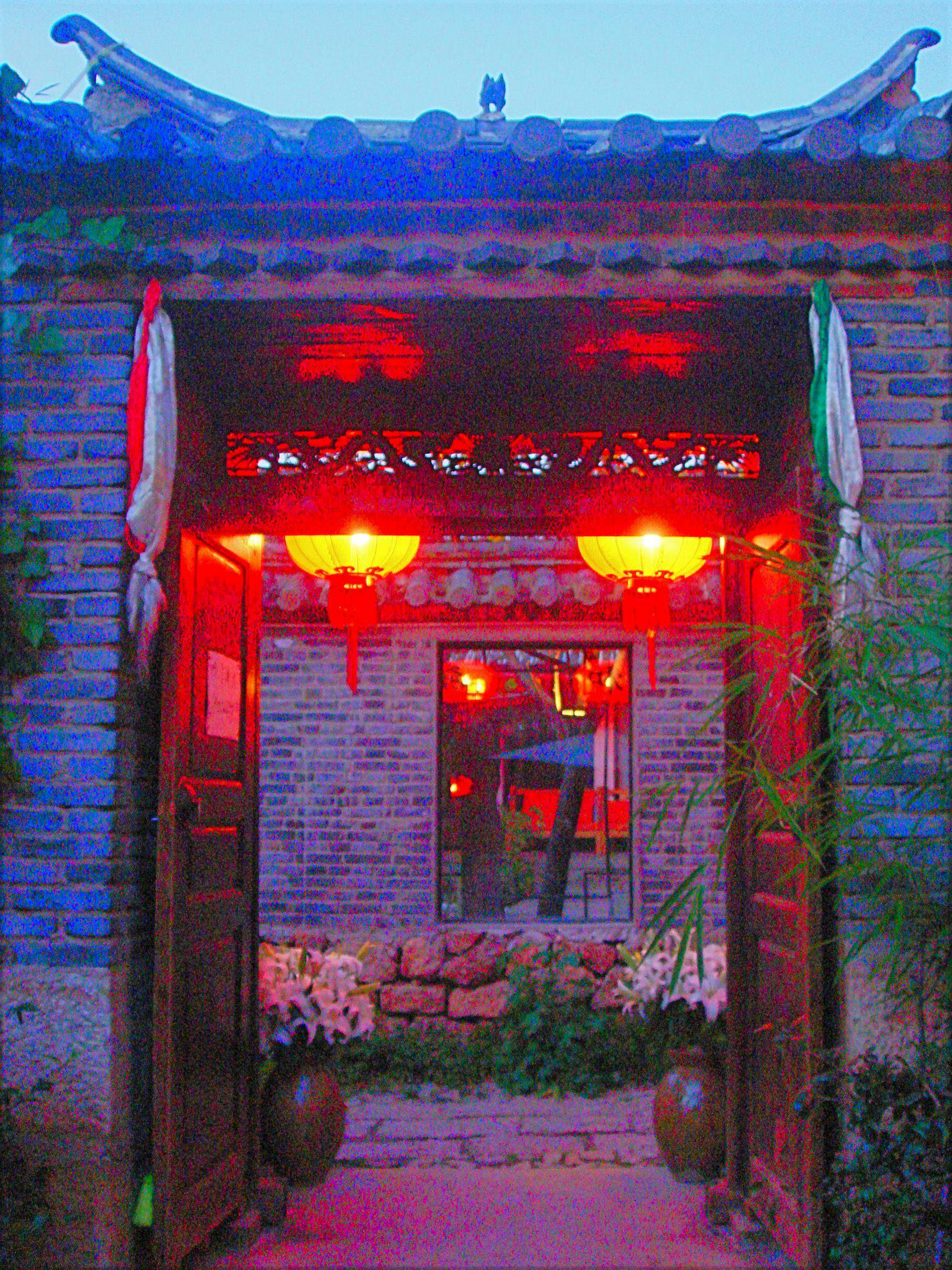}&
			\includegraphics[width=.11\textwidth,height=2.6cm]{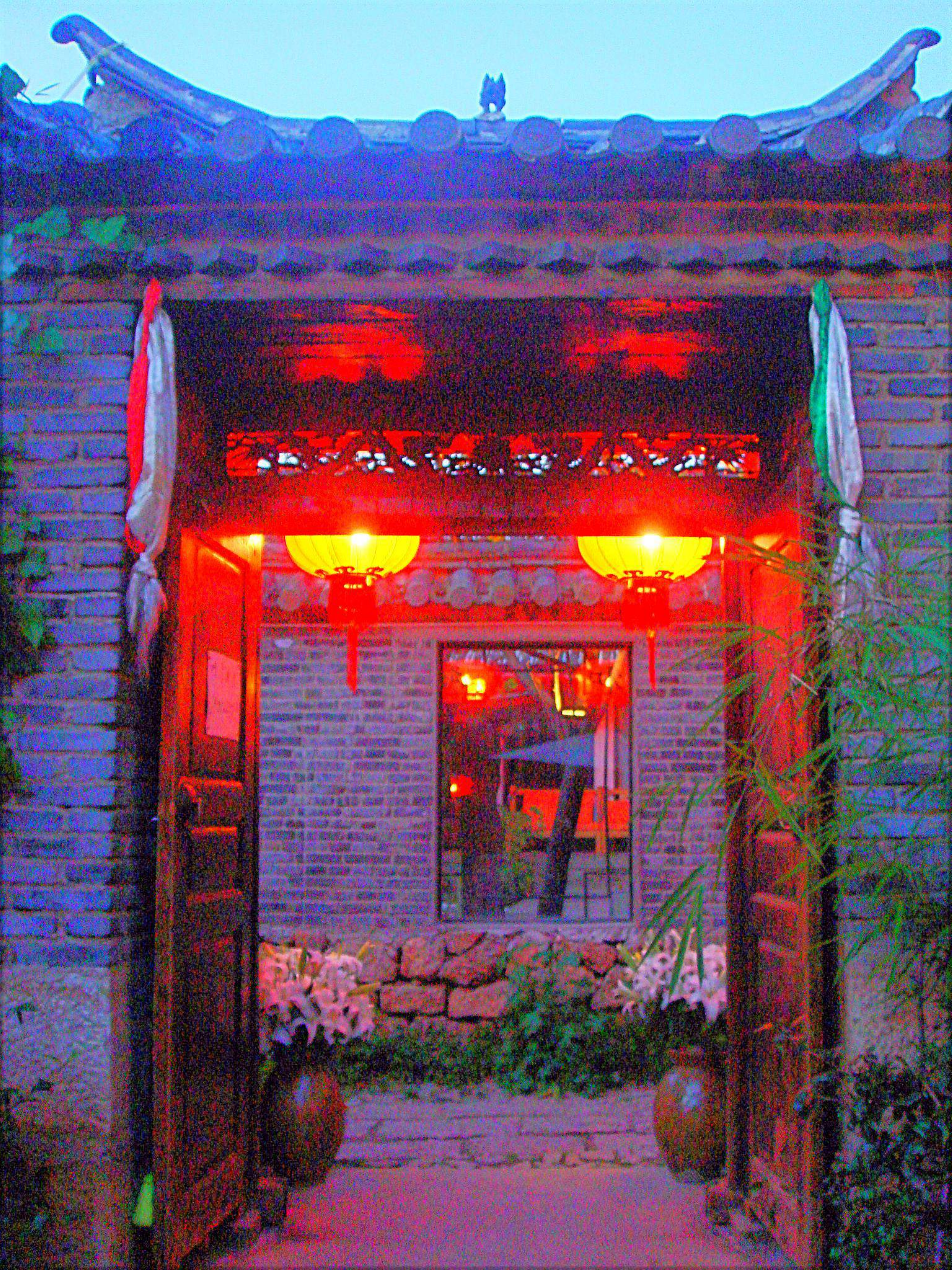}&\\
			(a) input& (b) RGB& (c) CIE Lab& (d) YCbCr\\
		\end{tabular}
	\end{center}
	\caption{Ablation study of the advantage of three channels adjustment (RGB, CIE Lab, and YCbCr color spaces).}
	\label{fig:colorspace}
\end{figure}

\begin{figure*}
	\begin{center}
		\begin{tabular}{c@{ }c@{ }c@{ }c@{ }c@{ }c@{ }c}
			\includegraphics[width=.16\textwidth,height=2.2cm]{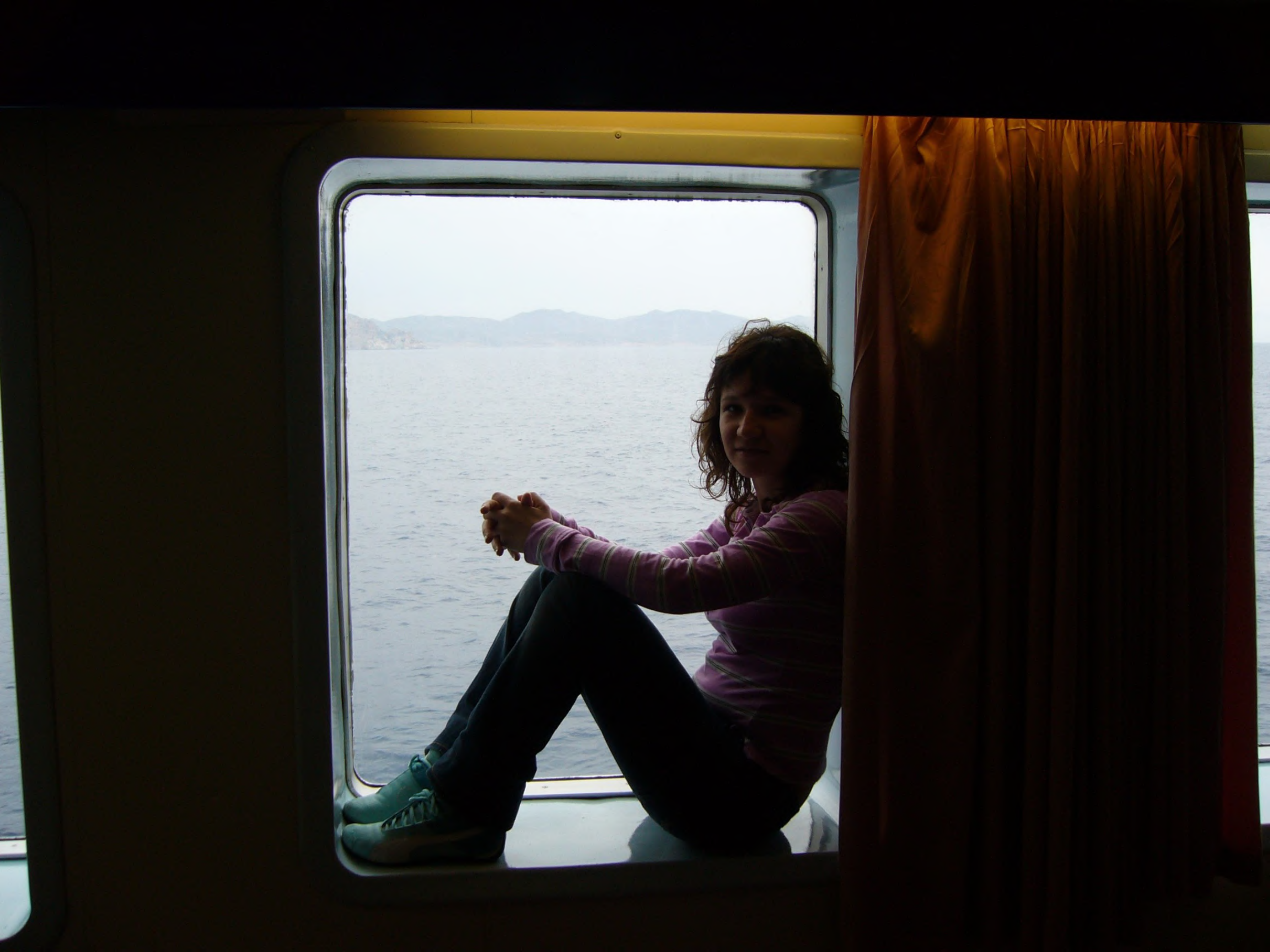}&
			\includegraphics[width=.16\textwidth,height=2.2cm]{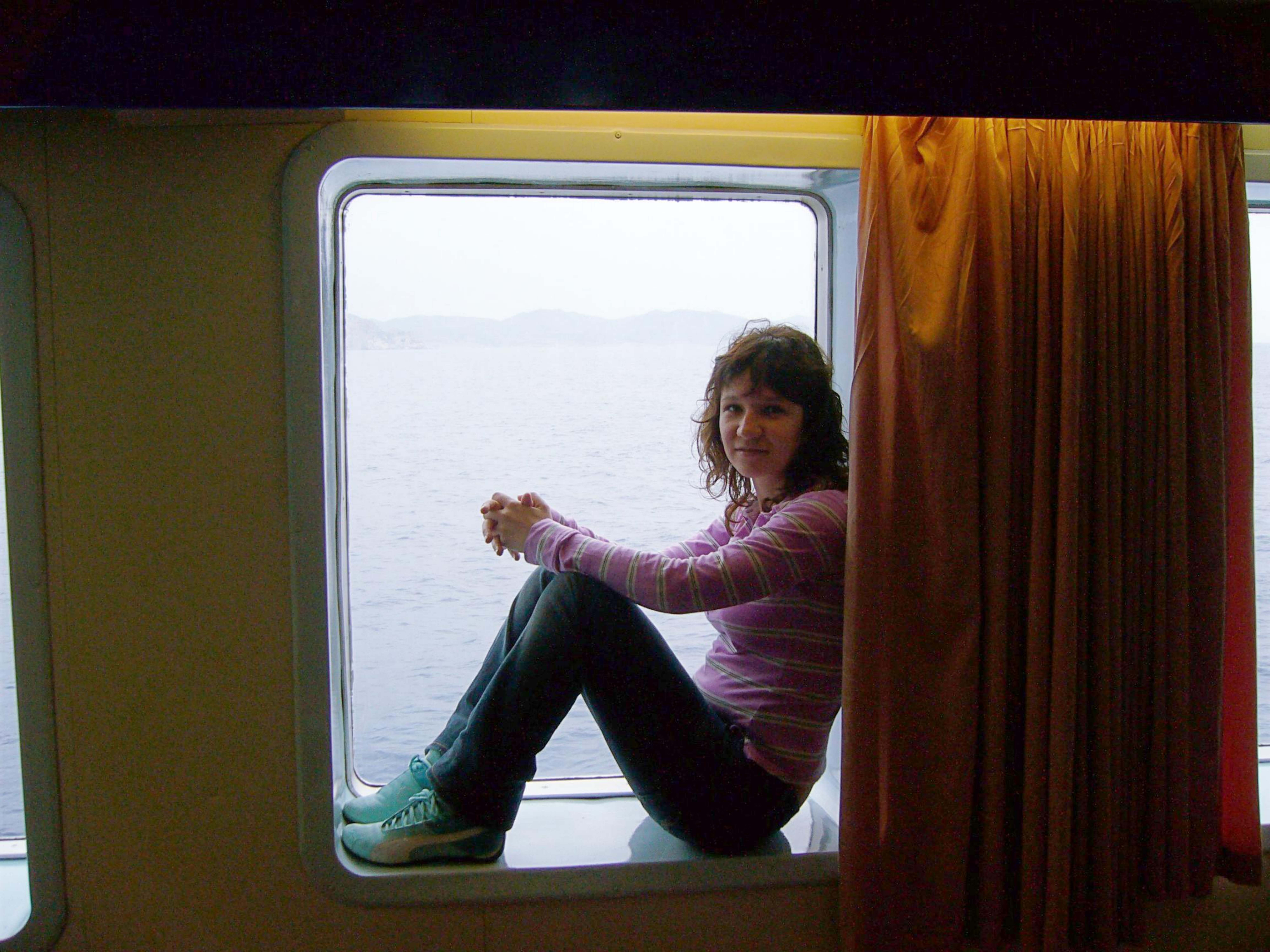}&
			\includegraphics[width=.16\textwidth,height=2.2cm]{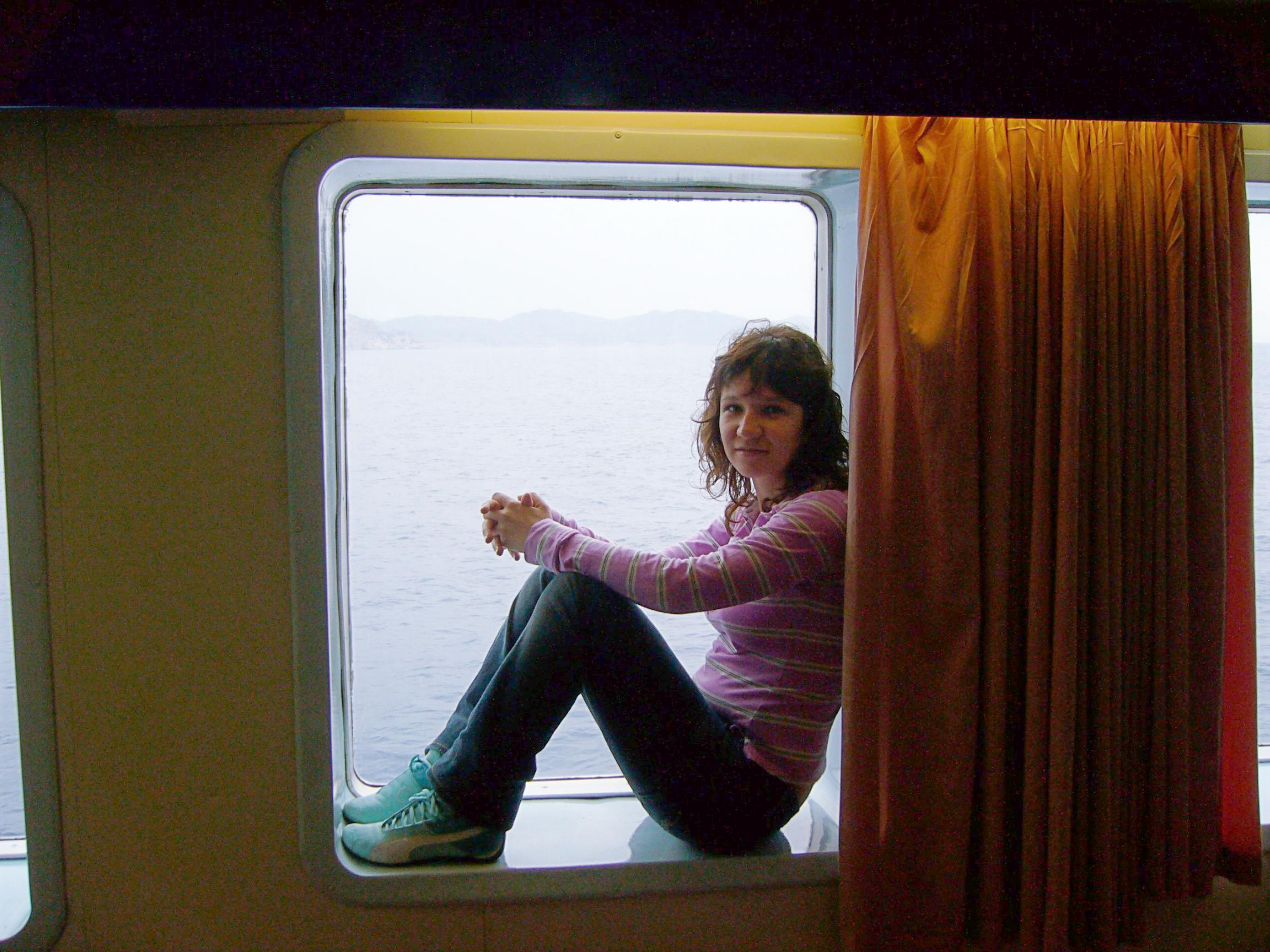}&
			\includegraphics[width=.16\textwidth,height=2.2cm]{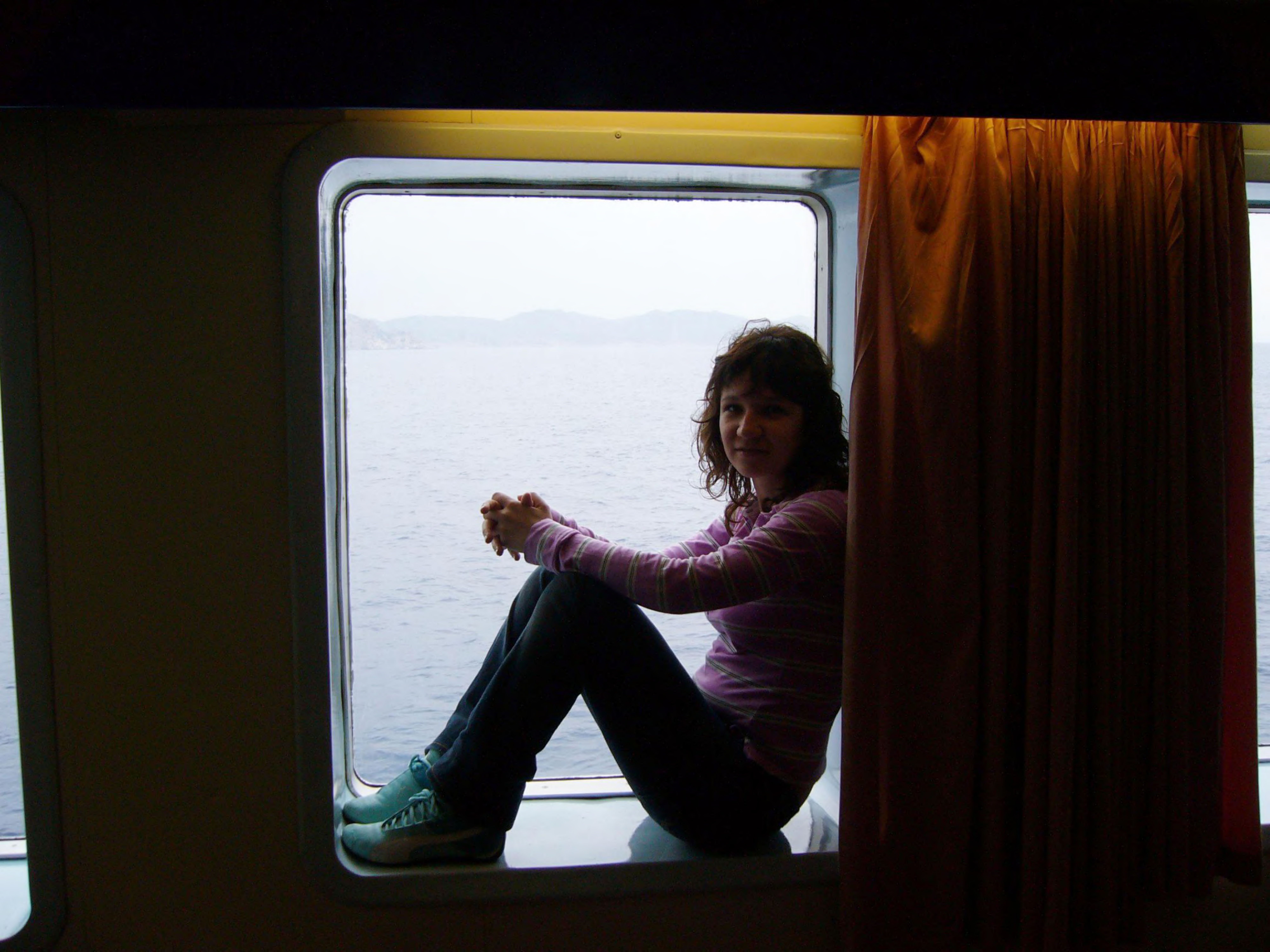}&
			\includegraphics[width=.16\textwidth,height=2.2cm]{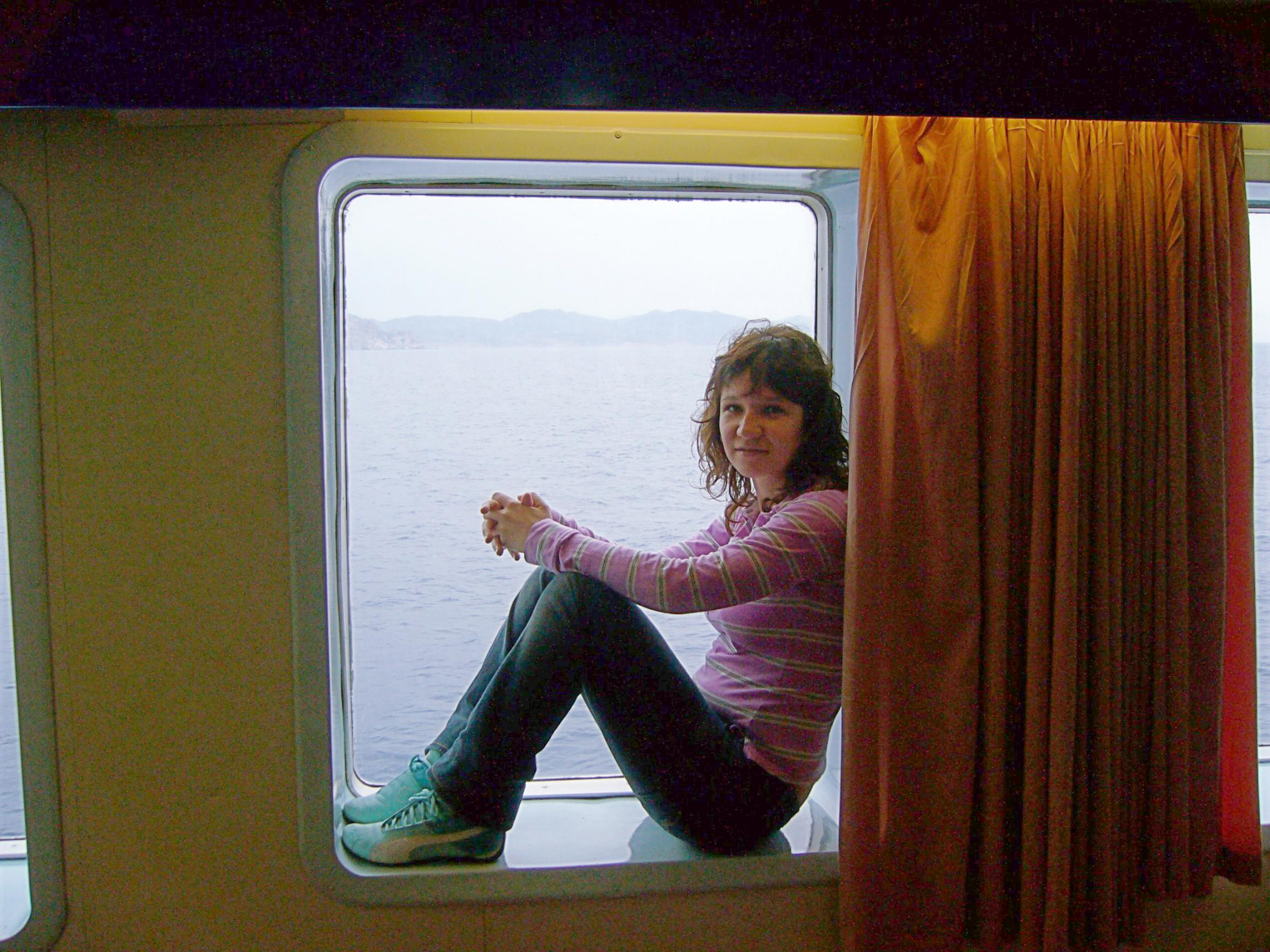}&
			\includegraphics[width=.16\textwidth,height=2.2cm]{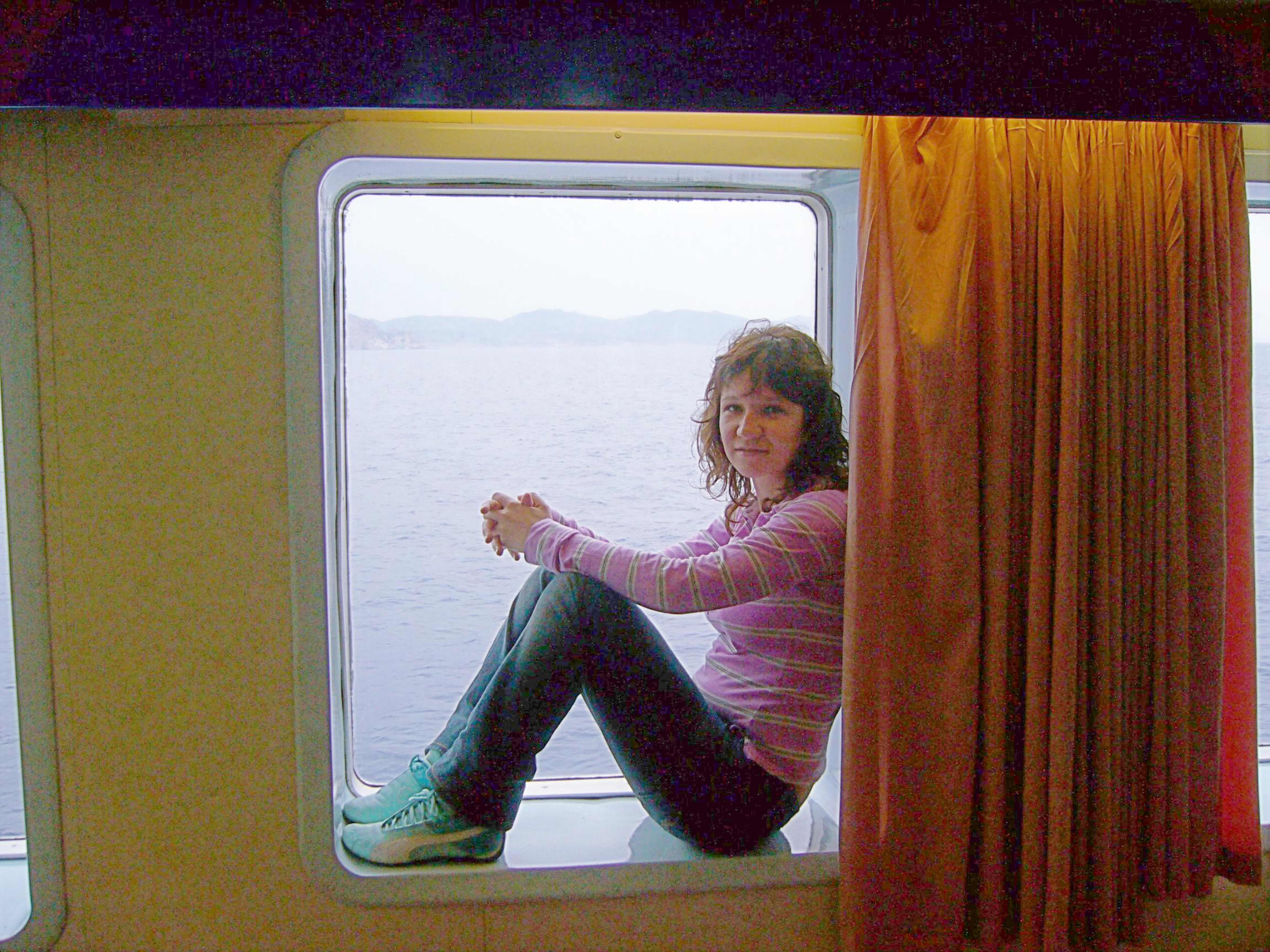}\\
			(a) input& (b) l3-f32-n8& (c) l7-f16-n8 &	(d) l7-f32-n1& (e) l7-f32-n8 & (f) l7-f32-n16\\
		\end{tabular}
	\end{center}
	\caption{Ablation study of the effect of parameter settings. $l$-$f$-$n$ represents the proposed Zero-DCE with $l$ convolutional layers, $f$ feature maps of each layer (except the last layer), and $n$ iterations.}
	\label{fig:parameter}
\end{figure*}

\begin{figure*}[t]
	\begin{center}
		\begin{tabular}{c@{ }c@{ }c@{ }c@{ }c@{ }c}
			\includegraphics[width=.18\textwidth,height=3.7cm]{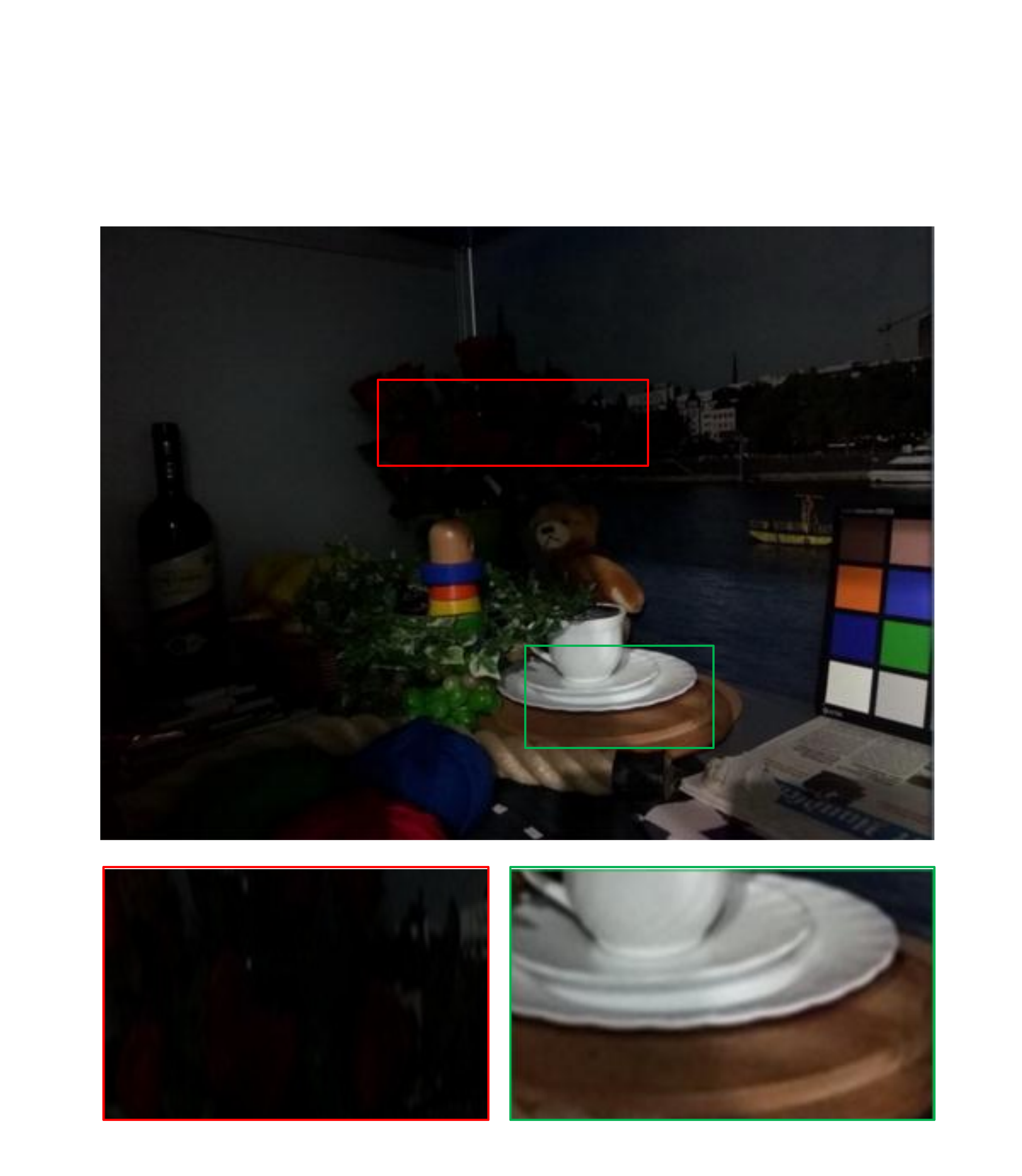}~~~&
			\includegraphics[width=.18\textwidth,height=3.7cm]{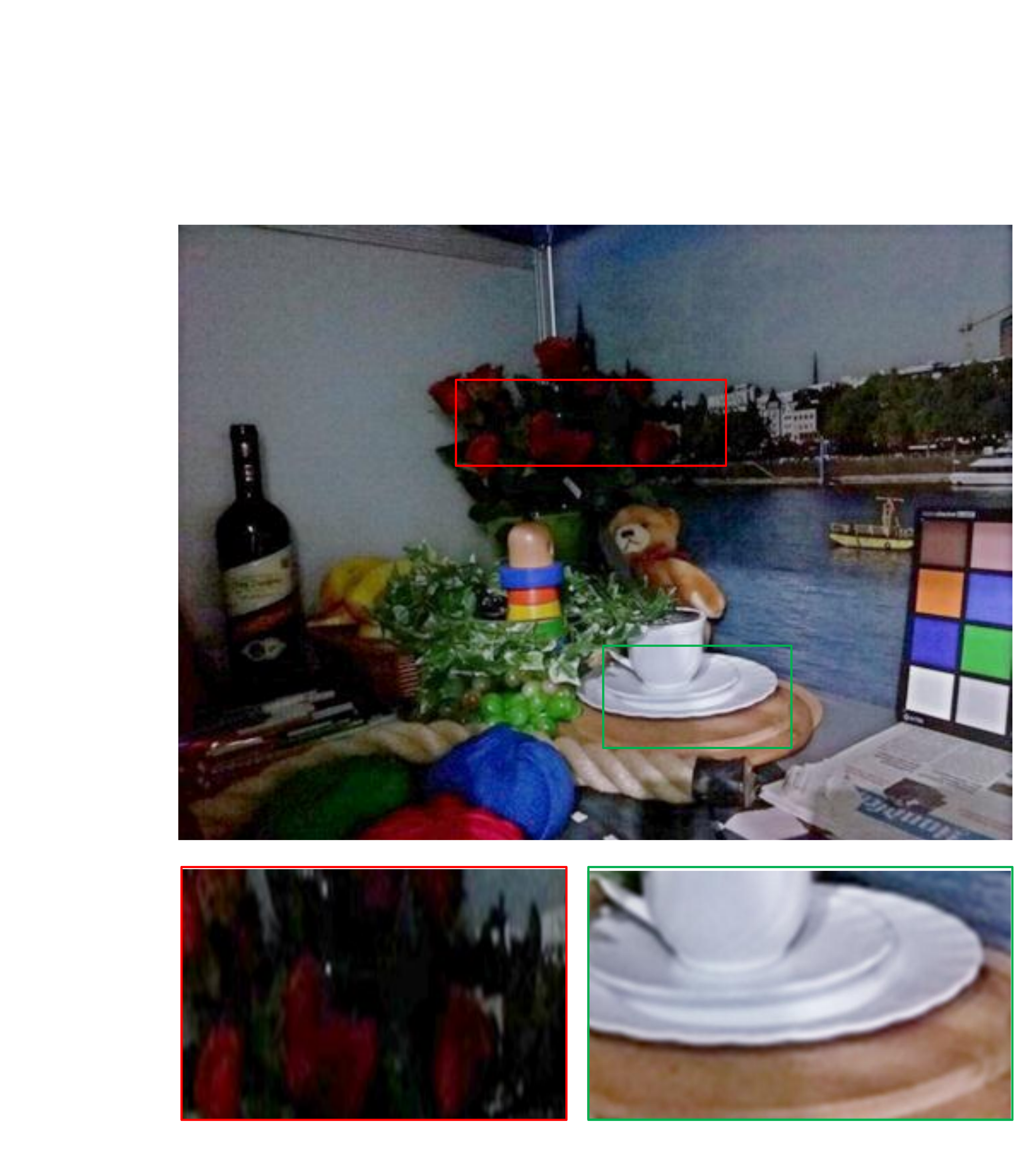}~~~&
			\includegraphics[width=.18\textwidth,height=3.7cm]{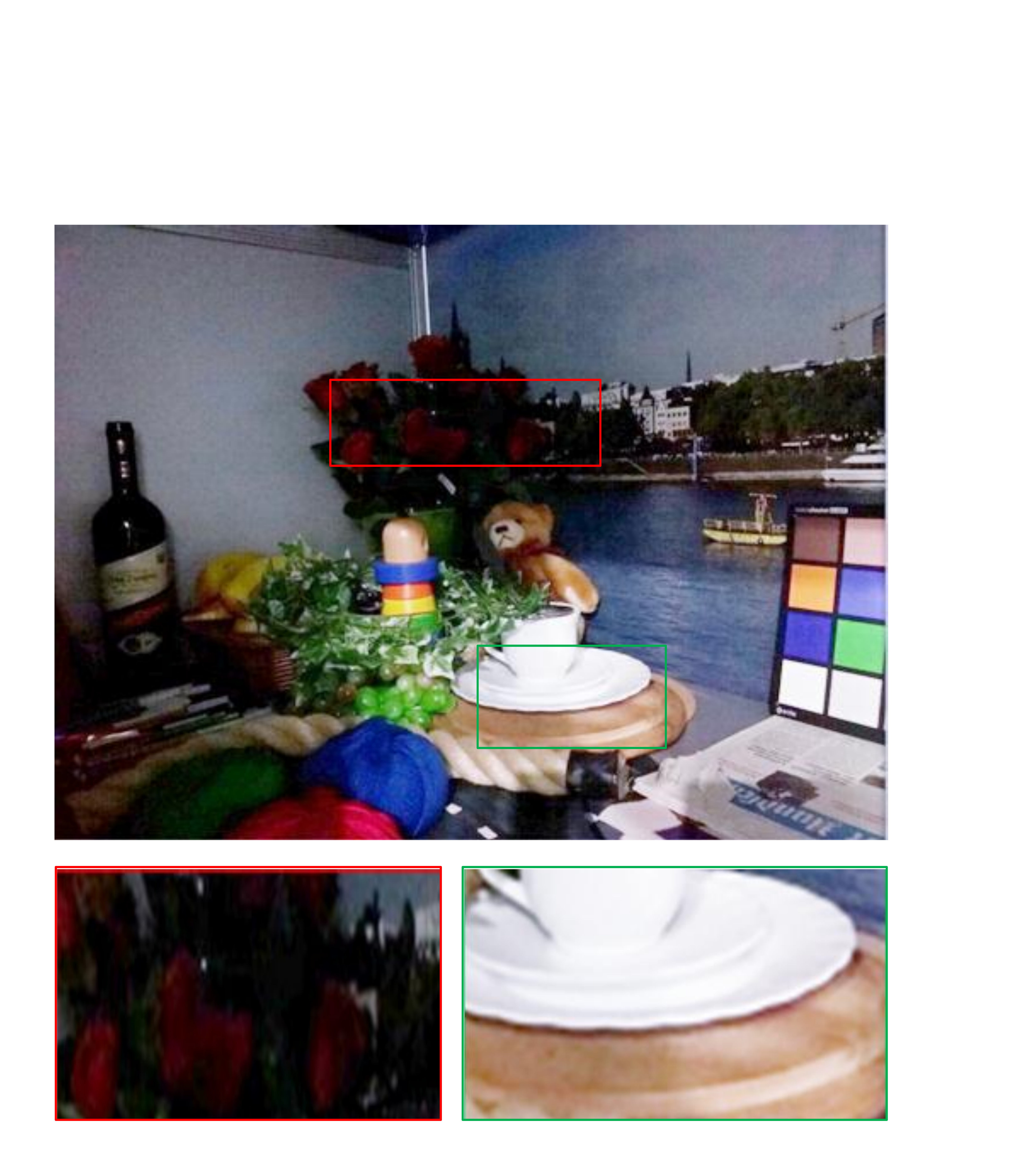}~~~&
			\includegraphics[width=.18\textwidth,height=3.7cm]{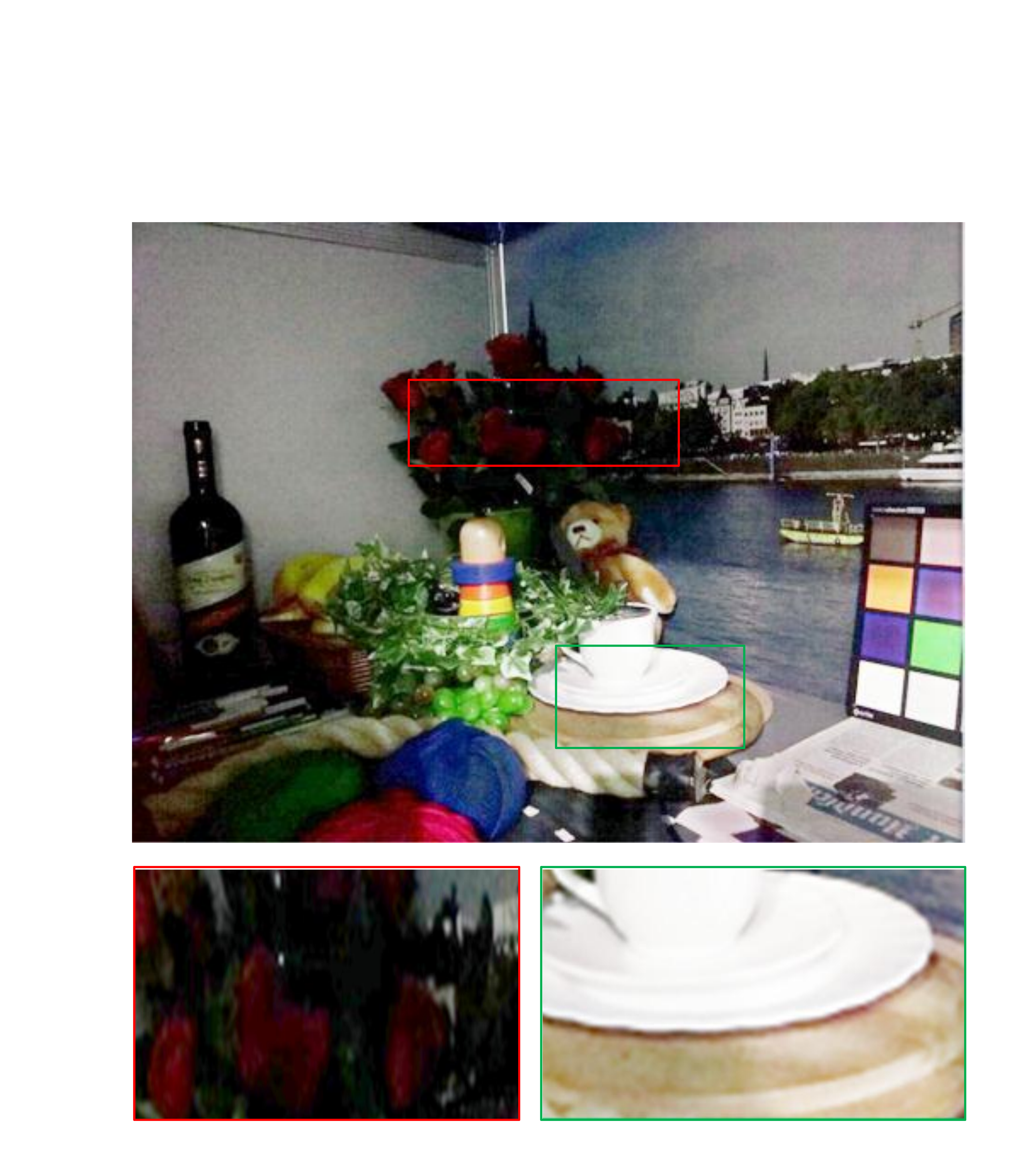}~~~&
			\includegraphics[width=.18\textwidth,height=3.7cm]{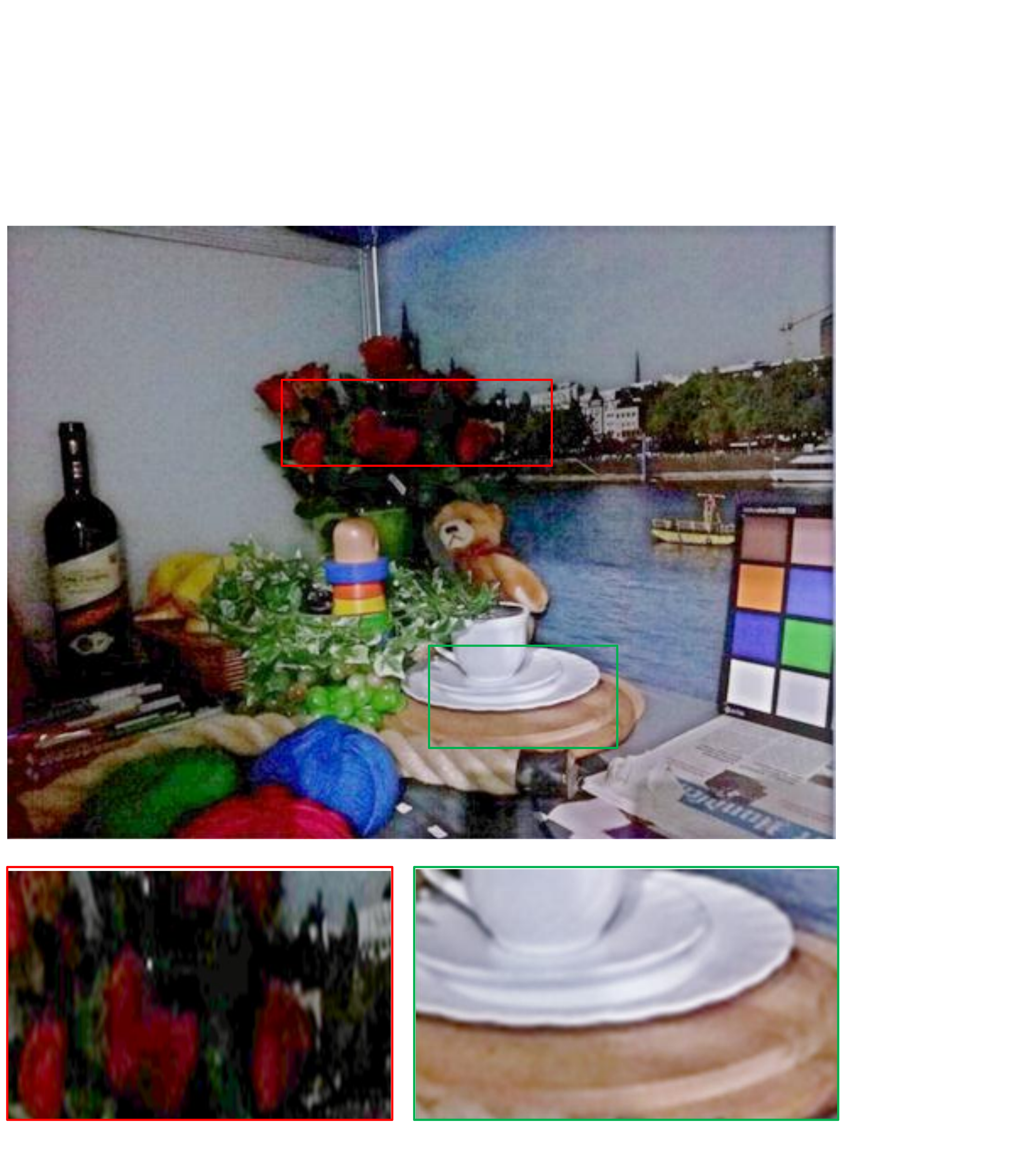}\\
			(a) input~& (b) Zero-DCE~& (c) Zero-DCE$_{Low}$~& (d) Zero-DCE$_{LargeL}$~& (e) Zero-DCE$_{LargeLH}$\\
		\end{tabular}
	\end{center}
	\caption{Ablation study of the impact of training data. Zero-DCE$_{Low}$ represents that the Zero-DCE was trained on only 900 low-light images out of 2,422 images in the original training set.   Zero-DCE$_{LargeL}$ represents that the Zero-DCE was trained on 9,000 unlabeled low-light images provided in the DARK FACE dataset~\cite{2019arXiv190404474Y}. Zero-DCE$_{LargeLH}$ represents that the Zero-DCE was trained on 4800 multi-exposure images from the data augmented combination of Part1 and Part2 subsets in the SICE dataset \cite{Cai2018}. (b) suggests that Zero-DCE has a good balance between over-enhancement and under-enhancement.}
	\label{fig:trainingdata}
\end{figure*}

\noindent
\textbf{Advantage of Three Channels Adjustment.}
To demonstrate the advantage of three channels adjustment, we try to adjust the illumination related channel only in CIE Lab and YCbCr color spaces using the same configurations as the adjustment in RGB color space, except removing the color constancy loss which is only available for three channels adjustment.

Specifically, we first transfer the input from RGB color space to CIE Lab (YCbCr) color space, then feed the L (Y) component to the DCE-Net for estimating a set of curve parameter maps, where we compute each loss in L (Y) channel in the phase of training.
At last, we adjust the L (Y) component using Equation~\ref{equ_3} with the estimated curve parameters.
After the adjustment of L (Y) component, the corresponding ab (CbCr) components are adjusted accordingly (equal proportion adjustment).
In Figure~\ref{fig:colorspace}, we show an example to demonstrate the advantage of three channels adjustment.
As observed, all the results show improved brightness and contrast, suggesting the effectiveness of both the single channel adjustment (CIE Lab and YCbCr color spaces) and the three-channel adjustment  (RGB color space) in improving the brightness of the given low-light image. However, the results adjusted in CIE Lab and YCbCr color spaces as shown in Figure~\ref{fig:colorspace}(c) and (d) have obvious color deviations (\eg, the color of wall) and over-saturation (\eg, the region of lantern).
The visual comparison suggests that three channels adjustment can better preserve the inherent color and reduce the risk of over-saturation.

\noindent
\textbf{Effect of Parameter Settings.} We evaluate the effect of parameters in Zero-DCE, consisting of the depth and width of the DCE-Net and the number of iterations.
A visual example is presented in Figure~\ref{fig:parameter}.
As observed in Figure~\ref{fig:parameter}(b), with just three convolutional layers, Zero-DCE$_{l3-f32-n8}$ can already produce satisfactory results, suggesting the effectiveness of zero-reference learning.
The Zero-DCE$_{l7-f32-n8}$ and Zero-DCE$_{l7-f32-n16}$ produce the most visually pleasing results with natural exposure and proper contrast.
By reducing the number of iterations to 1, an obvious decrease in performance is observed on Zero-DCE$_{l7-f32-n1}$ as shown in Figure~\ref{fig:parameter}(d).
This is because the curve with only single iteration has limited adjustment capability. This suggests the need for higher-order curves in our method.

The same tendency also can be found in the quantitative comparisons  in Table \ref{table_11}.  The comparison between the input and the enhanced results by Zero-DCE$_{l3-f32-n8}$ suggests the effectiveness of the proposed method despite the network only contains three convolutional layers.
The Zero-DCE$_{l7-f32-n1}$ achieves the worst quantitative performance due to the limited adjustment capability of only one-time iteration (\ie, n=1), suggesting the importance of using more iterations. When we increase the number of feature maps of each layer from 16 to 32, the quantitative performance is improved (\ie, Zero-DCE$_{l7-f16-n8}$ and Zero-DCE$_{l7-f32-n8}$). Increasing the number of iterations from 8 to 16 only boosts the average PSNR value marginally (\ie, Zero-DCE$_{l7-f32-n8}$ and Zero-DCE$_{l7-f32-n16}$).
Consequently, we choose Zero-DCE$_{f7-l32-n8}$ as the final model based on its good trade-off between efficiency and restoration performance.
\begin{table}[!t]
	\caption{Quantitative comparisons in terms of Peak Signal-to-Noise Ratio (PSNR, dB), Structural Similarity (SSIM)~\cite{SSIM}, and Mean Absolute Error (MAE). These comparisons are carried out on Part2 testing set. $l$-$f$-$n$ represents the proposed Zero-DCE with $l$ convolutional layers, $f$ feature maps of each layer (except the last layer), and $n$ iterations.}
	\centering
	\begin{tabular}{c|c|c|c}
		\hline
		\textbf{Method} & \textbf{PSNR$\uparrow$ } & \textbf{SSIM$\uparrow$}  & \textbf{MAE$\downarrow$} \\
		\hline
		input                 &   10.71    & 0.33       &209.65 \\
		l3-f32-n8   &  14.50     & 0.56       &119.40\\
		l7-f16-n8   &   15.67    & 0.58       &111.01  \\
		l7-f32-n1   &   12.21    &0.42  &172.89 \\
		l7-f32-n8    &  16.57 &0.59   &98.78 \\
		l7-f32-n16  &  16.79      & 0.57 & 98.70\\
		\hline
	\end{tabular}
	\label{table_11}
\end{table}

\begin{figure*}
	\begin{center}
		\begin{tabular}{c@{ }c@{ }c@{ }c@{ }c@{ }}
			\includegraphics[width=.16\textwidth,height=2.2cm]{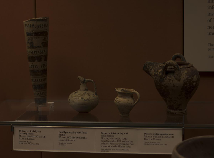}&
			\includegraphics[width=.16\textwidth,height=2.2cm]{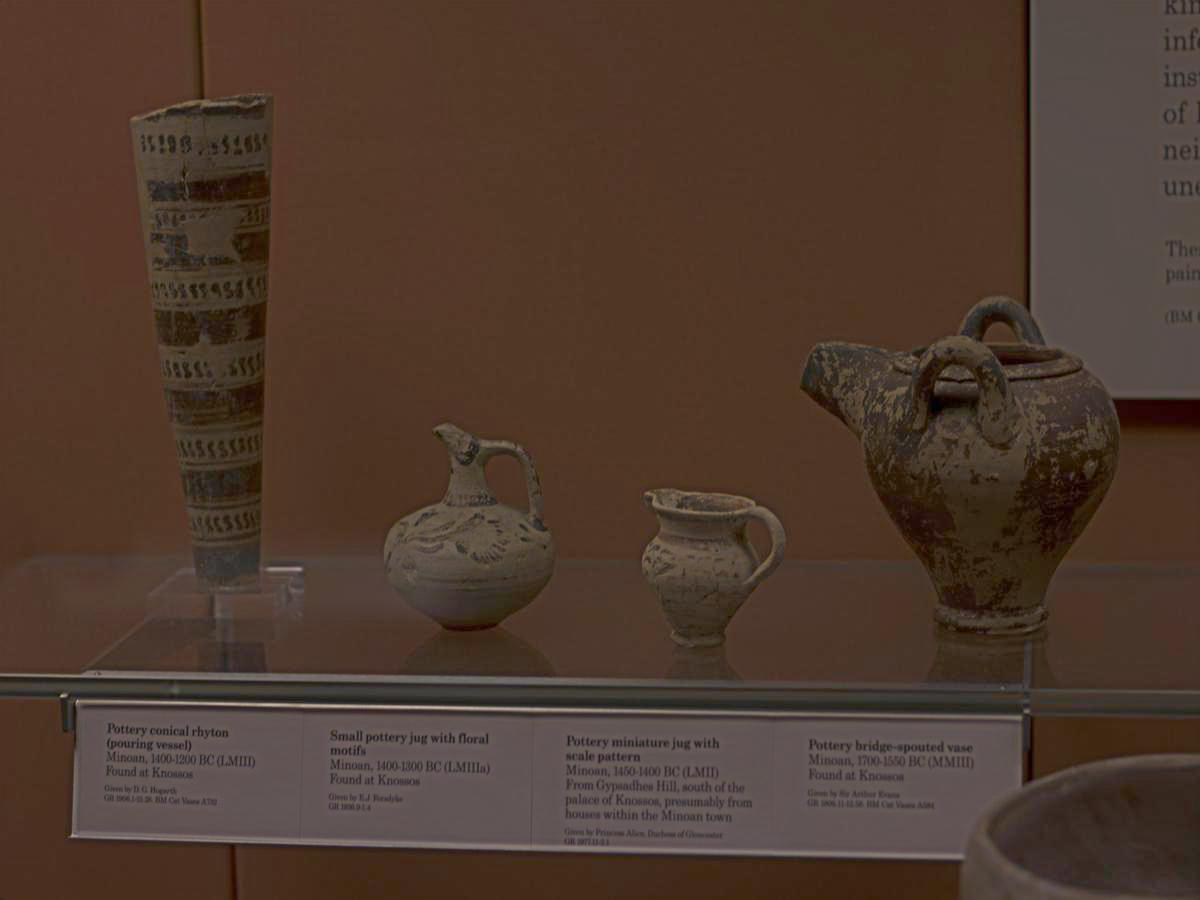} &	
			\includegraphics[width=.16\textwidth,height=2.2cm]{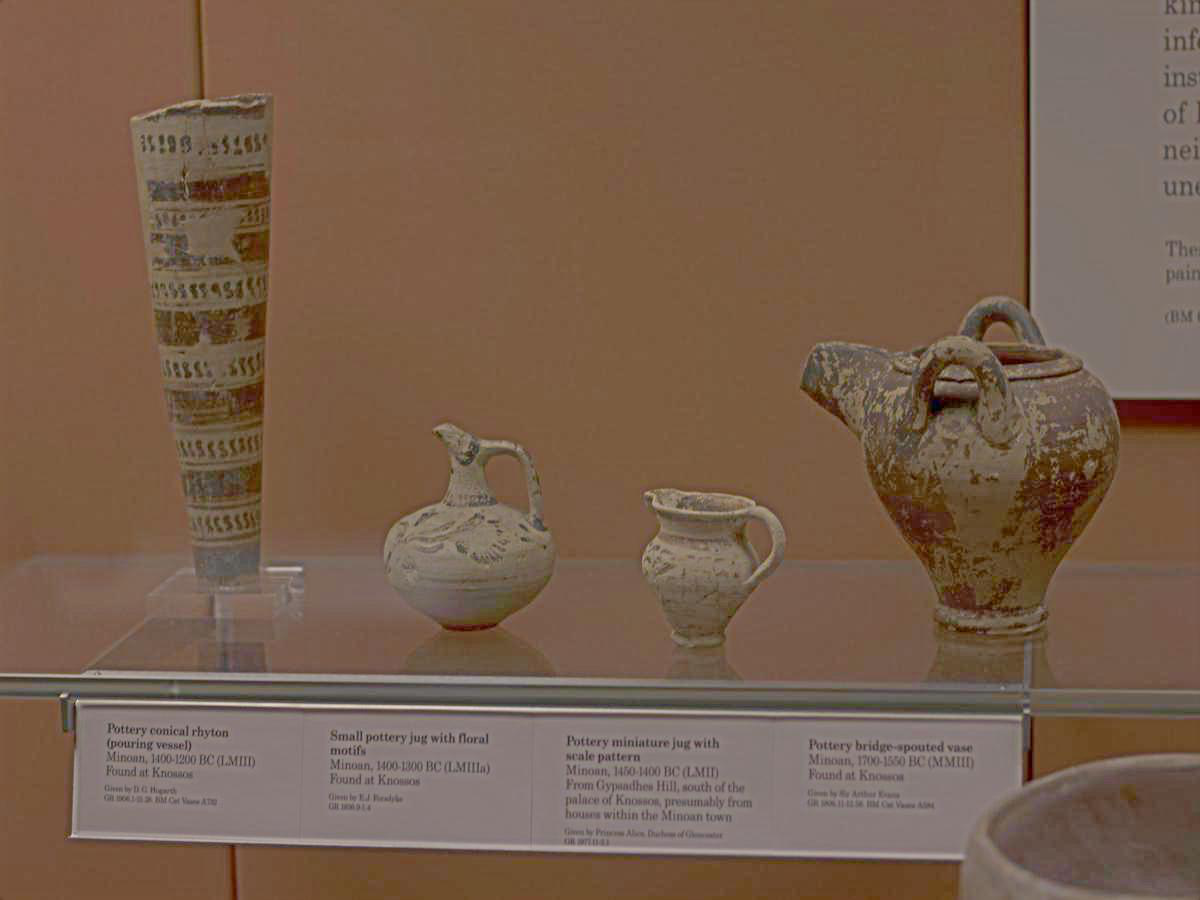}&		
			\includegraphics[width=.16\textwidth,height=2.2cm]{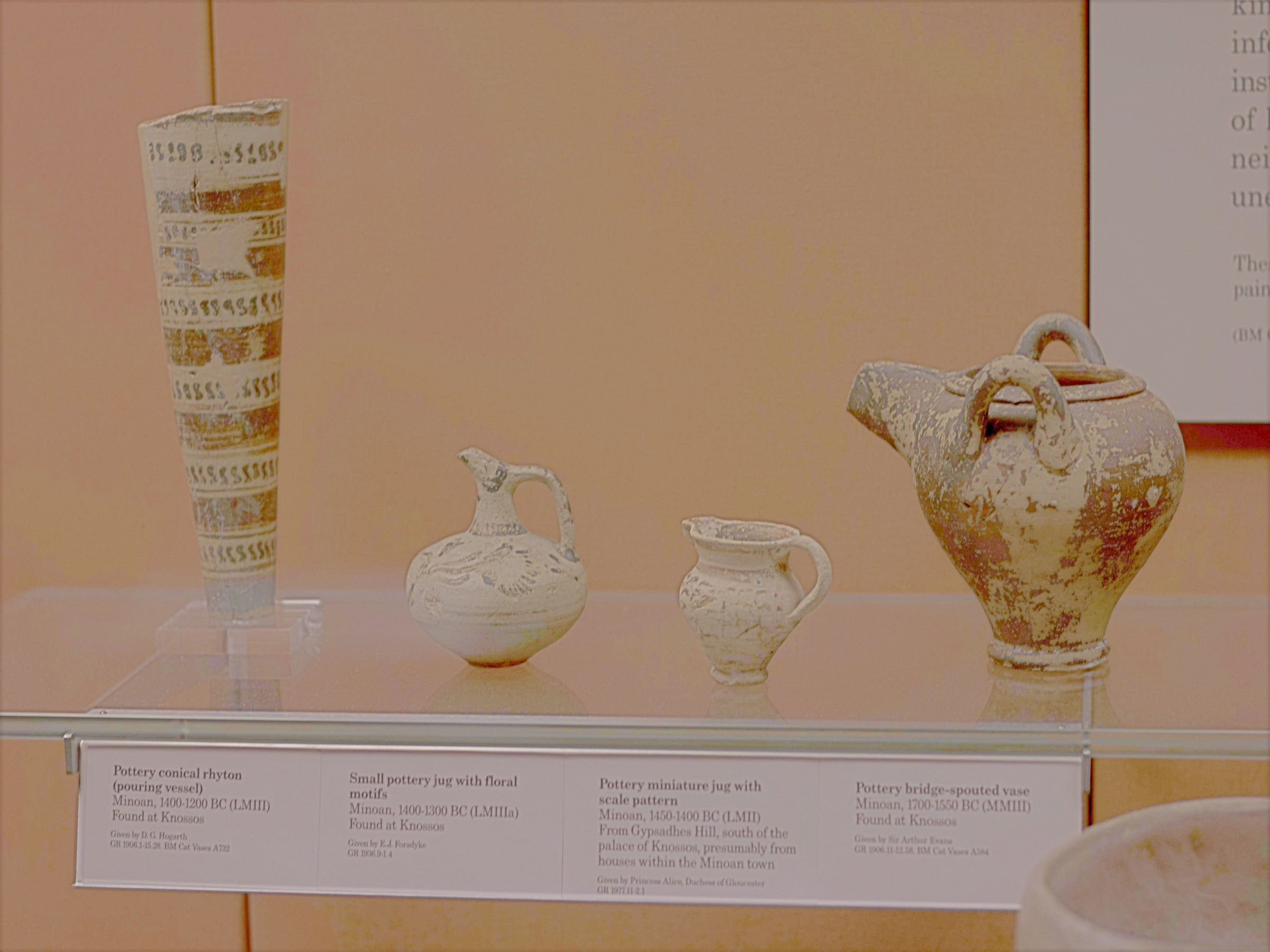}&
			\includegraphics[width=.16\textwidth,height=2.2cm]{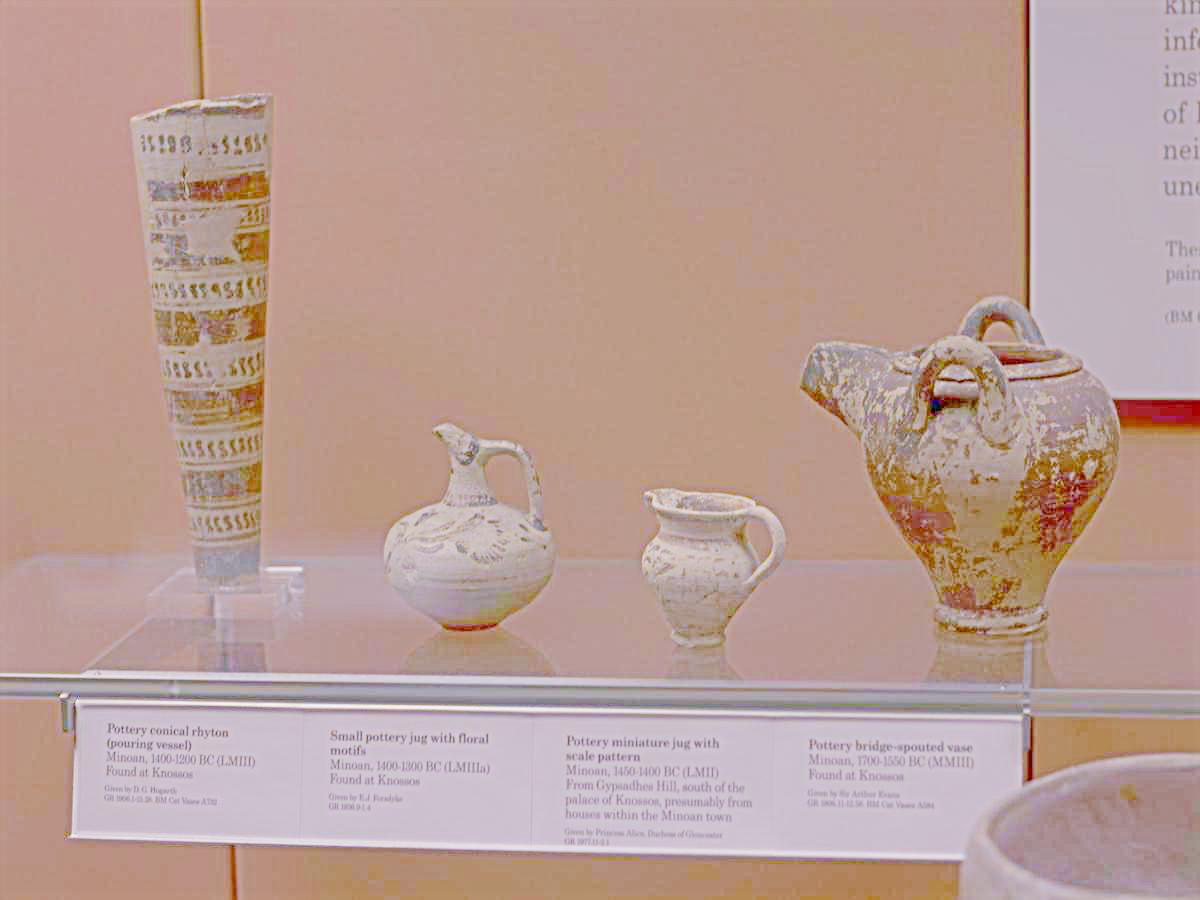}\\
			(a) input& (b) Zero-DCE$_{E0.4}$ & (c) Zero-DCE$_{E0.5}$& (d) Zero-DCE$_{E0.6}$ & (e) Zero-DCE$_{E0.7}$\\
		\end{tabular}
	\end{center}
	\caption{A visual comparison among the results generated by the Zero-DCE trained using different well-exposedness level, $E$, in exposure control loss (see Equation \eqref{equ_5}).}
	\label{fig:exposedness}
\end{figure*}

\begin{table}
	\caption{Quantitative comparisons in terms of  PSNR, SSIM, and MAE. These comparisons are carried out on Part2 testing set.}
	\centering
	\begin{tabular}{c|c|c|c}
		\hline
		\textbf{Method} & \textbf{PSNR$\uparrow$ } & \textbf{SSIM$\uparrow$}  & \textbf{MAE$\downarrow$} \\
		\hline
		input                 &   10.71    & 0.33       &209.65 \\
		Zero-DCE$_{E0.4}$  & 11.82     &  0.45     & 177.87\\
		Zero-DCE$_{E0.5}$  & 15.19     &   0.56    &114.47\\
		Zero-DCE$_{E0.6}$  &    16.57 &0.59   &98.78 \\ 
		Zero-DCE$_{E0.7}$   & 13.40    & 0.55 &156.83 \\
		\hline
	\end{tabular}
	\label{table_E}
\end{table}

\begin{figure*}
	\begin{center}
		\begin{tabular}{c@{ }c@{ }c@{ }c@{ }c@{ }c@{ }c}
			\includegraphics[width=.16\textwidth,height=2.2cm]{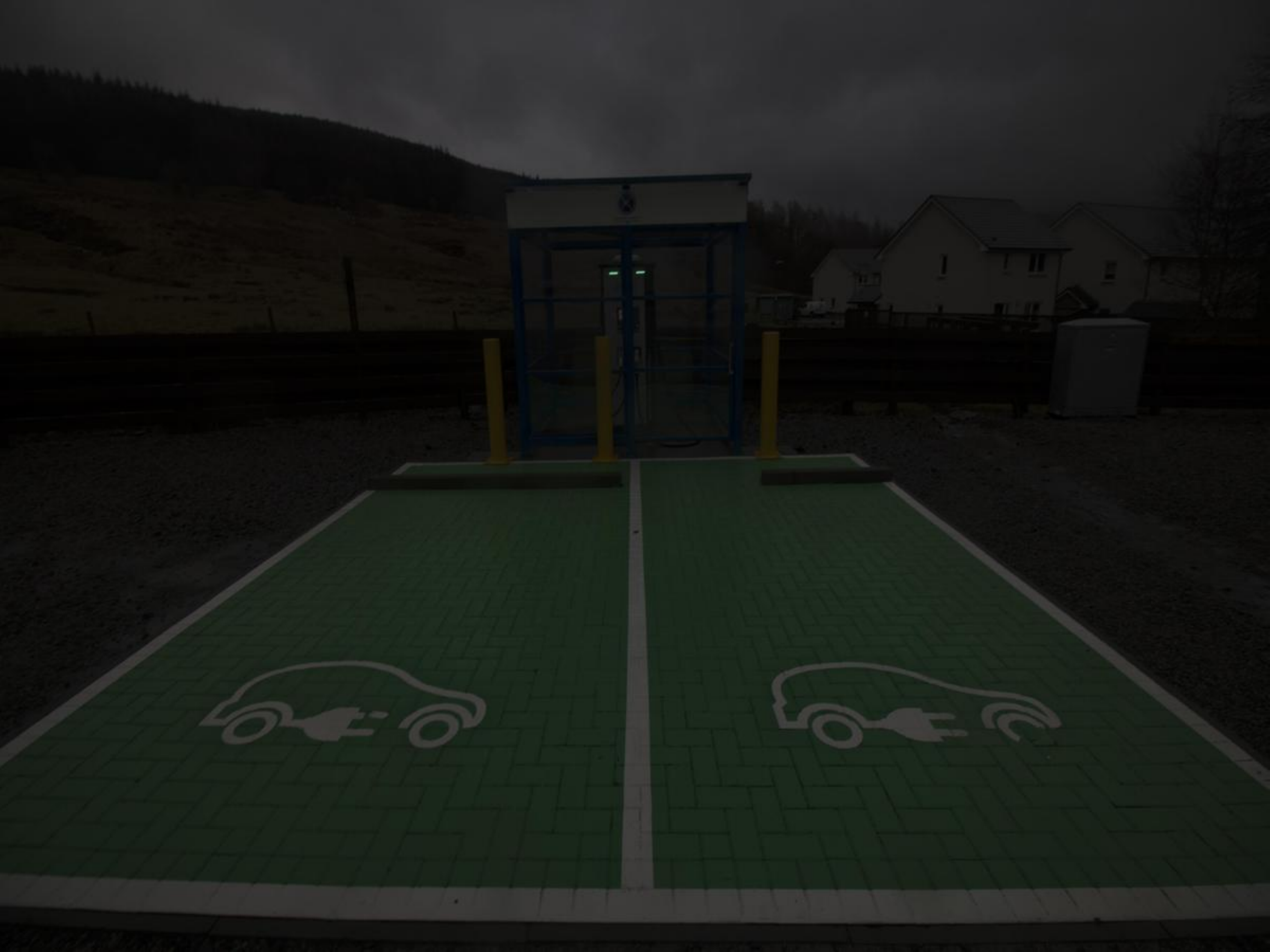}&
			\includegraphics[width=.16\textwidth,height=2.2cm]{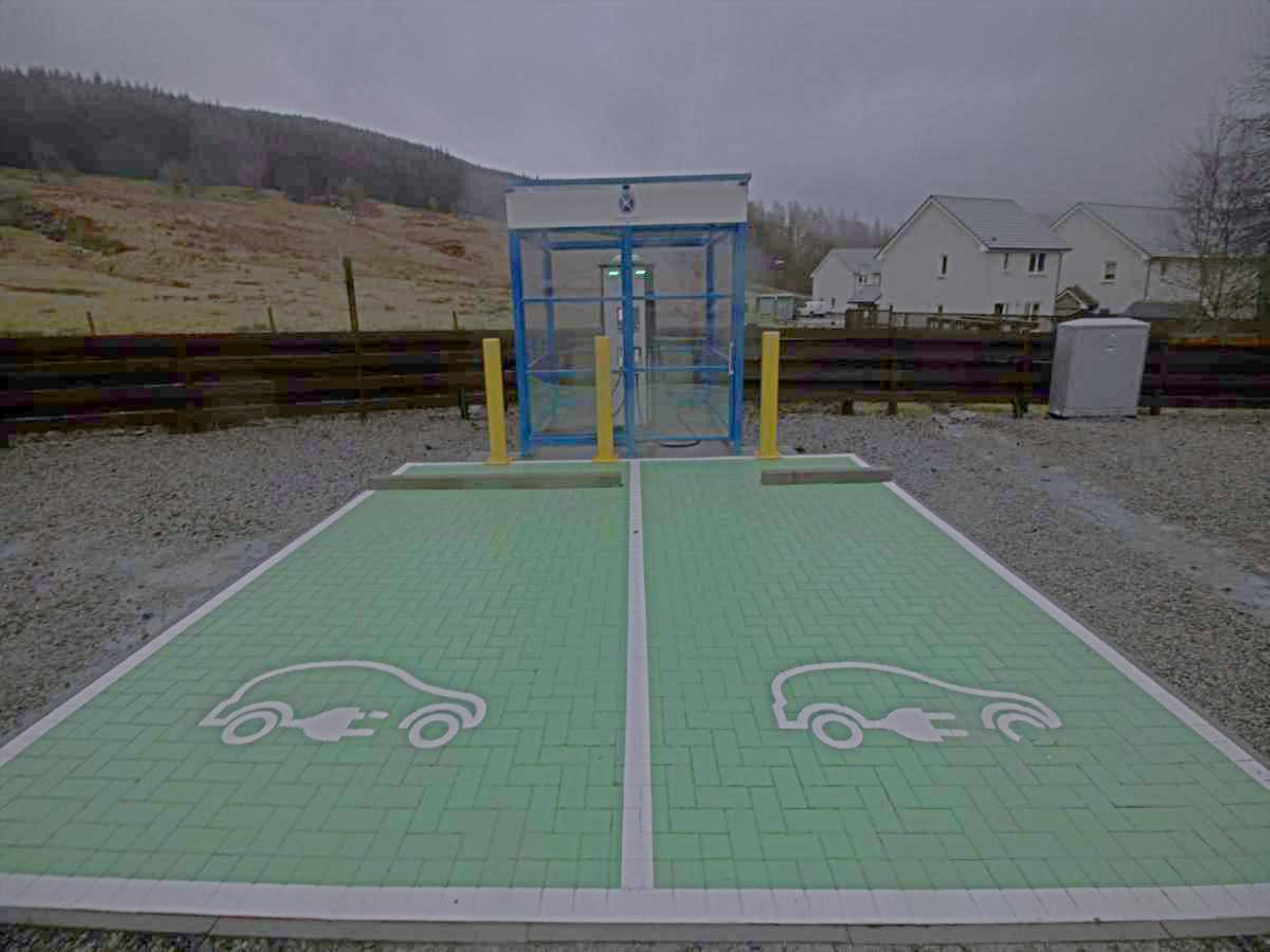}&
			\includegraphics[width=.16\textwidth,height=2.2cm]{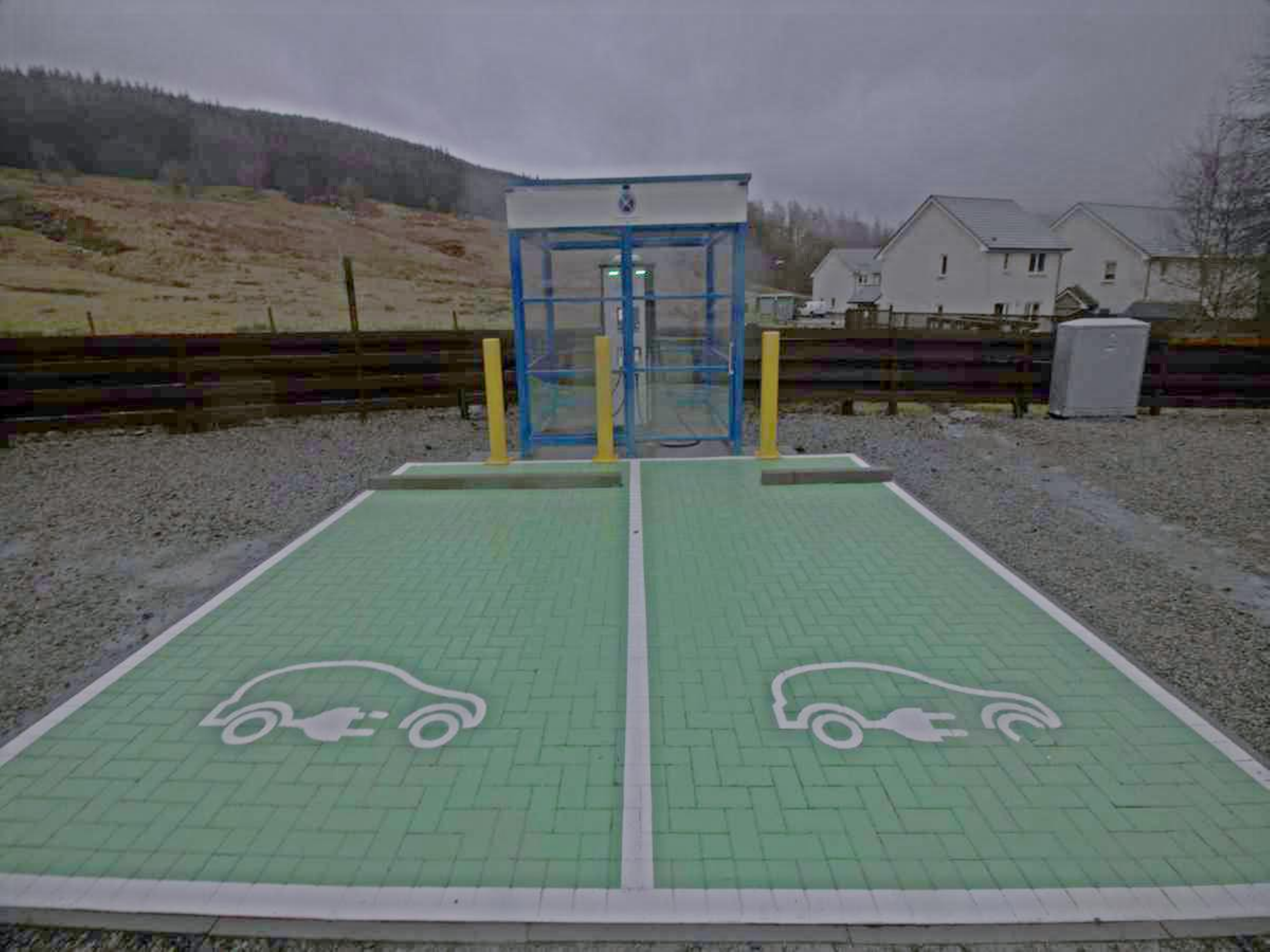}&
			\includegraphics[width=.16\textwidth,height=2.2cm]{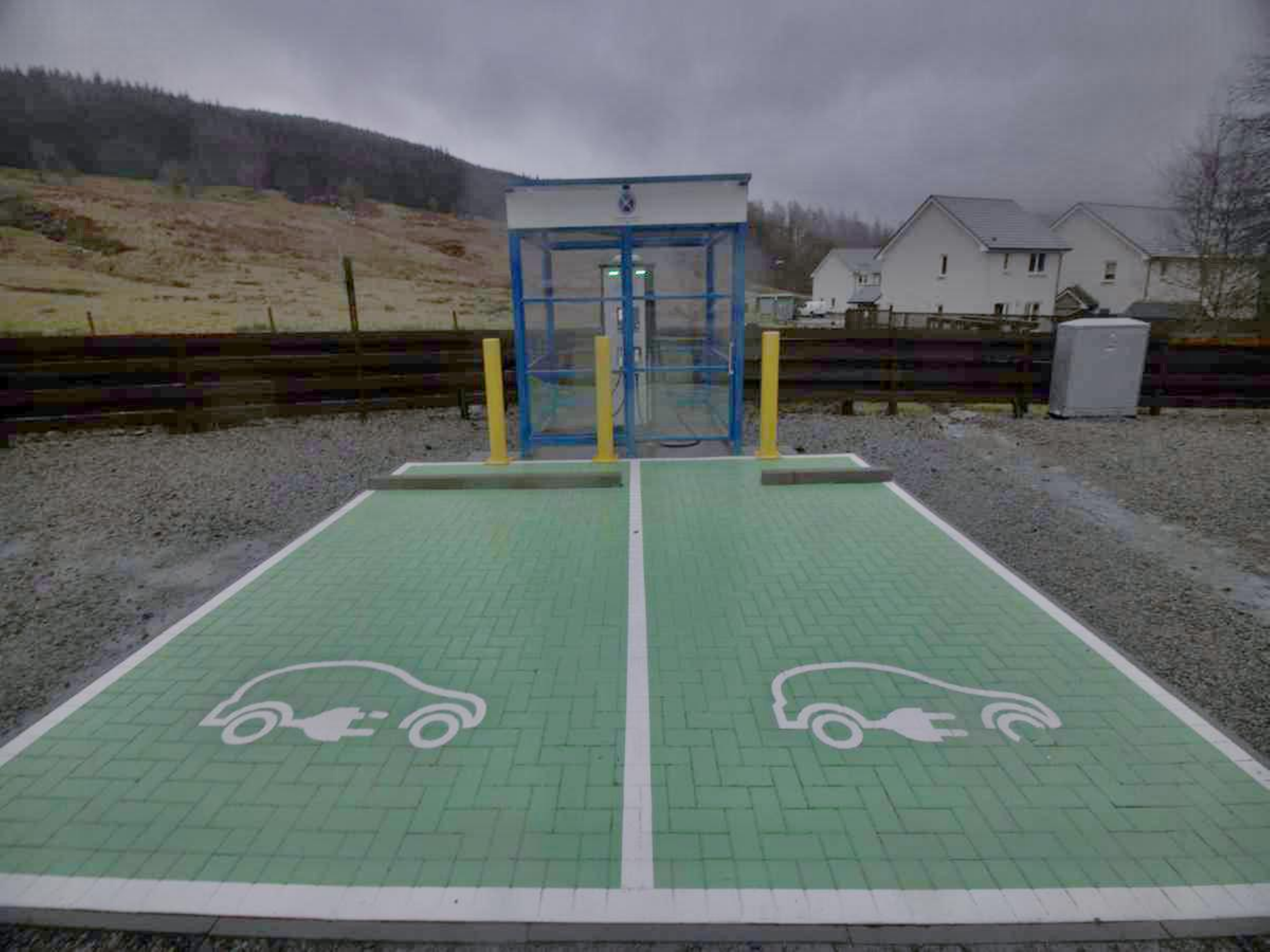}&
			\includegraphics[width=.16\textwidth,height=2.2cm]{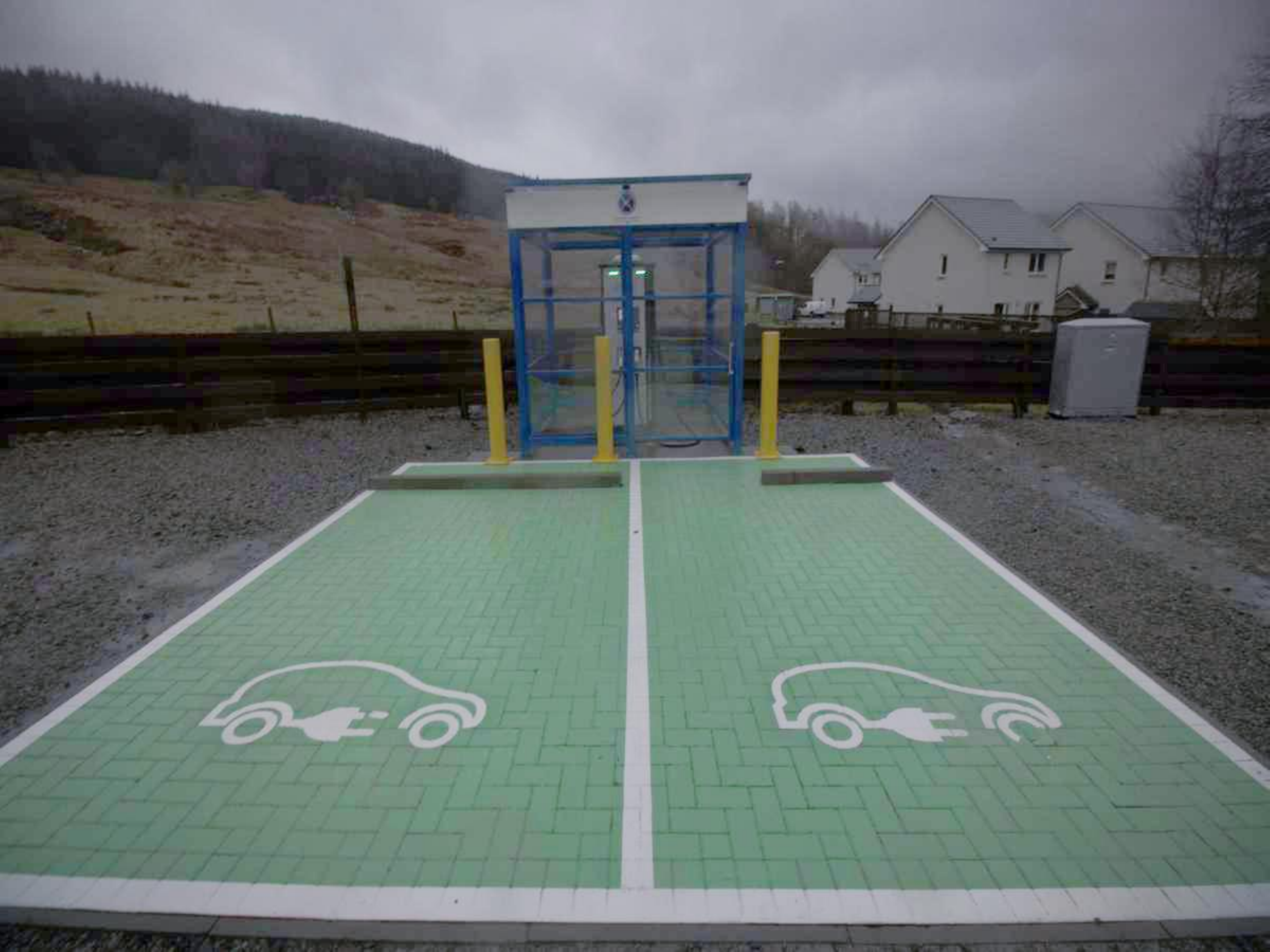}&
			\includegraphics[width=.16\textwidth,height=2.2cm]{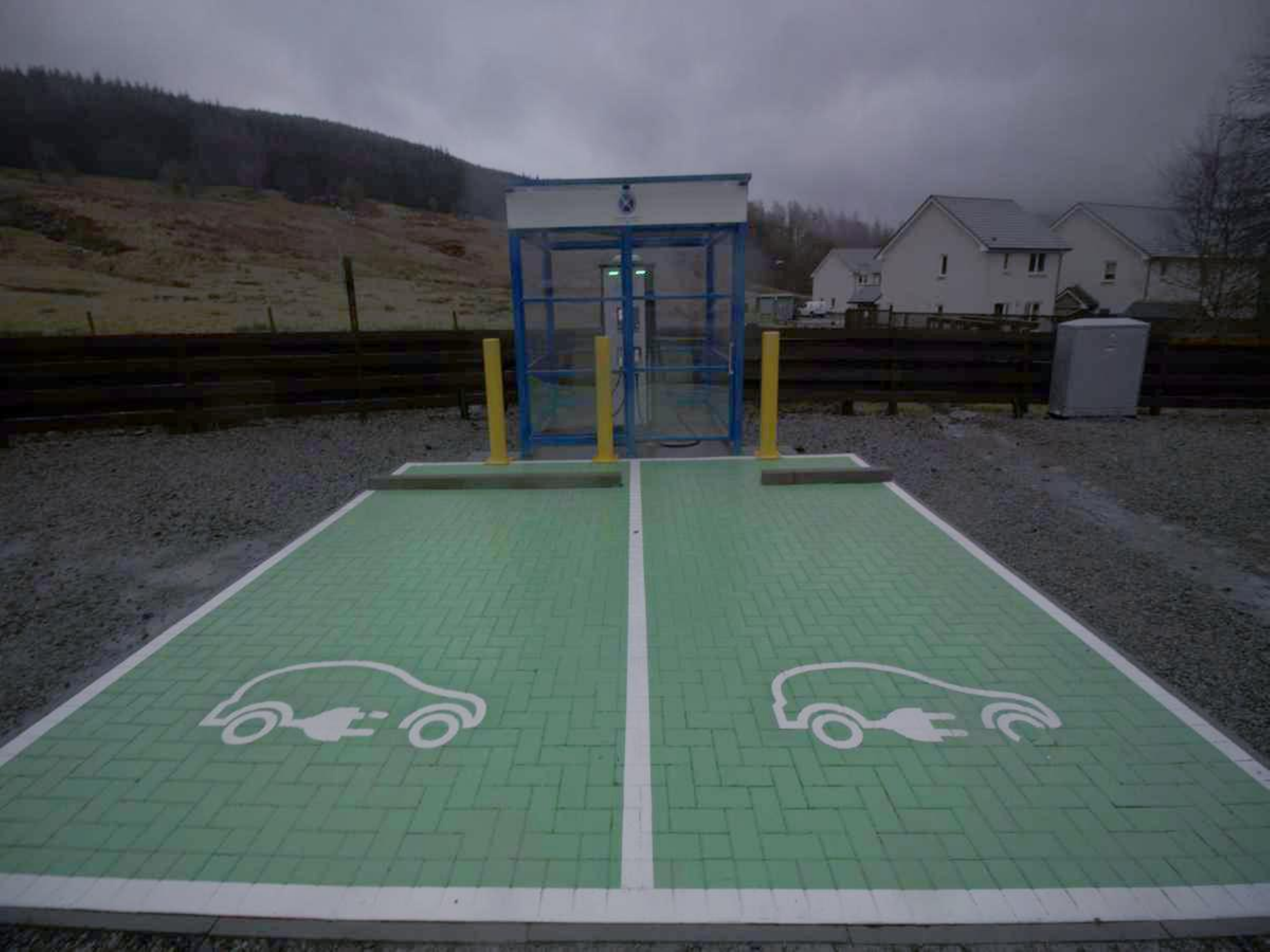}\\
			(a) input& (b) original resolution& (c) 2$\times$$\downarrow$ &	(d) 12$\times$$\downarrow$& (e) 100$\times$$\downarrow$ & (f) 300$\times$$\downarrow$\\
		\end{tabular}
	\end{center}
	\caption{A set of results by feeding different sizes of input to the modified framework. Even 300$\times$ downsampling does not hamper the performance much when compared with the original resolution as input. Here, $\downarrow$ represents the downsampling operation.}
	\label{fig:input_size}
\end{figure*}

\begin{table*}
	\caption{The statistic relations between enhancement performance and input sizes measured in PSNR and FLOPs. The FLOPs (in G) is computed  for an image of size 1200$\times$900$\times$3. ``number$\times$$\downarrow$'' indicates the times of downsampling the input image. These comparisons are carried out on Part2 testing set.}
	\centering
	\begin{tabular}{c|c|c|c|c|c|c|c|c|c|c}
		\hline
		Metrics	& original resolution & 2$\times$$\downarrow$& 4$\times$$\downarrow$& 6$\times$$\downarrow$& 12$\times$$\downarrow$& 20$\times$$\downarrow$&50$\times$$\downarrow$&75$\times$$\downarrow$ & 100$\times$$\downarrow$ & 300$\times$$\downarrow$\\
		\hline
		PSNR &16.09  &16.17 &16.29 &16.37 &16.42 &16.33 &15.85 &15.56 &15.35 &14.53   \\
		FLOPs &11.442  &2.887  &0.749 &0.352 &0.115 &0.064 & 0.040 &0.037 &0.036 &0.035    \\
		\hline
	\end{tabular}
	\label{table_111}
\end{table*}

\noindent
\textbf{Impact of Training Data.}
To test the impact of training data, we retrain the Zero-DCE on datasets different from that described in Sec.~\ref{Implementation}.
As shown in Figure~\ref{fig:trainingdata}(c) and (d), after removing the over-exposed training data, Zero-DCE tends to over-enhance the well-lit regions (\eg, the cup in the results of  Zero-DCE$_{Low}$ and Zero-DCE$_{LargeL}$), in spite of using more low-light images, (\ie, Zero-DCE$_{LargeL}$). Such results indicate the rationality and necessity of the usage of multi-exposure training data in the training process of our network.  In addition, the Zero-DCE can better recover the dark regions (\eg. the roses)  when more multi-exposure training data are used (\ie, Zero-DCE$_{LargeLH}$), as shown in Figure~\ref{fig:trainingdata}(e). The proposed Zero-DCE as shown in Figure~\ref{fig:trainingdata}(b) has a good balance between over-enhancement and under-enhancement. For a fair comparison with other deep learning-based methods, we use a comparable amount of training data with them although more training data can bring better visual performance to our approach.

\noindent
\textbf{Effect of Well-Exposedness Level.}
We study the effect of well-exposedness level $E$ used in the exposure control loss on the enhancement performance of our method. We set four different well-exposedness levels $E$ (\ie, 0.4, 0.5, 0.6, 0.7) to train our network, denoted as Zero-DCE$_{E0.4}$, Zero-DCE$_{E0.5}$, Zero-DCE$_{E0.6}$ (\ie, our final Zero-DCE model), and Zero-DCE$_{E0.7}$, respectively.  A set of visual results are shown in Figure \ref{fig:exposedness}. The quantitative comparisons are presented in Table \ref{table_E}.

As shown in Table \ref{table_E}, Zero-DCE$_{E0.6}$ achieves the best quantitative scores.
Zero-DCE$_{E0.5}$ obtains comparable performance to  Zero-DCE$_{E0.6}$. The quantitative performance of Zero-DCE$_{E0.4}$ and Zero-DCE$_{E0.7}$ is slightly inferior to that of Zero-DCE$_{E0.6}$ and Zero-DCE$_{E0.5}$. As observed in Figure \ref{fig:exposedness},  Zero-DCE$_{E0.5}$ and Zero-DCE$_{E0.6}$ obtain visually pleasing brightness. In contrast,  Zero-DCE$_{E0.4}$ produces under-exposure while  Zero-DCE$_{E0.7}$ over-enhances the input image. Finally, we choose  Zero-DCE$_{E0.6}$ as the final model based on its good qualitative and quantitative performance.

\begin{table}
	\caption{Ablation study between Zero-DCE and Zero-DCE++ in terms of PSNR, trainable parameters (\#P), and FLOPs (in G). The FLOPs is computed  for an image of size  1200$\times$900$\times$3. The average PSNR values are computed on Part2 testing set.}
	\centering
	\begin{tabular}{c|c|c||c|c|c}
		\hline
		Zero-DCE & DSconv& Pshared& PSNR$\uparrow$   &\#P$\downarrow$ & FLOPs$\downarrow$ \\
		\hline
		$\checkmark$& & &16.57  &79,416 &84.99    \\
		&$\checkmark$&  &16.51 &11,926 &0.375    \\
		&& $\checkmark$& 16.24&67,299 &0.540    \\
		&$\checkmark$& $\checkmark$&16.42 &10.561  & 0.115    \\
		\hline
	\end{tabular}
	\label{table_222}
\end{table}

\begin{figure}[!h]
	\begin{center}
		\begin{tabular}{c@{ }c@{ }c@{ }}
			\includegraphics[width=.15\textwidth,height=4cm]{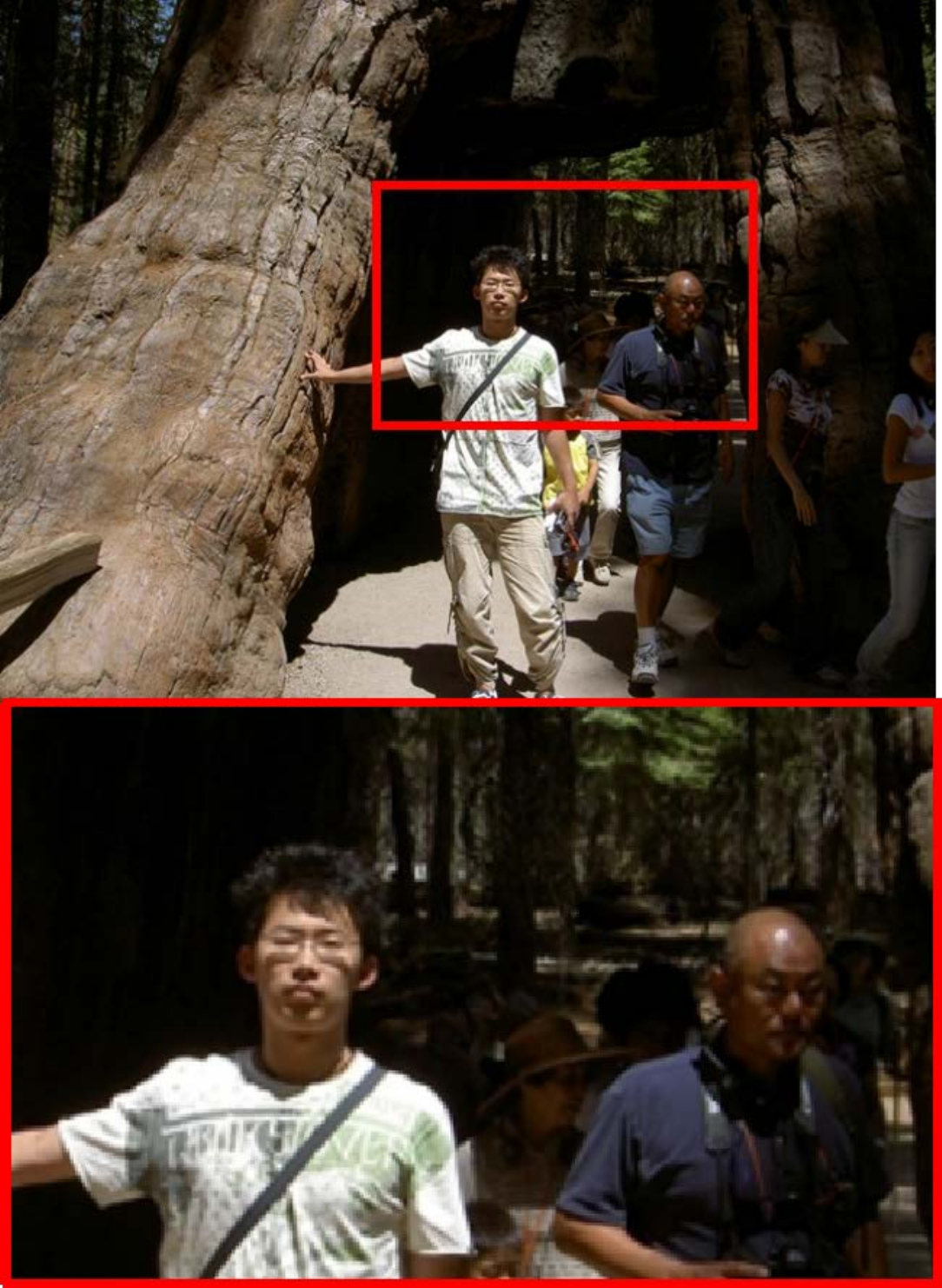}&
			\includegraphics[width=.15\textwidth,height=4cm]{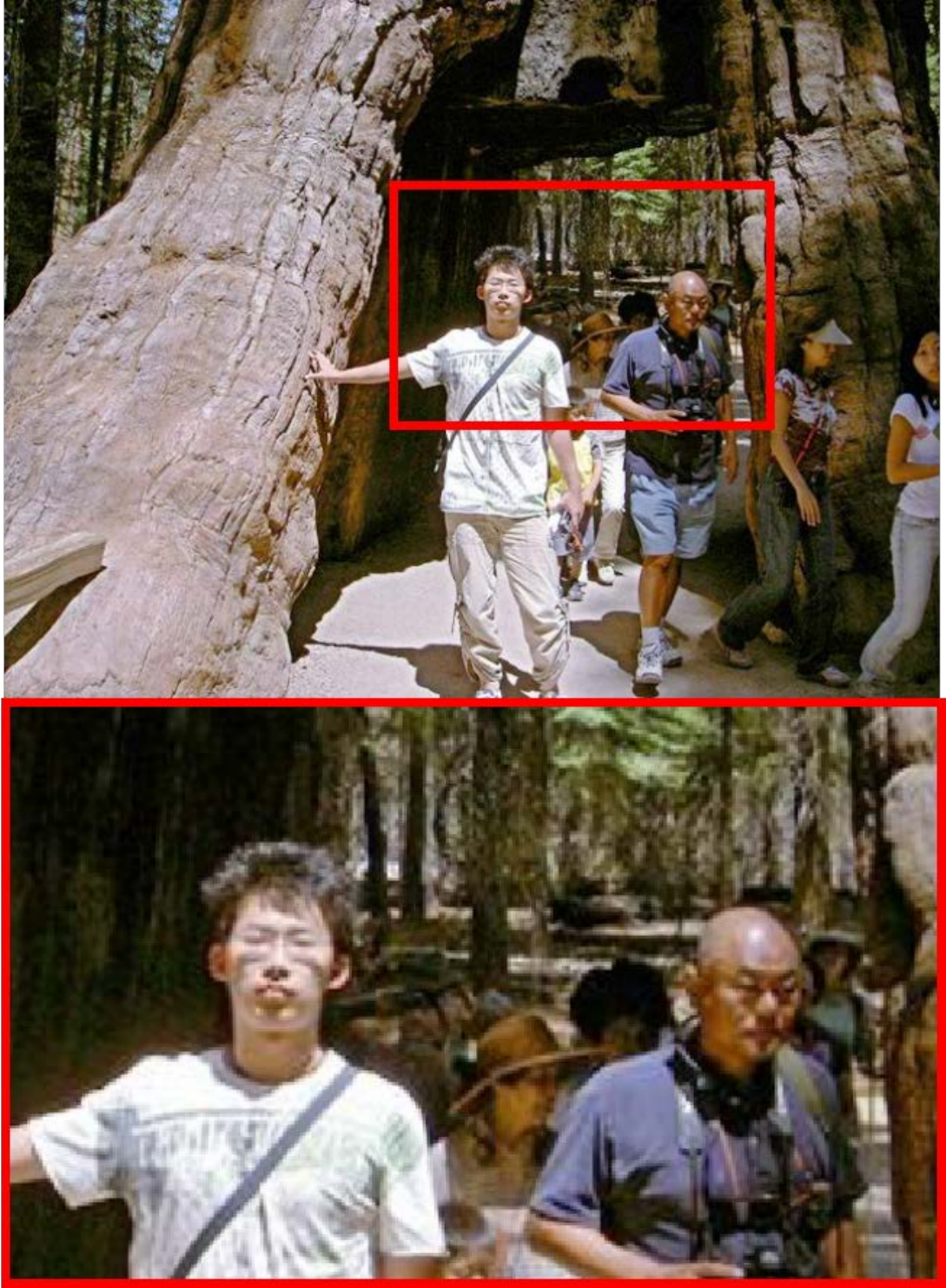} &	\includegraphics[width=.15\textwidth,height=4cm]{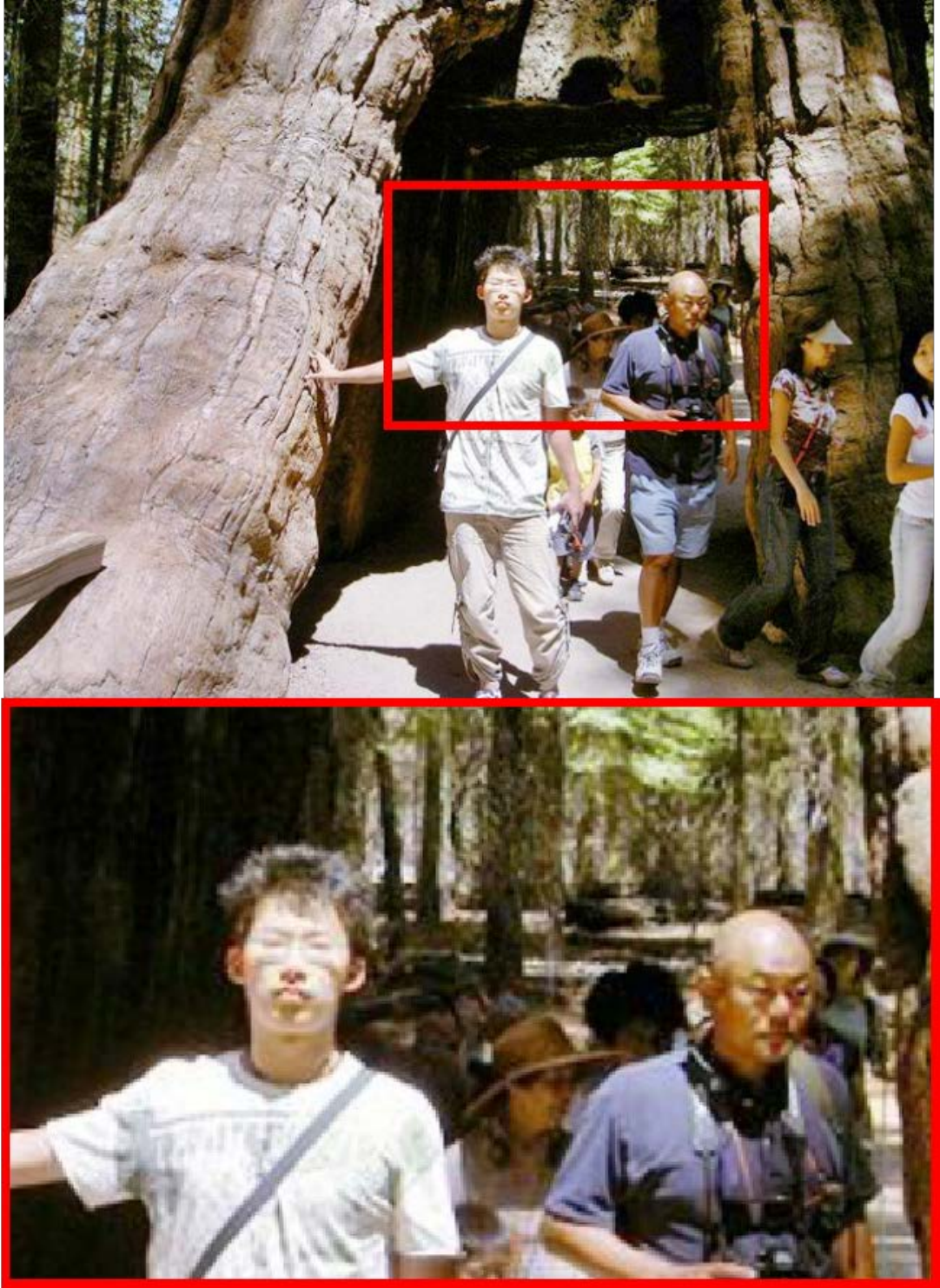}\\
			(a) input& (b) Zero-DCE& (c) Zero-DCE++\\
		\end{tabular}
	\end{center}
	\caption{A visual comparison between the results generated by Zero-DCE and Zero-DCE++. Zero-DCE shows better capability in handling extreme lighting conditions.}
	\label{fig:zero_dce++}
\end{figure}

\begin{table*}[!t]
	\caption{User study (US)$\uparrow$/Perceptual index (PI)$\downarrow$ scores on the image sets (NPE, LIME, MEF, DICM, VV). Higher US score indicates better human subjective visual quality while lower PI value indicates better perceptual quality. The best result is in red whereas the second best one is in blue under each case.}
	\centering
	\begin{tabular}{c|c|c|c|c|c|c}
		\hline
		\textbf{Method}               & \textbf{NPE} & \textbf{LIME} &\textbf{MEF}  & \textbf{DICM} & \textbf{VV} & \textbf{Average}\\
		\hline
		SRIE~\cite{Fu2016}           &  3.65/{\color{blue}2.79} & 3.50/{\color{red}2.76}               &  3.22/2.61           &  3.42/3.17             & 2.80/3.37                  &3.32/2.94 \\
		LIME~\cite{Guo2017}          &  3.78/3.05           & {\color{red}3.95}/3.00  &  3.71/2.78            &  3.31/3.35             & 3.21/{\color{blue}3.03} & 3.59/3.04\\
		Li \etal~\cite{Li2018}           &  3.80/3.09           &3.78/3.02       & 2.93/3.61                &3.47/3.43                & 2.87/3.37                        & 3.37/3.72\\
		LightenNet \cite{LightenNet} &3.76/2.88 & 3.02/2.84 & 3.07/2.51& 3.11/{\color{blue}3.13}&2.55/3.29 & 2.70/{\color{blue}2.93}\\
		MBLLEN \cite{MBLLEN} &3.81/{\color{red}2.77} & 3.77/3.18& 3.21/3.04& 3.07/3.19& 2.72/3.63& 3.33/3.16  \\
		RetinexNet~\cite{Chen2018}   &  3.30/3.18           & 2.32/3.08         &2.80/2.86        &  2.88/3.24       & 1.96/{\color{red}2.95}              &2.58/3.06\\
		Wang \etal~\cite{Wang2019}      & {\color{blue}3.83}/2.83  & 3.82/2.90     &3.13/2.72           & 3.44/3.20      &  2.95/3.42 & 3.43/3.01  \\
		EnlightenGAN~\cite{Jiang2019}&  {\color{red}3.90}/2.96 & {\color{blue}3.84}/{\color{blue}2.83} & 3.75/{\color{blue}2.45} &{\color{blue}3.50}/{\color{blue}3.13} & 3.17/4.71              &3.63/3.22\\
		Zero-DCE                     &  3.81/2.84 & 3.80/{\color{red}2.76}               &  {\color{red}4.13}/{\color{red}2.43}&  {\color{red}3.52}/{\color{red}3.04} & {\color{blue}3.24}/3.33     &{\color{red}3.70}/{\color{red}2.88}\\
		Zero-DCE++        & 3.79/2.93  &  3.81/2.97          &{\color{blue}4.10}/2.50 &3.48/3.21   & {\color{red}3.26}/3.31   &{\color{blue}3.69}/2.98\\
		\hline
	\end{tabular}
	\label{label:user study}
\end{table*}

\begin{table}[t]
	\caption{Quantitative comparisons in terms of PSNR, SSIM, and MAE  on the Part2 testing set. The best result is in red whereas the second best one is in blue under each case.}
	\centering
	\begin{tabular}{c|c|c|c}
		\hline
		\textbf{Method} & \textbf{PSNR$\uparrow$ } & \textbf{SSIM$\uparrow$}  & \textbf{MAE$\downarrow$} \\
		\hline
		SRIE~\cite{Fu2016}           &  14.41      & 0.54 & 127.08\\
		LIME~\cite{Guo2017}          &  16.17      & 0.57 & 108.12\\
		Li \etal~\cite{Li2018}          &  15.19      &   0.54 &114.21         \\
		RetinexNet~\cite{Chen2018}   &  15.99      & 0.53 & 104.81 \\
		LightenNet \cite{LightenNet} & 13.17 & 0.55&140.92  \\
		MBLLEN  \cite{MBLLEN} &15.02 &0.52 &119.14 \\
		Wang \etal~\cite{Wang2019}      &  13.52      &0.49  & 142.01\\
		EnlightenGAN~\cite{Jiang2019}&  16.21    & {\color{red}0.59} & {\color{blue}102.78}\\
		Zero-DCE                      &  {\color{red}16.57}     & {\color{red}0.59} & {\color{red}98.78}\\
		Zero-DCE++                      &  {\color{blue}16.42}     & {\color{blue}0.58} & 102.87\\
		\hline
	\end{tabular}
	\label{table_1}
\end{table}

\noindent
\textbf{Zero-DCE VS. Zero-DCE++.}
We first analyze the effect of input sizes on the enhancement performance in our method. As specified in Sec.~\ref{Zero-DCE++}, we first replace convolutional layers of DCE-Net with depthwise separation convolutions and reuse the curve parameter maps across eight iterations. Then, we feed the different sizes of input to the modified framework. The statistic relations between enhancement performance and input sizes are summarized  in Table~\ref{table_111}. We also show several results by feeding  different sizes of input to the modified framework in Figure~\ref{fig:input_size}. As shown in Table~\ref{table_111} and Figure~\ref{fig:input_size}, downsampling the sizes of input has unnoticeable effect on the enhancement performance but significantly saves computational cost (measured in FLOPs). As shown, results of 12$\times$$\downarrow$ achieve the highest average PSNR value, thus we adopt it as the default operation in Zero-DCE++.

Then, we conduct an ablation study to compare the network structures between Zero-DCE and Zero-DCE++ by replacing the modified component.  The ablated models include the Zero-DCE with the depthwise separable convolutions (denoted as DSconv) and the Zero-DCE that shares the curve parameter maps in different iteration stages (denoted as Pshared). The input of Zero-DC is the original resolution image while 12$\times$ downsampling operation as default is used in ``DSconv'' and ``Pshared''. The quantitative comparison results of the ablated models are presented in Table \ref{table_222}. 

As shown in Table \ref{table_222}, both ``DSconv'' and ``Pshared'' have fewer trainable parameters and FLOPs than Zero-DCE. Introducing ``DSconv'' and ``Pshared'' slightly decreases the PSNR values.  The trainable parameters and FLOPs of the combination of ``DSconv'' and ``Pshared'' (\ie, Zero-DCE++) are significantly decreased with negligible  decrease in PSNR value. The results suggest the effectiveness of such modifications. In Figure \ref{fig:zero_dce++}, Zero-DCE still outperforms Zero-DCE++ in some challenging cases. For example, Zero-DCE can more effectively handle challenging lighting without introducing over-/under-exposure when compared with Zero-DCE++.
One could choose between Zero-DCE and Zero-DCE++ according to the specific requirements on quality and efficiency.

\begin{figure*}[t]
	\begin{center}
		\begin{tabular}{c@{ }c@{ }c@{ }c@{ }c@{ }c@{ }}
			\includegraphics[height=5.1cm,width=3.5cm]{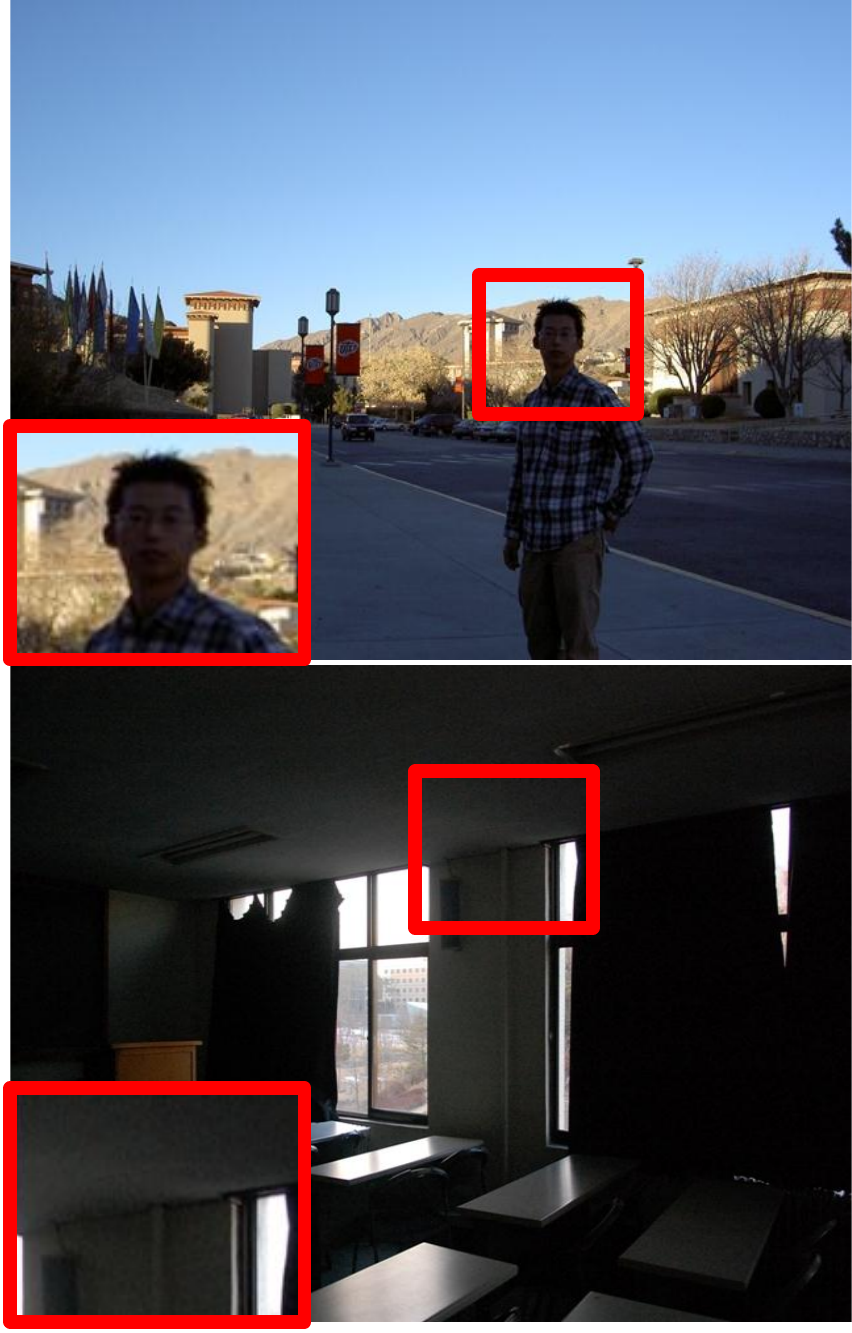}&~~
			\includegraphics[height=5.1cm,width=3.5cm]{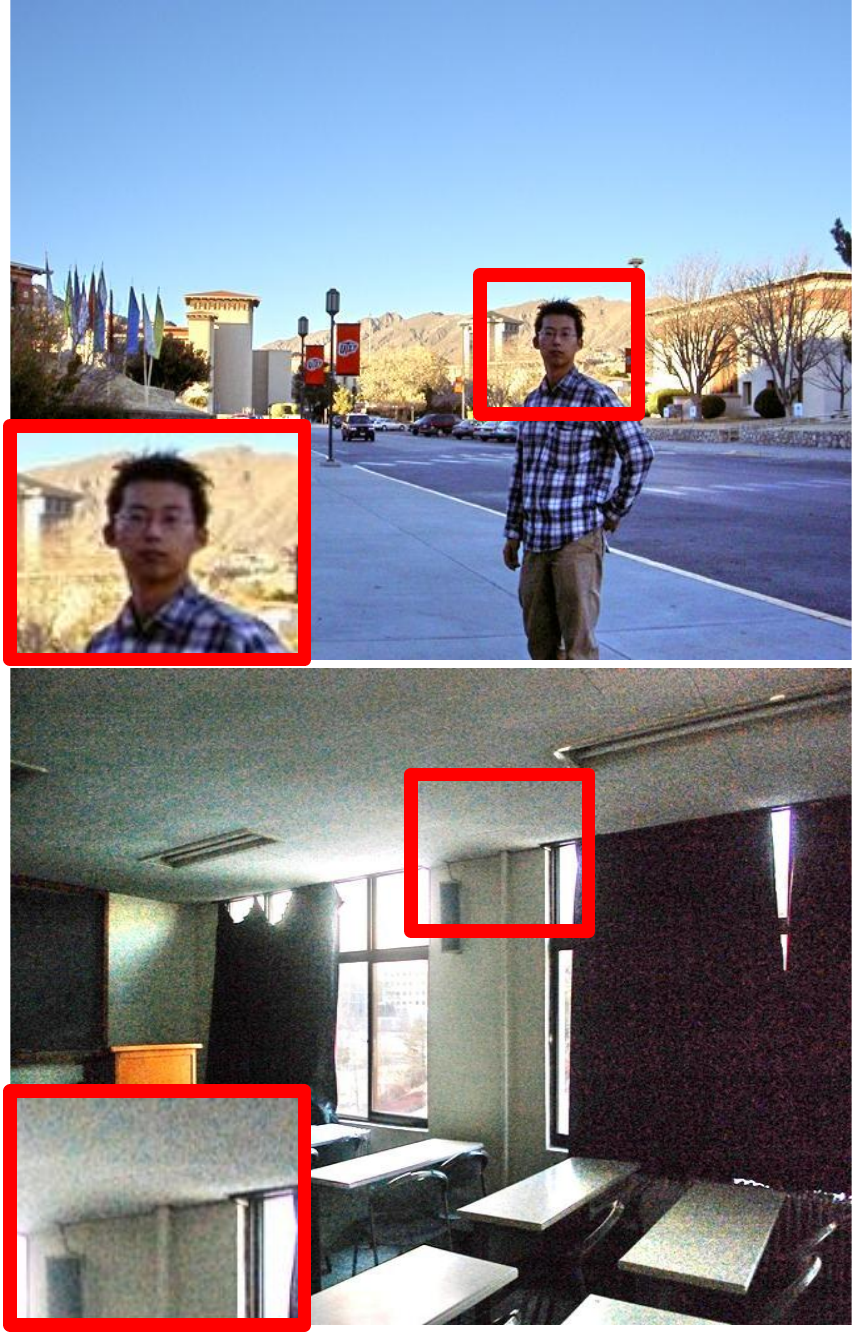}&~~
			\includegraphics[height=5.1cm,width=3.5cm]{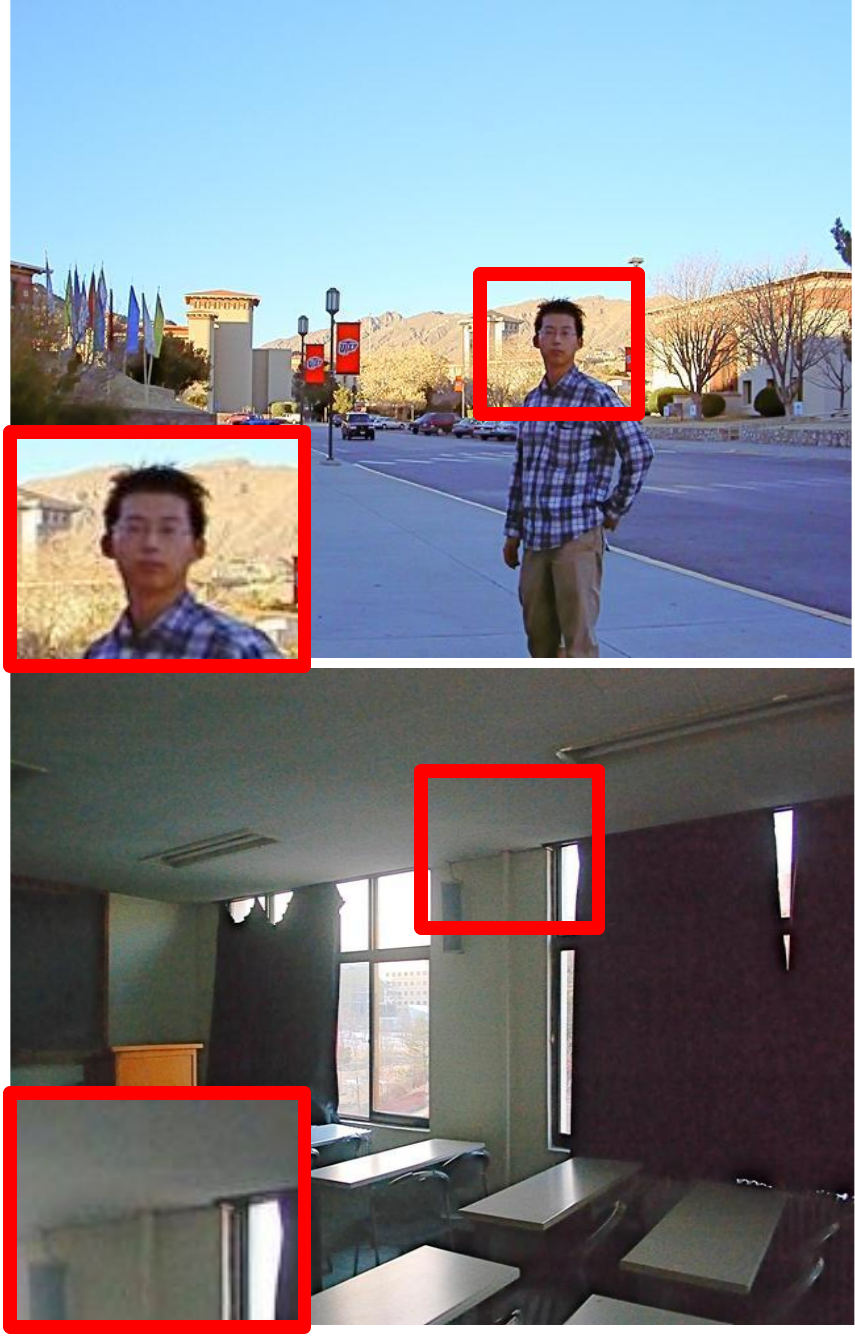}&~~
			\includegraphics[height=5.1cm,width=3.5cm]{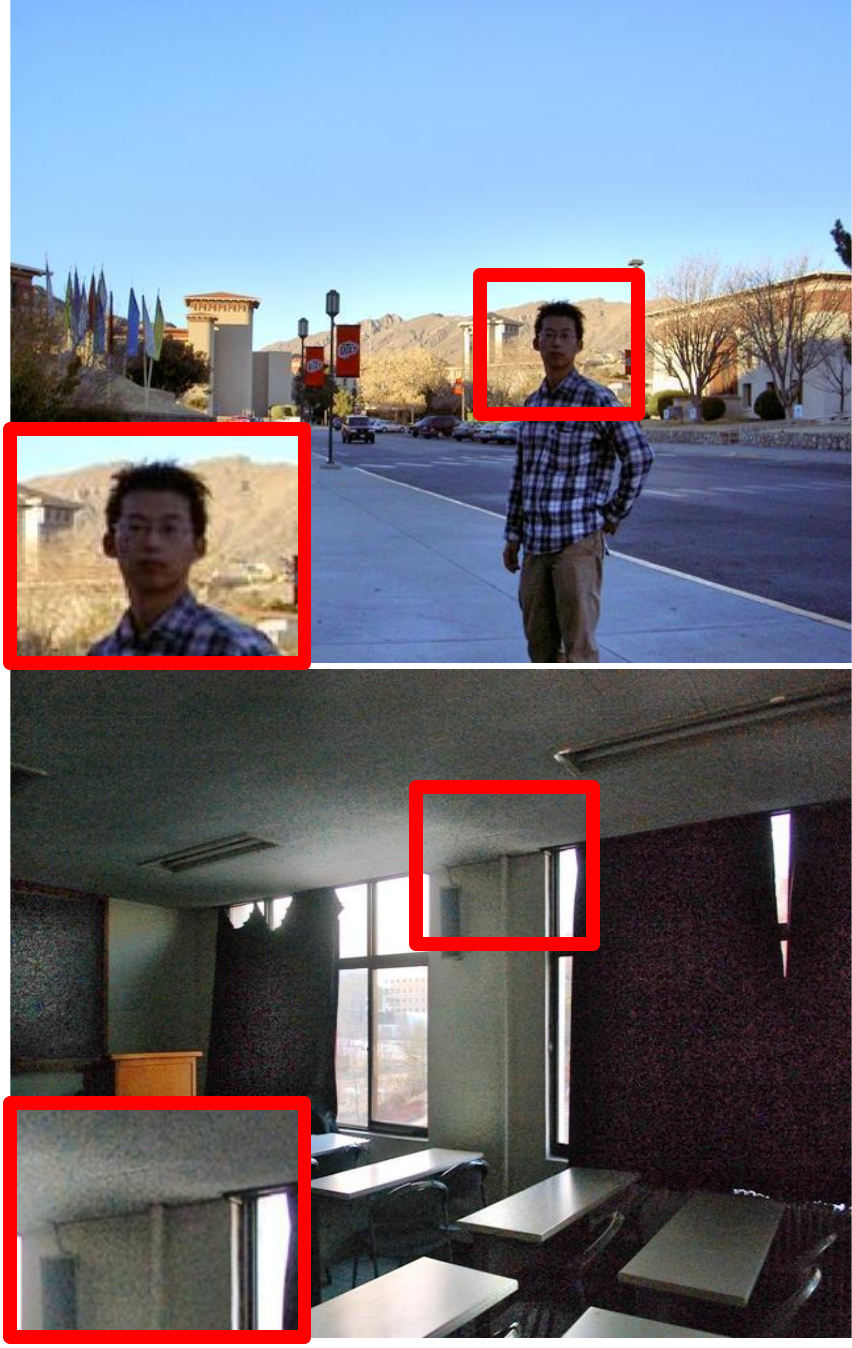}&~~
			\includegraphics[height=5.1cm,width=3.5cm]{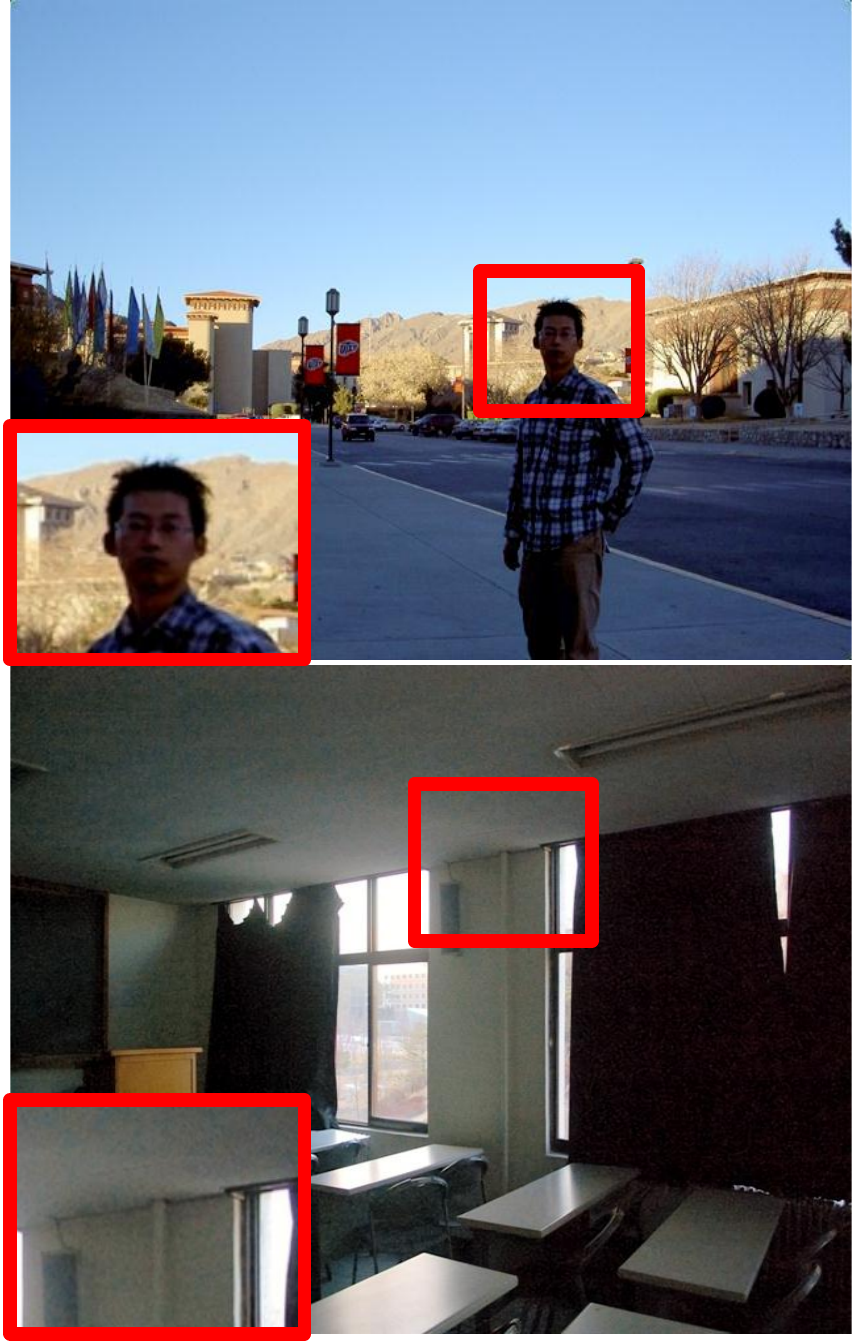}\\
			(a) inputs &  (b)  LIME~\cite{Guo2017} & (c) Li \etal~\cite{Li2018} & (d) LightenNet~\cite{LightenNet} & (e)  MBLLEN~\cite{MBLLEN} \\
			\includegraphics[height=5.1cm,width=3.5cm]{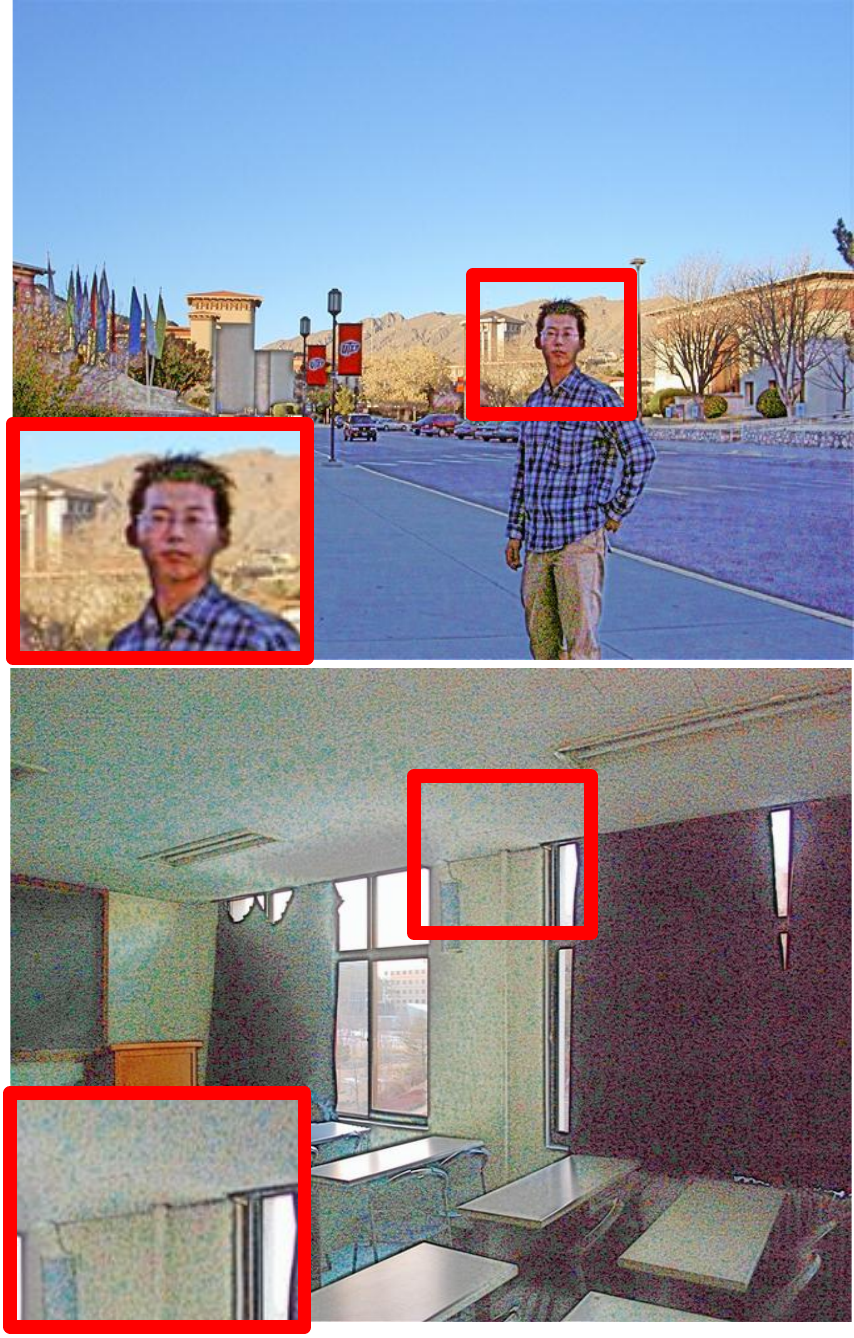}&
			\includegraphics[height=5.1cm,width=3.5cm]{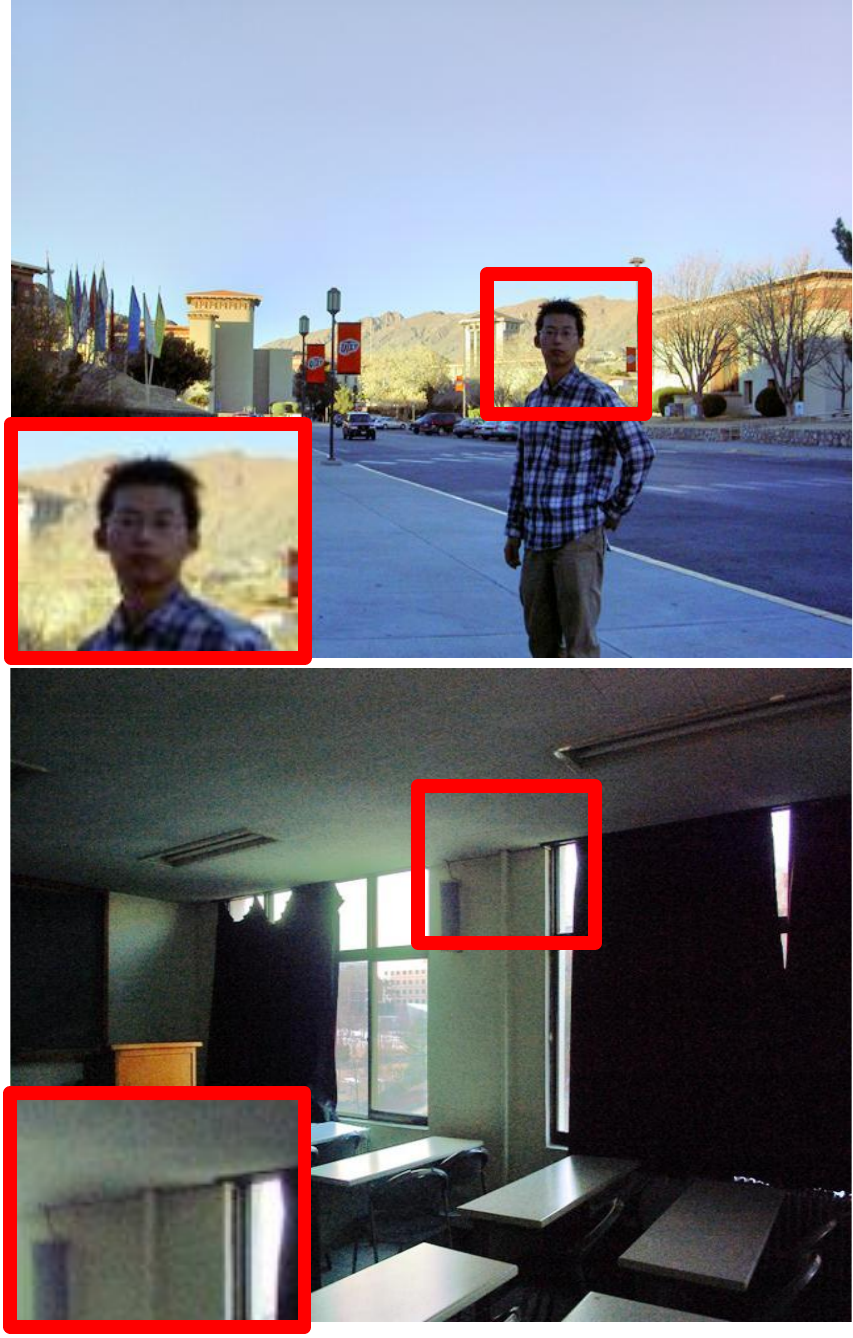}&
			\includegraphics[height=5.1cm,width=3.5cm]{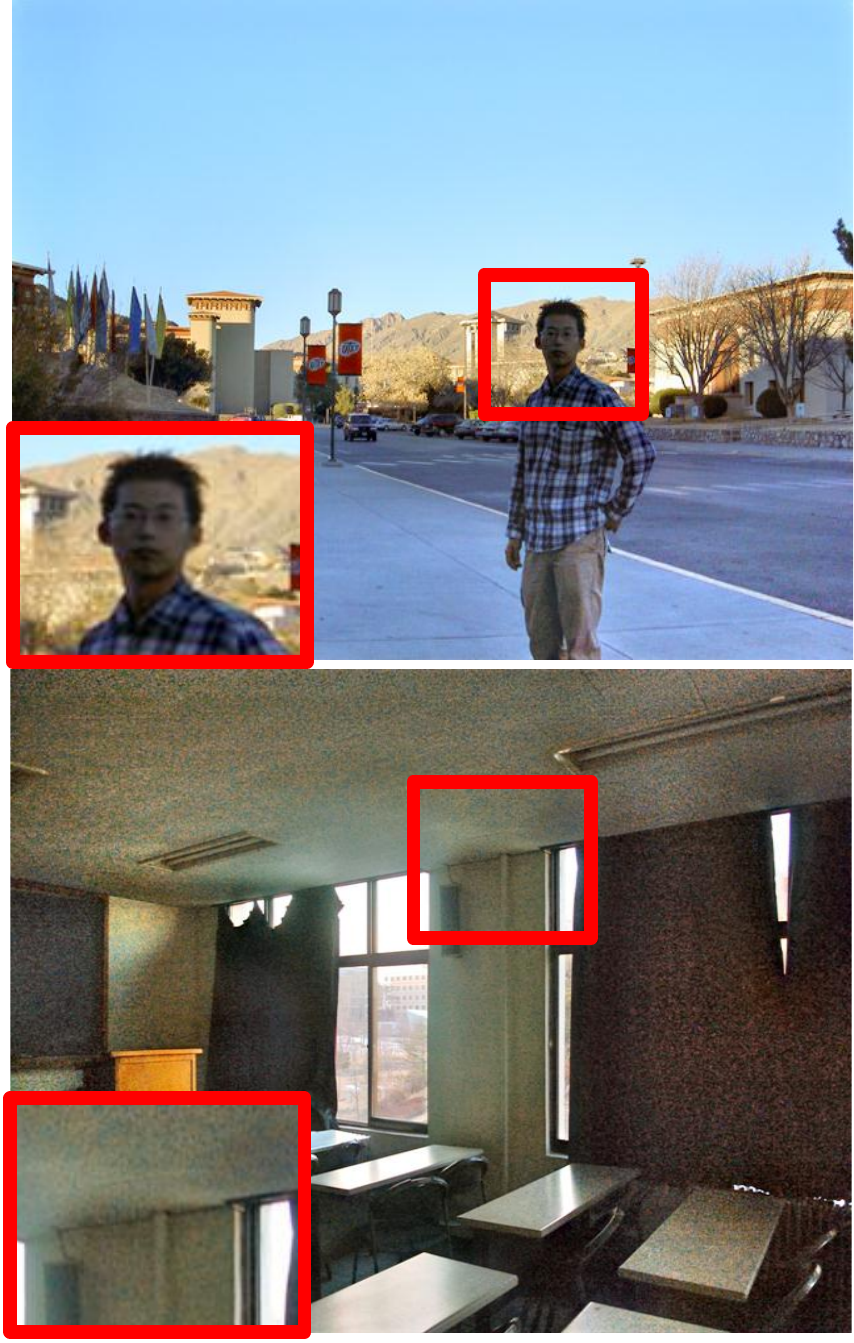}&
			\includegraphics[height=5.1cm,width=3.5cm]{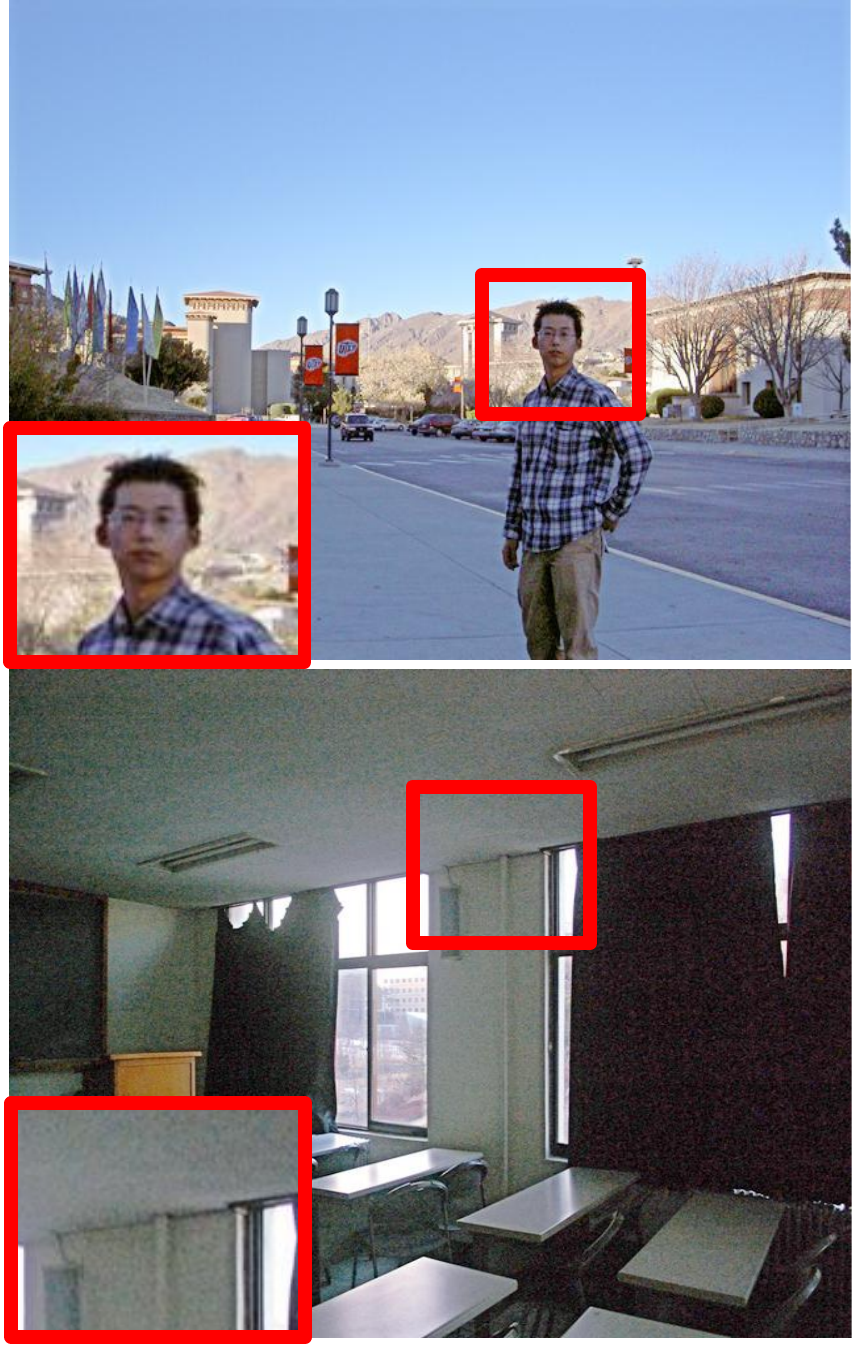}&
			\includegraphics[height=5.1cm,width=3.5cm]{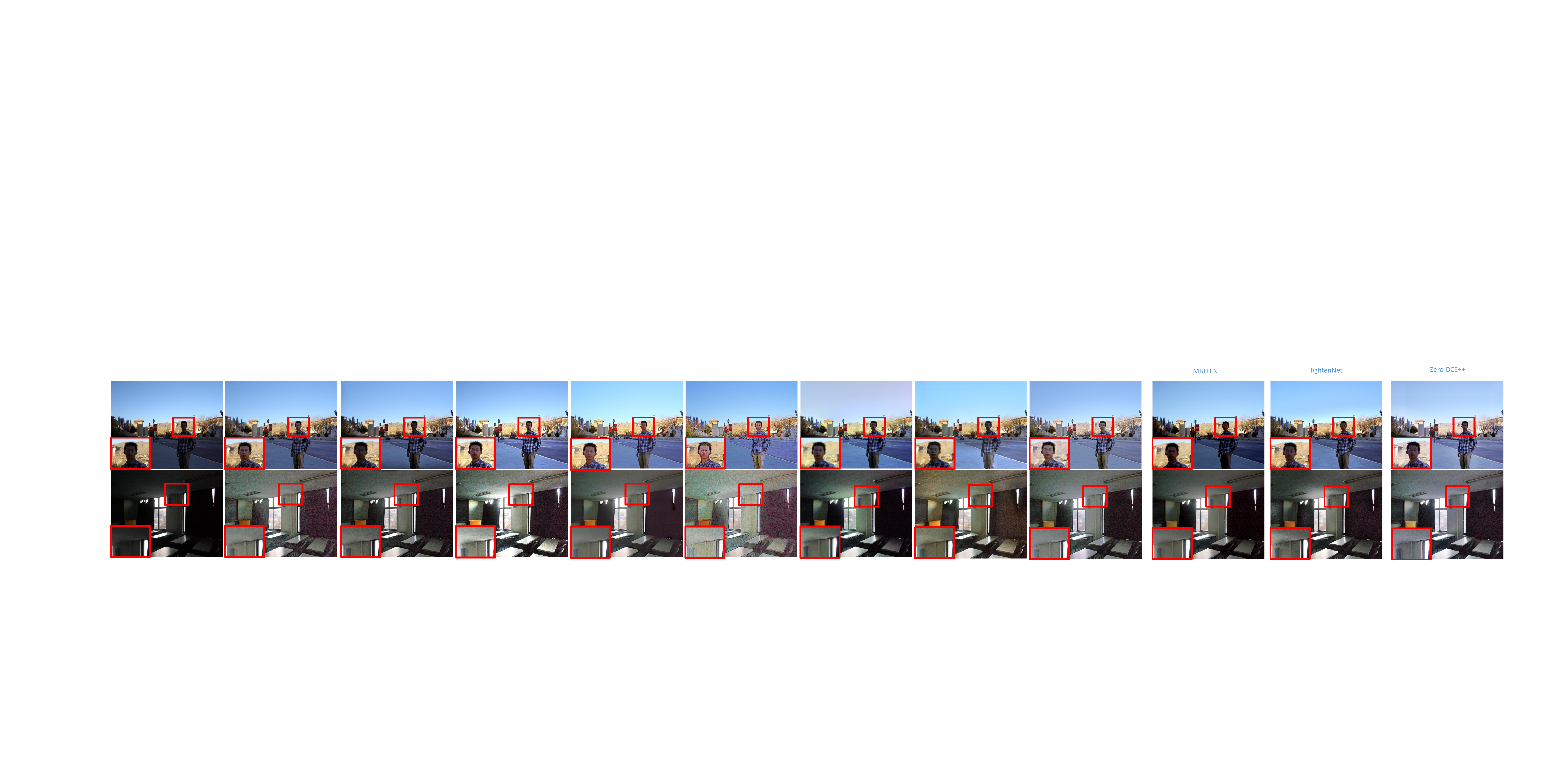}\\
			(f) RetinexNet~\cite{Chen2018} & (g)  Wang \etal~\cite{Wang2019} & (h) EnlightenGAN~\cite{Jiang2019} & (i) Zero-DCE & (j) Zero-DCE++\\
		\end{tabular}
	\end{center}
	\caption{Visual comparisons on typical low-light images.}
	\label{fig:visual_results}
\end{figure*}

\begin{figure*}[!t]
	\begin{center}
		\begin{tabular}{c@{ }c@{ }c@{ }c@{ }c@{ }c@{ }}
			\includegraphics[height=4.5cm,width=3.5cm]{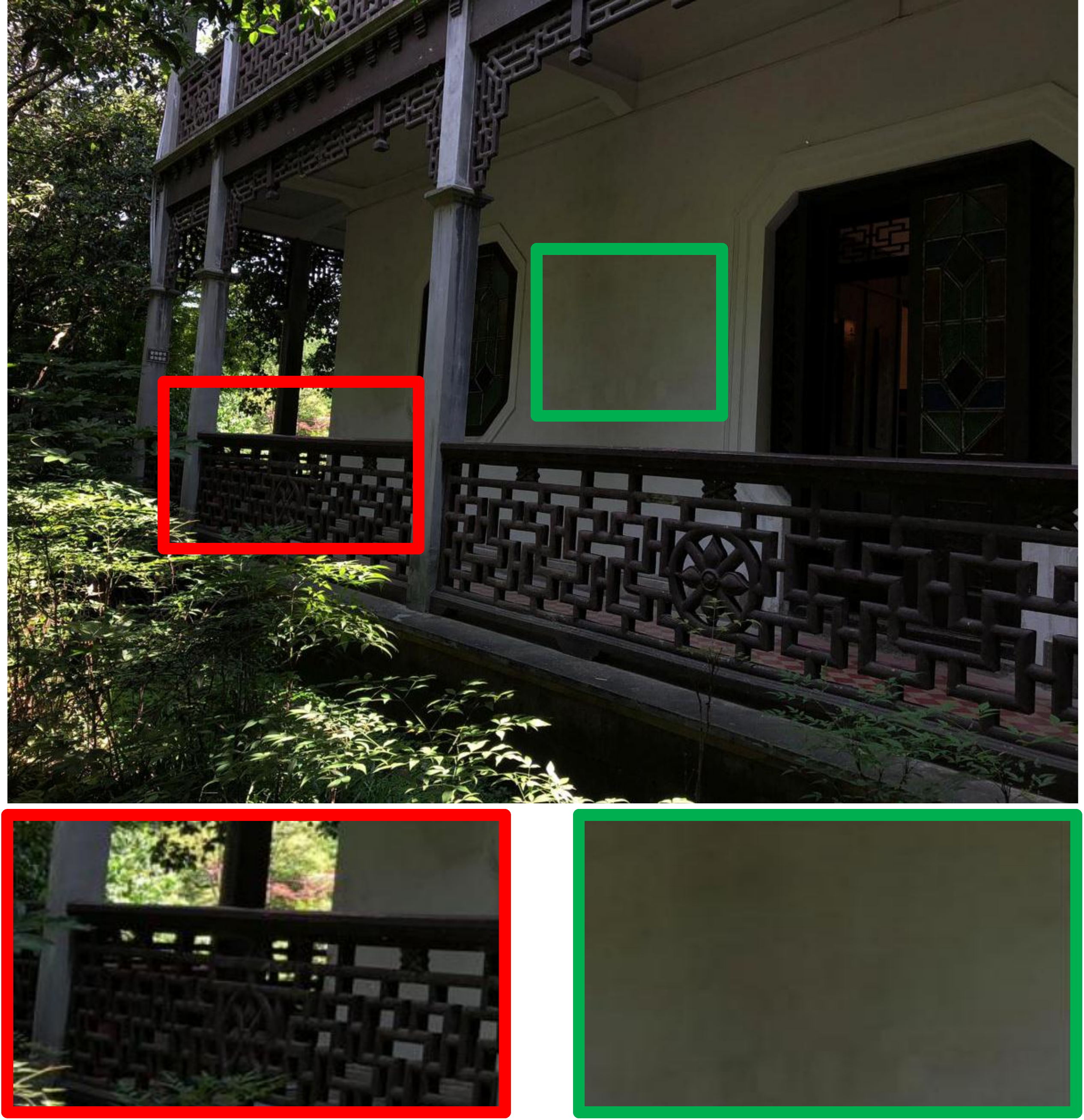}&~~
			\includegraphics[height=4.5cm,width=3.5cm]{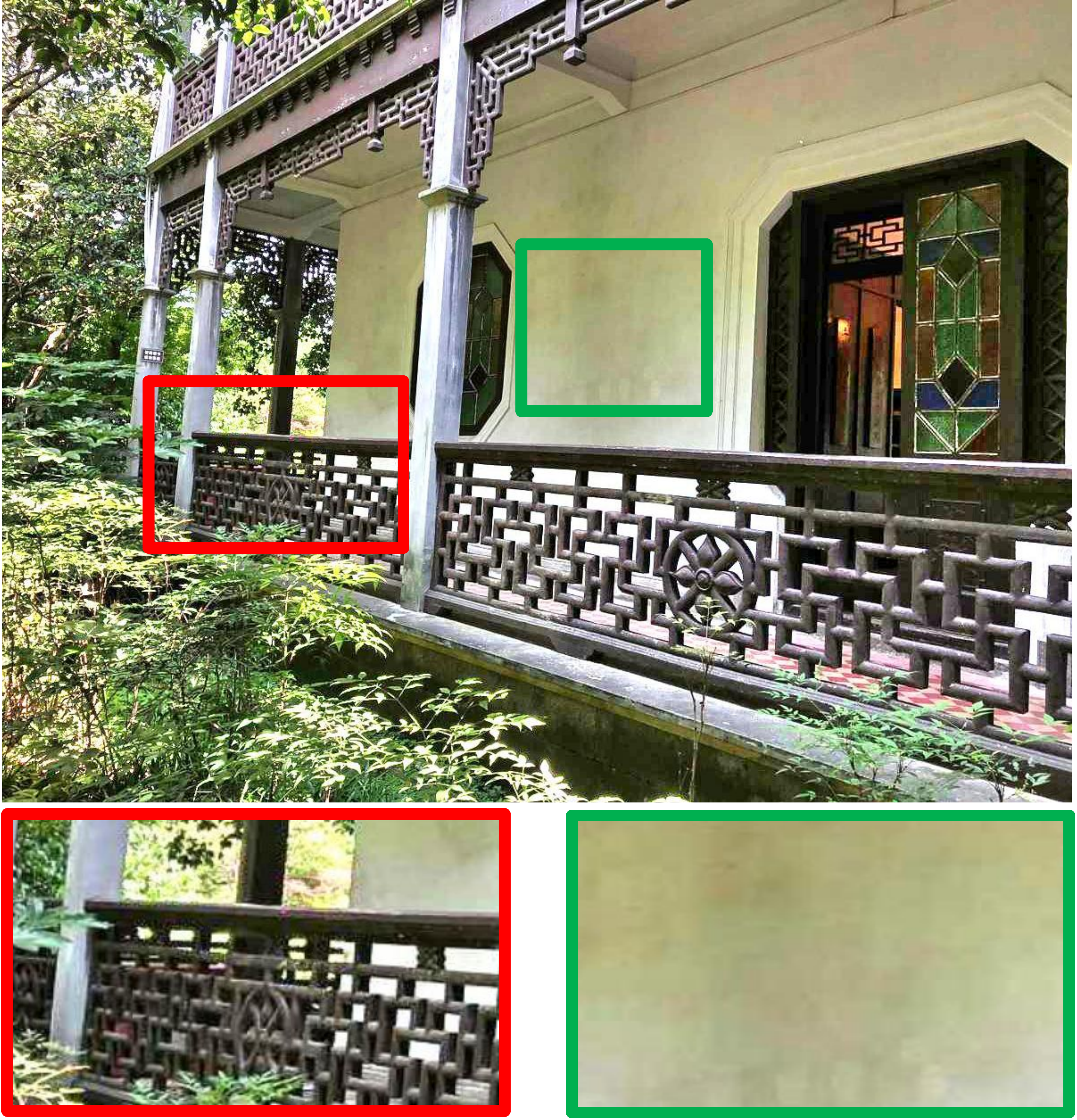}&~~
			\includegraphics[height=4.5cm,width=3.5cm]{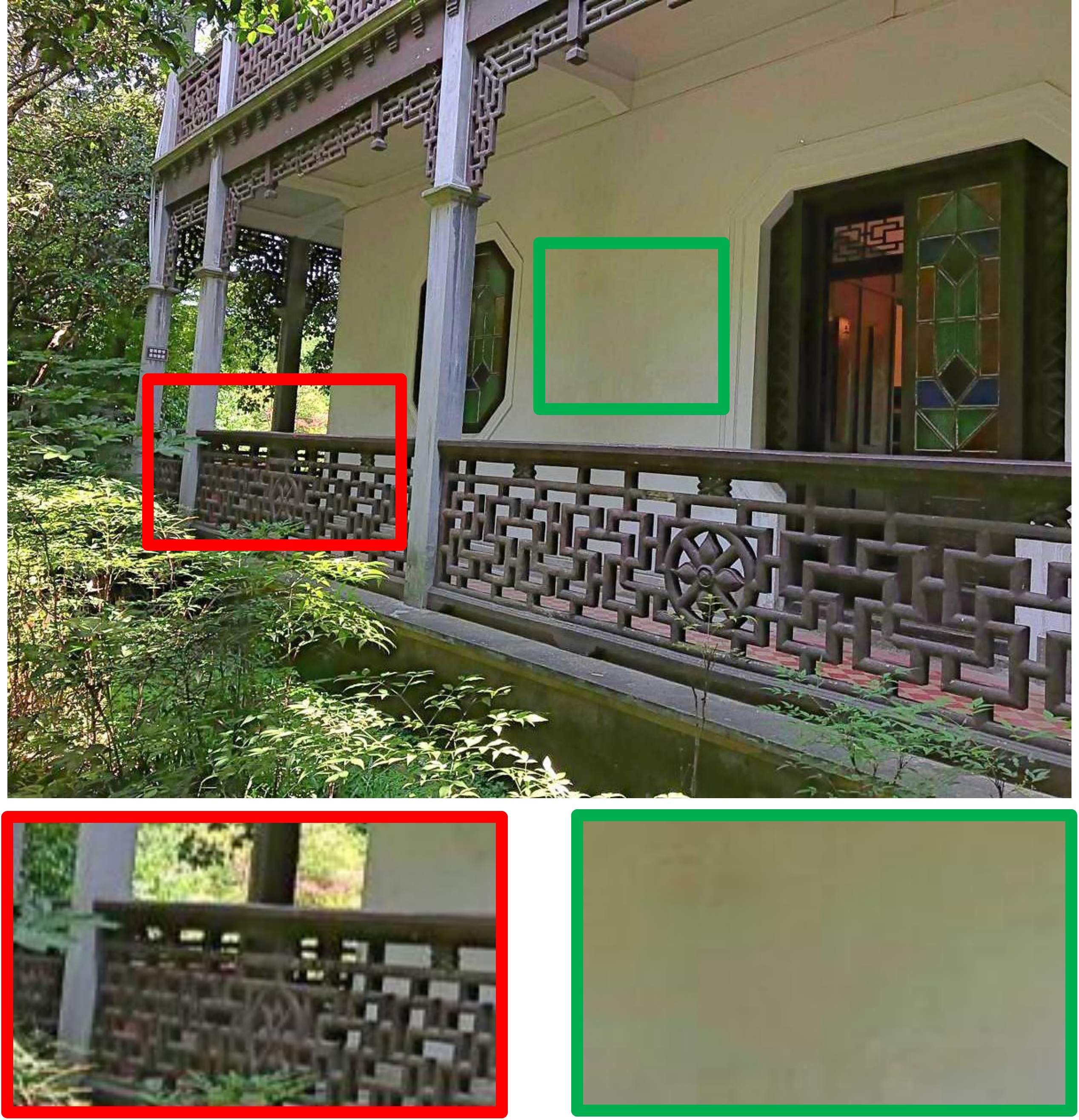}&~~
			\includegraphics[height=4.5cm,width=3.5cm]{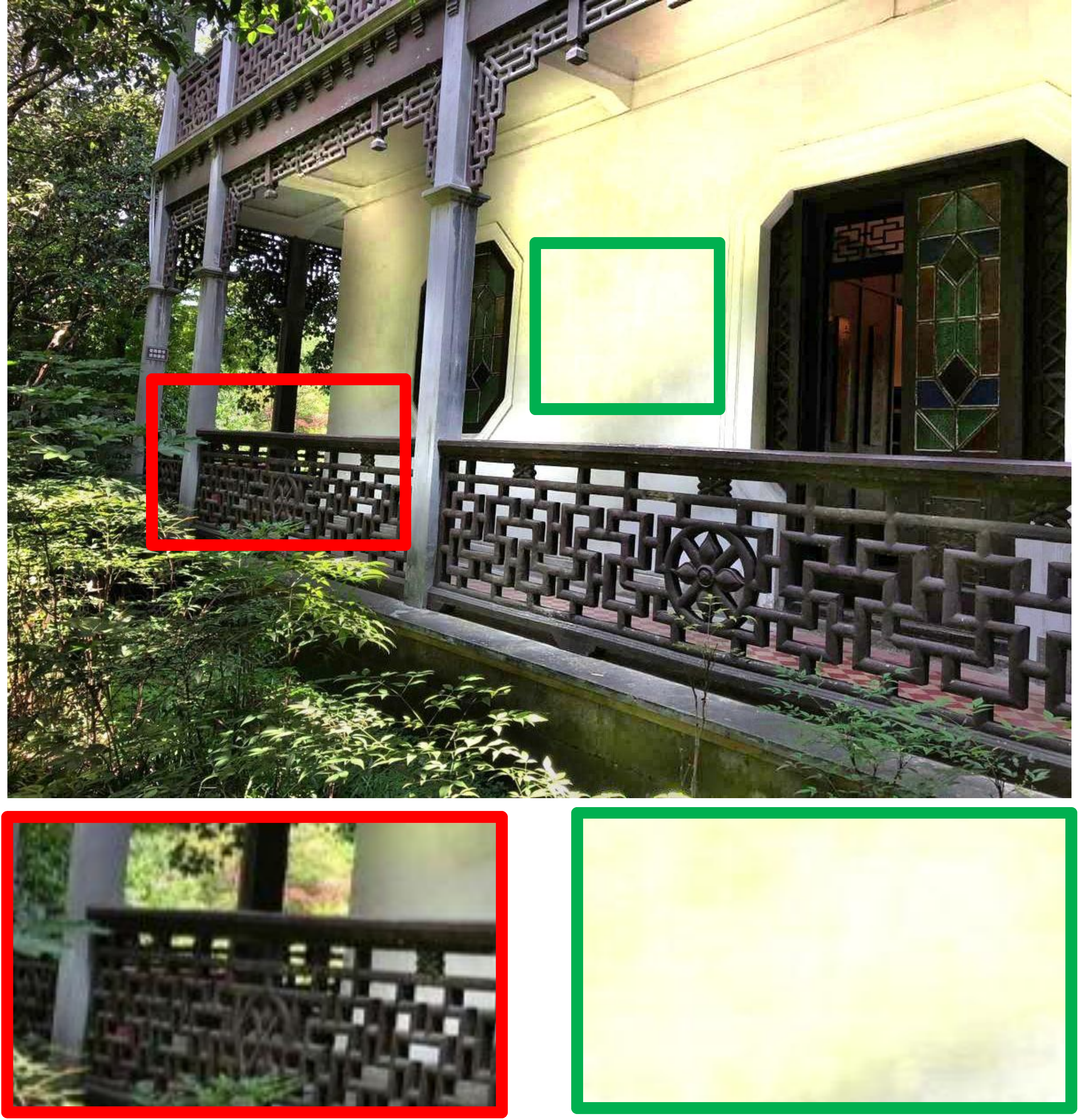}&~~
			\includegraphics[height=4.5cm,width=3.5cm]{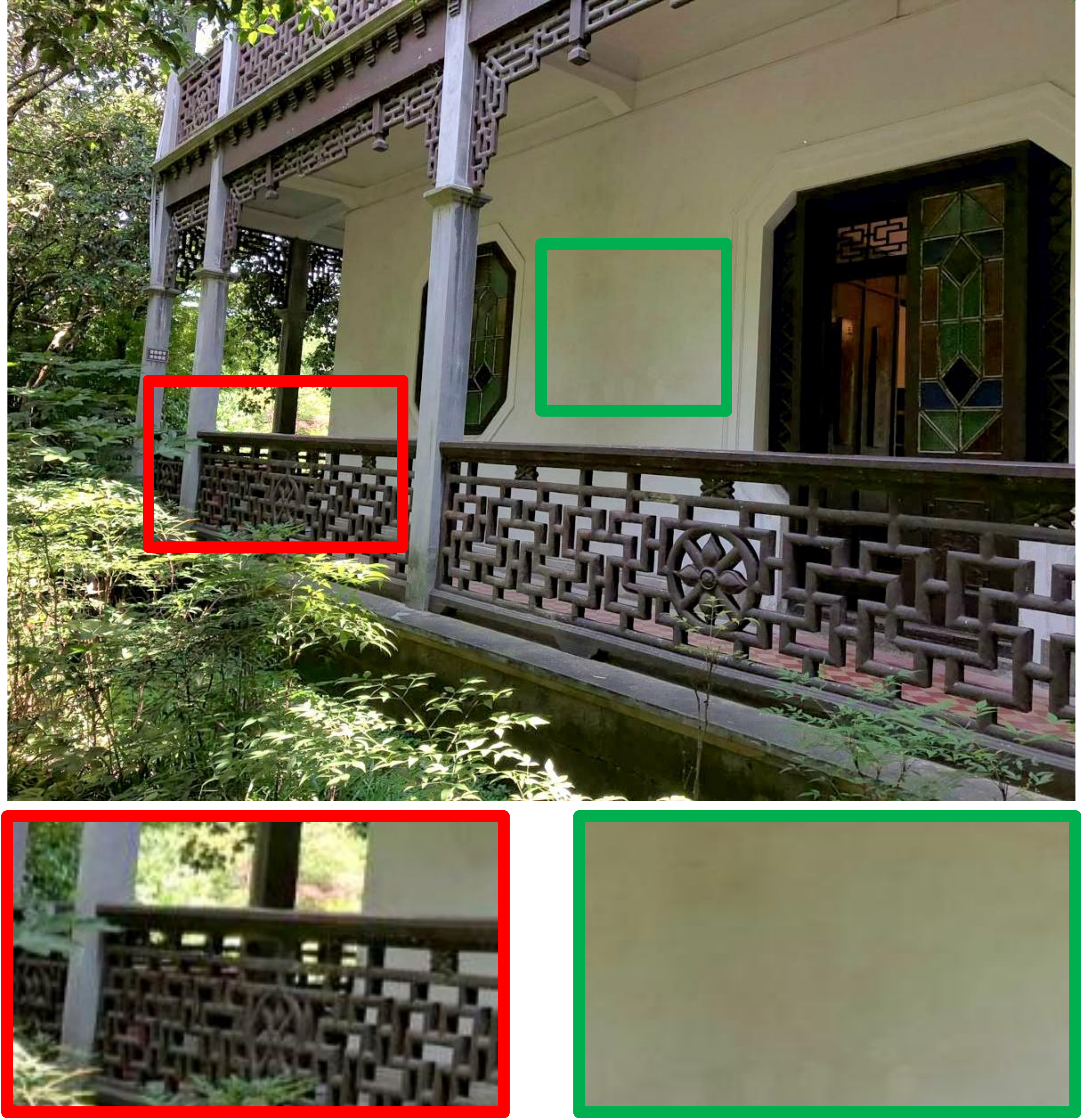}\\
			(a) input &  (b)  LIME~\cite{Guo2017} & (c) Li \etal~\cite{Li2018} & (d) LightenNet~\cite{LightenNet} & (e) MBLLEN~\cite{MBLLEN}\\
			\includegraphics[height=4.5cm,width=3.5cm]{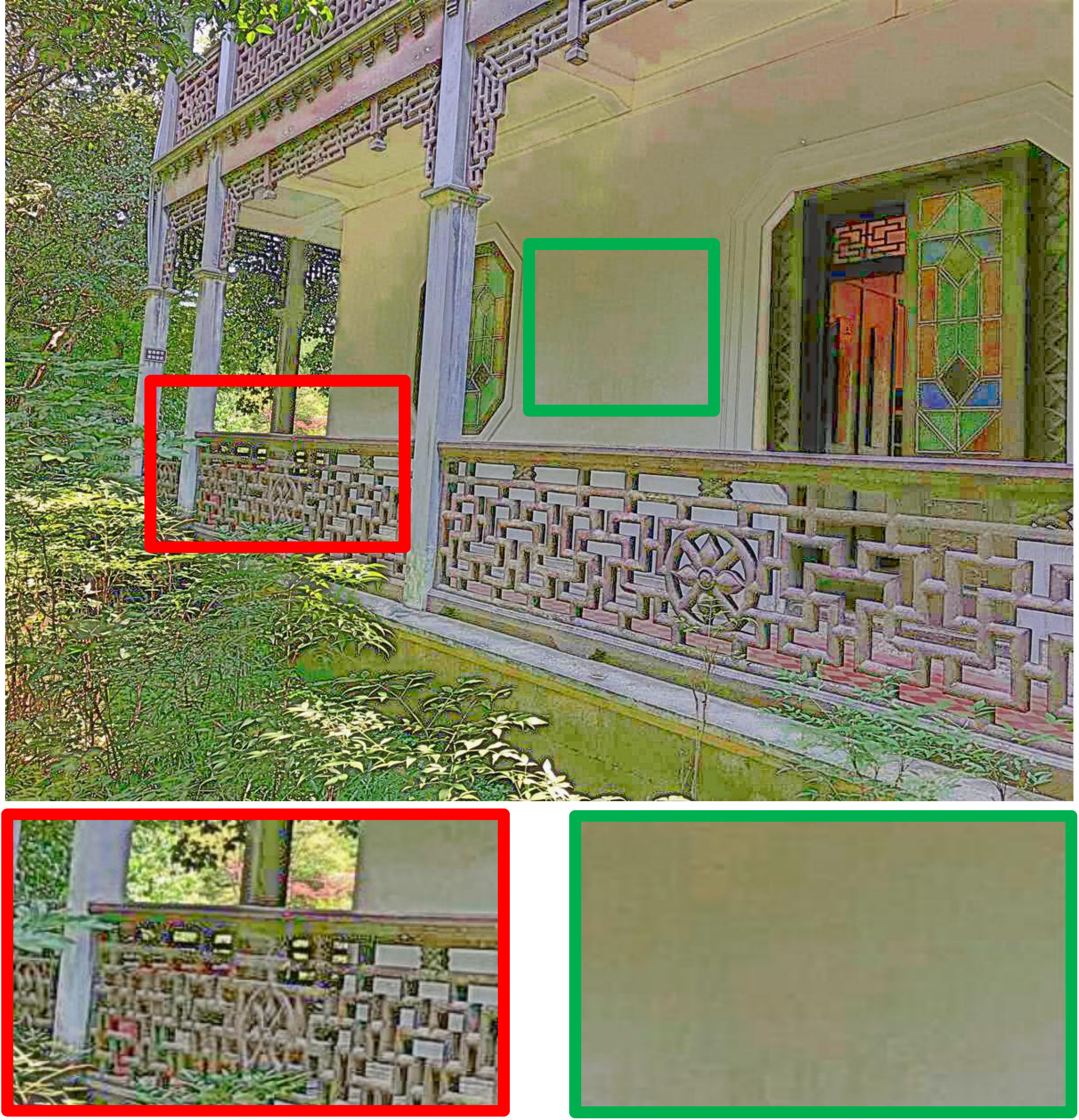}&
			\includegraphics[height=4.5cm,width=3.5cm]{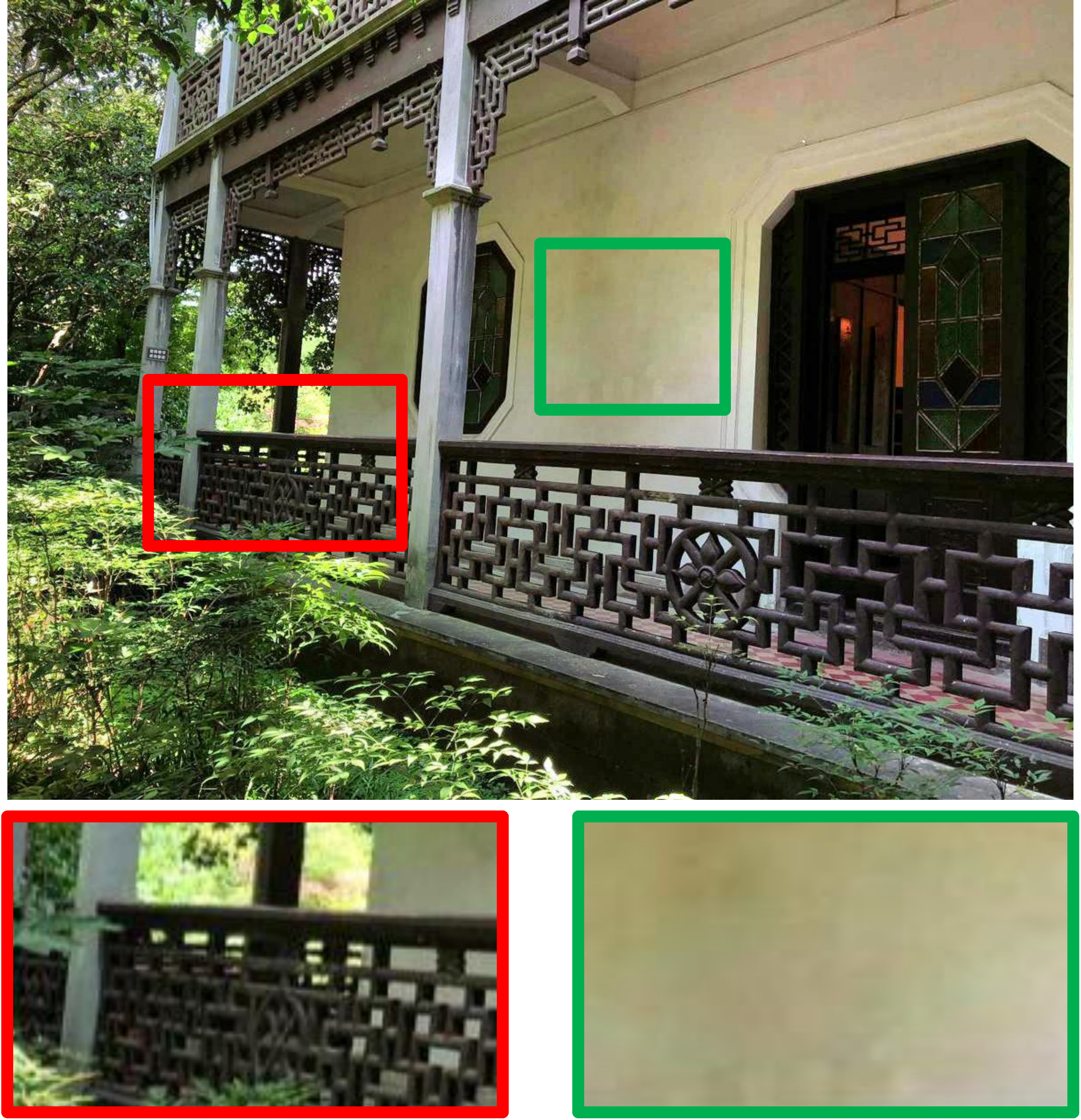}&
			\includegraphics[height=4.5cm,width=3.5cm]{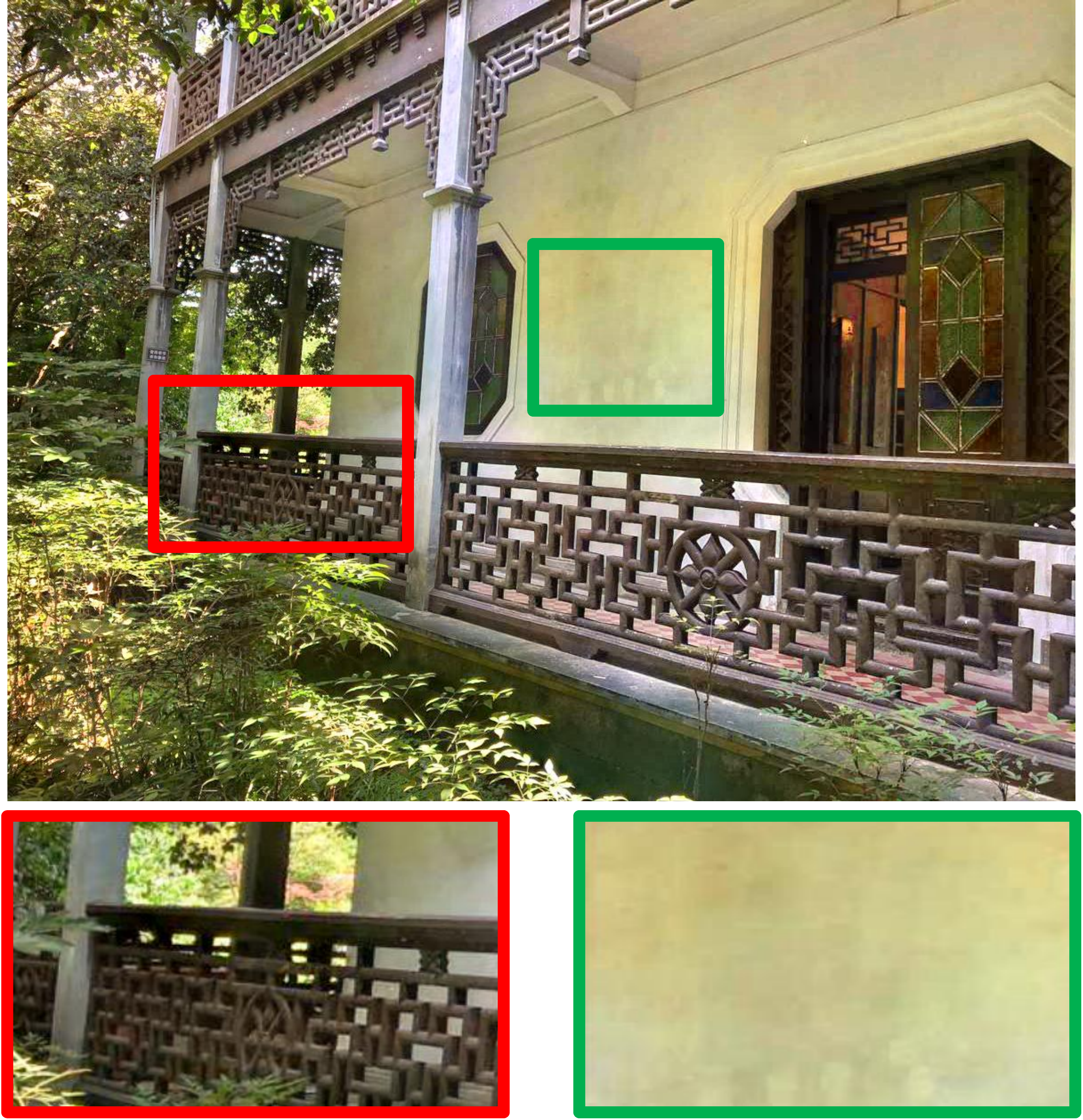}&
			\includegraphics[height=4.5cm,width=3.5cm]{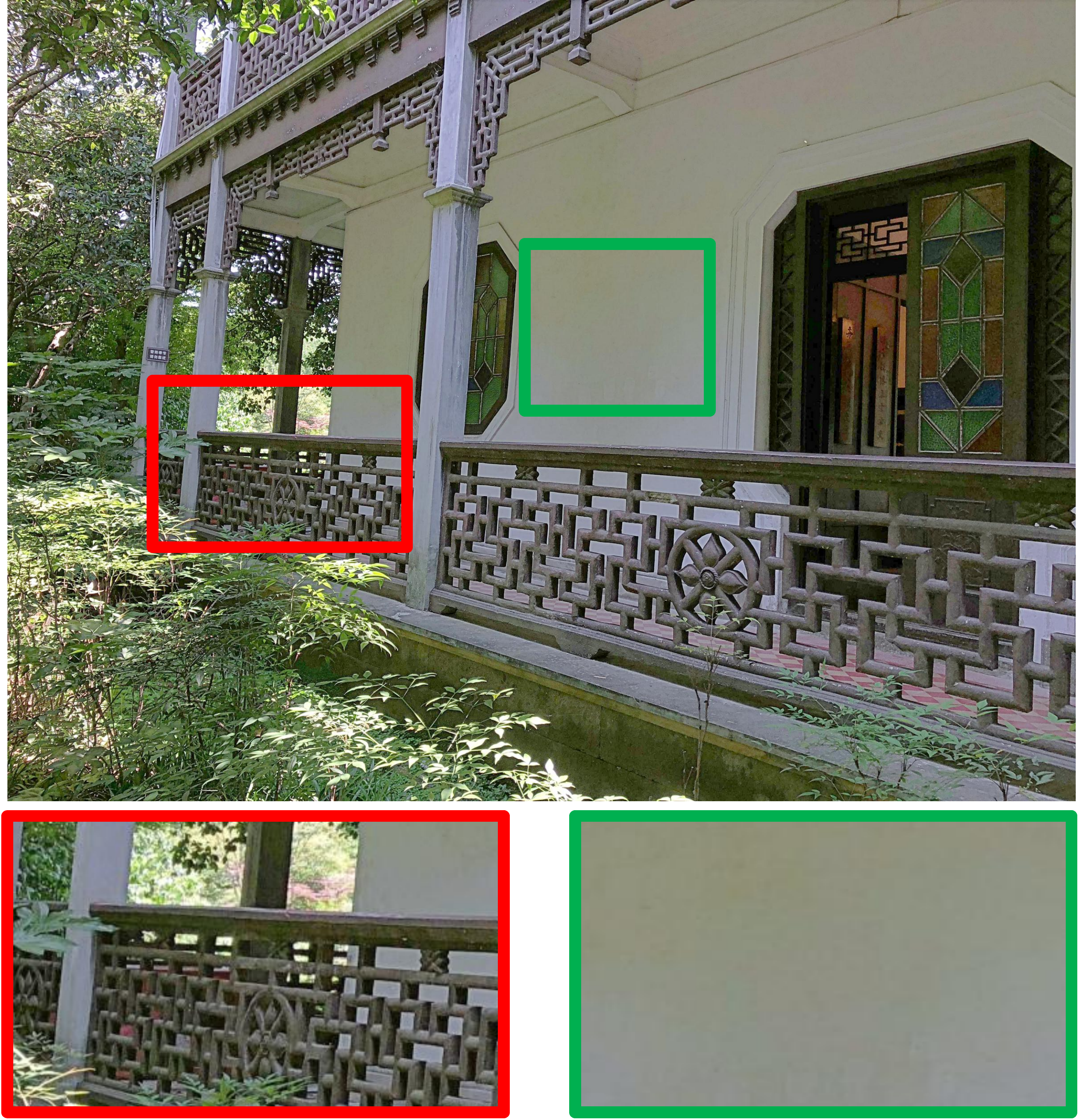}&
			\includegraphics[height=4.5cm,width=3.5cm]{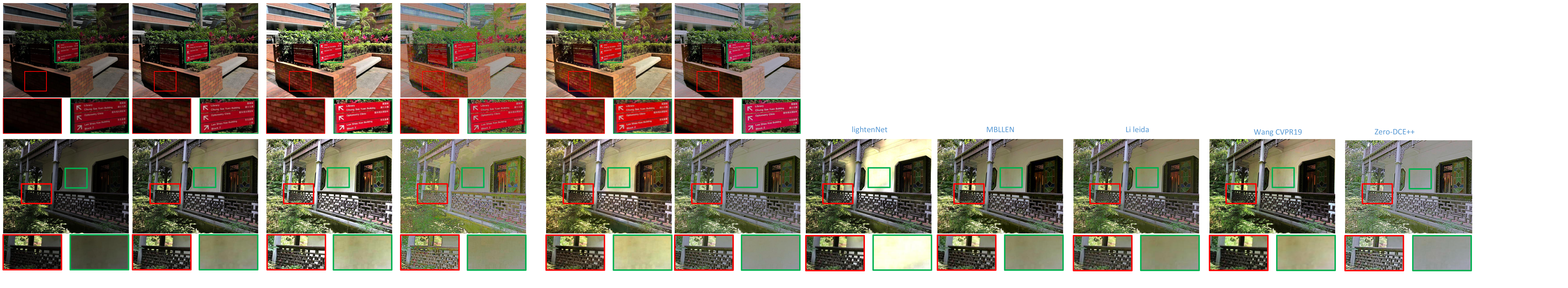}\\
			(f) RetinexNet~\cite{Chen2018} & (g)  Wang \etal~\cite{Wang2019} & (h) EnlightenGAN~\cite{Jiang2019} & (i) Zero-DCE & (j) Zero-DCE++\\
		\end{tabular}
	\end{center}
	\caption{Visual comparisons on a low-light image sampled from the Part2 subset testing set.}
	\label{fig.part2}
\end{figure*}

\subsection{Benchmark Evaluations}
In this section, we conduct qualitative and quantitative experiments to compare different methods. We also investigate the performance of different methods on face detection in the dark.

\subsubsection{Visual and Perceptual Comparisons}
We present the visual comparisons on typical low-light images in Figure~\ref{fig:visual_results}.
For challenging back-lit regions (\eg, the face in Figure~\ref{fig:visual_results}(a)),  Zero-DCE yields natural exposure and clear details while SRIE~\cite{Fu2016}, LIME~\cite{Guo2017}, LightenNet~\cite{LightenNet}, MBLLEN~\cite{MBLLEN}, Wang \etal~\cite{Wang2019}, and EnlightenGAN~\cite{Jiang2019} cannot recover the face clearly. In comparison, RetinexNet~\cite{Chen2018} produces over-exposed artifacts.
In the second example featuring an indoor scene, our method enhances dark regions and preserves color of the input image simultaneously. The result is visually pleasing without obvious noise and color casts.
In contrast,  Li \etal~\cite{Li2018} and MBLLEN \cite{MBLLEN} over-smooth the details while other baseline methods amplify noise and even produce color deviation (\eg, the color of wall). Overall, Zero-DCE++ obtains the comparable performance to Zero-DCE in both indoor and outdoor scenes.

We also show the results of different methods on the image sampled from the Part2 subset testing set. The comparison results are presented in Figure \ref{fig.part2}. Compared with the results of other methods that remain the low-light regions or introduce obvious artifacts, the proposed Zero-DCE and Zero-DCE++ not only produce more clear details but also do not introduce blocking artifacts. 
Our method tends to generate the results with proper contrast, clear details, vivid color, and less noise.

We perform a user study to quantify the subjective visual quality of various methods.
We process low-light images from the image sets (NPE, LIME, MEF, DICM, VV) by different methods.
For each enhanced result, we display it on a screen and provide the input image as a reference.
A total of 15 human subjects are invited to independently score the visual quality of the enhanced image.
These subjects are trained by observing the results from 
\begin{enumerate}
	\item whether the results contain over-/under-exposed artifacts or over-/under-enhanced regions;
	\item whether the results introduce color deviation; and
	\item  whether the results have unnatural texture and obvious noise.
\end{enumerate}

\begin{table*}[t]
	\caption{Runtime (RT, in second), trainable parameters (\#P), and FLOPs (in G) comparisons.  ``-'' indicates that the result is not available. The best result is in red whereas the second best one is in blue under each case.}
	\centering
	\begin{tabular}{c|c|c|c|c}
		\hline
		\textbf{Method} & \textbf{RT} &\textbf{\#P} & \textbf{FLOPs} & \textbf{Platform} \\
		\hline
		SRIE~\cite{Fu2016}           &  12.1865     &     -      &     -       & MATLAB (CPU) \\
		LIME~\cite{Guo2017}          &  0.4914      &     -      &    -        & MATLAB (CPU)\\
		Li \etal~\cite{Li2018}       &  90.7859     &    -   &     -           & MATLAB (CPU) \\
		LightenNet \cite{LightenNet} & 25.7716& {\color{blue}29,532}&30.54 & MATLAB (CPU) \\
		MBLLEN  \cite{MBLLEN}  & 13.9949&450,171 &301.12 &TensorFlow (GPU)\\
		RetinexNet~\cite{Chen2018}   &  0.1200        &  555,205     &   587.47           & TensorFlow (GPU)\\
		Wang \etal~\cite{Wang2019}      &  0.0210    &  998,816  &    {\color{blue}0.19}            & TensorFlow (GPU)\\
		EnlightenGAN~\cite{Jiang2019}&  0.0078      & 8,636,675  & 273.24    & PyTorch (GPU)\\
		Zero-DCE                     &  {\color{blue}0.0025}     & 79,416 &   84.99      & PyTorch (GPU)\\
		Zero-DCE++                     &  {\color{red}0.0012}   &  {\color{red}10,561} &  {\color{red}0.12}        & PyTorch (GPU)\\
		\hline
	\end{tabular}
	\label{table_2}
\end{table*}

The scores of visual quality range from 1 to 5 (worst to best quality). The step size is set to 1.
The average subjective scores for each image set are reported in Table~\ref{label:user study}.
As summarized in Table~\ref{label:user study}, Zero-DCE achieves the highest average User Study (US) score  for a total of 199 testing images from the above-mentioned image sets while  Zero-DCE++ achieves the second-highest  US score. The Zero-DCE and Zero-DCE++ obtain  similar subject scores, which further indicates that the effectiveness and robustness of Zero-DCE++. 
For the MEF, DICM, and VV sets, our results are most favored by the subjects.
All in all, the user study demonstrates that our method can produce a better performance on diverse low-light images from the human subjective visual perspective.

In addition to the US score, we employ a non-reference perceptual index (PI)~\cite{PI,Ma2017,NIQE} to evaluate the perceptual quality. The PI metric is originally used to measure perceptual quality in image super-resolution. It has also been used to assess the performance of other image restoration tasks, such as image dehazing~\cite{Qu2019}. A lower PI value indicates better perceptual quality. The PI values are reported in Table~\ref{label:user study} too.
Similar to the user study, the proposed Zero-DCE is superior to other competing methods in terms of the average PI values.
It obtains the best perceptual quality on LIME, MEF, and DICM sets. Zero-DCE++ also produces competing average PI values.

\subsubsection{Quantitative Comparisons}

We employ the full-reference image quality assessment metrics PSNR, SSIM~\cite{SSIM}, and MAE metrics to quantitatively compare the performance of different methods on the Part2 testing set.  A higher SSIM value indicates a result is closer to the ground truth in terms of structural properties. A higher PSNR (lower MAE) value indicates a result is closer to the ground truth in terms of pixel-level image content.
In Table~\ref{table_1}, the proposed Zero-DCE achieves the best values under all cases, despite that it does not use any paired or unpaired training data. In contrast, Zero-DCE++ obtains comparable performance to Zero-DCE, such as the second-best quantitative scores of PSNR and SSIM values on Part2 testing set. 

Our Zero-DCE is  computationally efficient, benefited from the simple curve mapping form and lightweight network structure. Further,  Zero-DCE++ extremely  speeds up Zero-DCE and only costs  few computational resources. Table~\ref{table_2} shows the runtime\footnote[4]{Runtime is measured on a PC with an Nvidia GTX 2080Ti GPU and Intel I7 6700 CPU, except for Wang \etal~\cite{Wang2019}, which has to run on GTX 1080Ti GPU.}, trainable parameters, and FLOPs  of different methods averaged on 32 images of size 1200$\times$900$\times$3. For conventional methods and 	LightenNet \cite{LightenNet}, only the codes of CPU version are available. 

Compared with current methods, our method achieves the fastest runtime with a large margin (\ie, Zero-DCE: 0.0025s and Zero-DCE++: 0.0012s). Moreover, the runtime of Zero-DCE++ is only 0.0012s, which is really faster than current methods. The runtime of Zero-DCE++ is 17.5 times and 6.5 times faster than recent deep learning-based methods Wang \etal~\cite{Wang2019} and EnlightenGAN~\cite{Jiang2019}, respectively. Zero-DCE++ only contains a tiny network structure that has 10,561 trainable parameters and costs 0.12G FLOPs, which are extremely suitable for practical applications.

\subsubsection{Face Detection in the Dark}
\begin{figure}[!t]
	\centering
	\centerline{\includegraphics[width=1\linewidth]{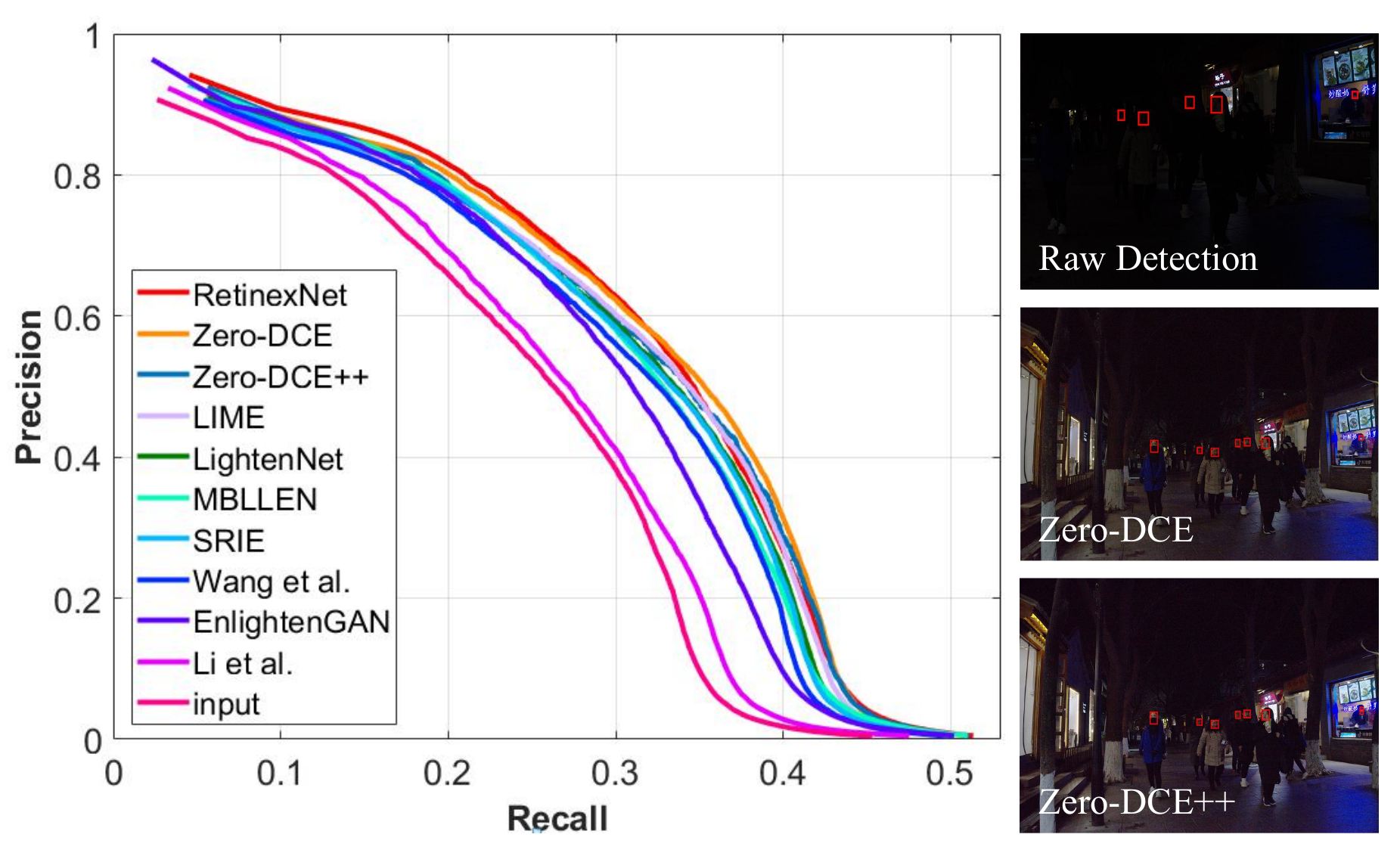}}
	\caption{The P-R curves of face detection in the dark. Best viewed on a color screen in high resolution with zoom in.}
	\label{fig:PR}
\end{figure}

We investigate the performance of low-light image enhancement methods on the face detection task under low-light conditions. Specifically, we use the DARK FACE dataset~\cite{2019arXiv190404474Y} that composes 10,000 images taken in the dark.
Since the bounding boxes of test set are not publicly available, we perform an evaluation on the training and validation sets, which totally consists of 6,000 images.
A state-of-the-art deep face detector, Dual Shot Face Detector (DSFD)~\cite{DSFD}, trained on WIDER FACE dataset~\cite{WIDER}, is used as the baseline model. We feed the results of different low-light image enhancement methods to  DSFD~\cite{DSFD}. We  depict the precision-recall (P-R) curves under IoU threshold 0.5 in Figure~\ref{fig:PR} and compare the average precision (AP) under different IoU thresholds (\ie, 0.5, 0.7, 0.9) using the evaluation tool\footnote[5]{\url{https://github.com/Ir1d/DARKFACE_eval_tools}} provided in DARK FACE dataset~\cite{2019arXiv190404474Y}. 
The AP results are presented in Table~\ref{table:AP}.

\begin{table}[!t]
	\caption{The average precision (AP) for face detection in the dark under different IoU thresholds (0.5, 0.7, 0,9). The best result is in red whereas the second best one is in blue under each case.}
	\centering
	\begin{tabular}{c|c|c|c}
		\hline
		\multirow{2}{*}{\textbf{Method}}&\multicolumn{3}{c}{\textbf{IoU thresholds}} \\
		\cline{2-4}
		\cline{2-4}
		 & \textbf{0.5}  & \textbf{0.7} & \textbf{0.9} \\
		\hline
		input & 0.231278 &   0.007296       &0.000002  \\
		SRIE~\cite{Fu2016}           &  0.288193     &  0.012621        &   {\color{blue}{0.000007}}  \\
		LIME~\cite{Guo2017}          &  0.293970      &  0.013417       &  {\color{blue}{0.000007}}    \\
		Li \etal~\cite{Li2018}       &  0.243714     & 0.008616       & 0.000003    \\
		LightenNet \cite{LightenNet} & 0.290128& 0.012581 &0.000005\\
		MBLLEN  \cite{MBLLEN}  &0.289232 &0.013696 &{\color{blue}{0.000007}} \\
		RetinexNet~\cite{Chen2018}   &  {\color{red}{0.304933}}   &{\color{red}{0.017545}} &0.000005           \\
		Wang \etal~\cite{Wang2019}      & 0.280068    & 0.011107 & 0.000003    \\
		EnlightenGAN~\cite{Jiang2019}&       0.276574   &0.013204  & {\color{red}{0.000009}}    \\
		Zero-DCE                     &  {\color{blue}{0.303135}}  &{\color{blue}{0.014772}}  & 0.000005      \\
		Zero-DCE++                     & 0.297977   & 0.014587 &    0.000005      \\
		\hline
	\end{tabular}
	\label{table:AP}
\end{table}

As shown in Figure~\ref{fig:PR}, after image enhancement, the precision of DSFD~\cite{DSFD} increases considerably compared to that using the input images without enhancement. Among different methods, RetinexNet~\cite{Chen2018}, Zero-DCE, and Zero-DCE++ perform the best. These three methods are comparable but Zero-DCE and Zero-DCE++ perform better in the high recall area. 
As presented in Table \ref{table:AP}, the AP scores of all methods drop when we set higher IoU thresholds. When the IoU threshold is set to 0.9, the performance of all methods is extremely poor. Under the IoU thresholds of 0.5 and 0.7,  Zero-DCE and Zero-DCE++ obtain similar AP scores that are just a little lower than the best result produced by RetinexNet~\cite{Chen2018}.  However, the subjective and quantitative results of RetinexNet~\cite{Chen2018} are unsatisfactory as shown before. In contrast, our method does not require paired training data and balances the subject enhancement performance, application performance, and computational cost well. 
Observing the examples, our Zero-DCE and Zero-DCE++ lighten up the faces in the extremely dark regions and preserves the well-lit regions, thus improves the performance of face detector in the dark.
\section{Conclusion}
\label{sec:Conclusion}
We proposed a deep network for low-light image enhancement. It can be trained end-to-end with zero reference images. This is achieved by formulating the low-light image enhancement task as an image-specific curve estimation problem, and devising a set of differentiable non-reference losses. By re-designing the network structure, reformulating the curve estimation, and controlling the sizes of input image, the proposed Zero-DCE can be further improved, which is significant light-weight and fast for practical applications. Our method excels in both enhancement performance and efficiency. Experiments demonstrate the superiority of our method against existing light enhancement methods.

\ifCLASSOPTIONcompsoc
\section*{Acknowledgments}
\else
\section*{Acknowledgment}
\fi
This research was conducted in collaboration with SenseTime. This work is supported by A*STAR through the Industry Alignment Fund - Industry Collaboration Projects Grant. It is also partially supported by  Singapore MOE AcRF Tier 1 (2018-T1-002-056) and NTU SUG. Chunle Guo is sponsored by CAAI-Huawei MindSpore Open Fund.

\ifCLASSOPTIONcaptionsoff
  \newpage
\fi

{
\bibliographystyle{IEEEtran}
\bibliography{egbib}

\begin{thebibliography}{10}
\providecommand{\url}[1]{#1}
\csname url@samestyle\endcsname
\providecommand{\newblock}{\relax}
\providecommand{\bibinfo}[2]{#2}
\providecommand{\BIBentrySTDinterwordspacing}{\spaceskip=0pt\relax}
\providecommand{\BIBentryALTinterwordstretchfactor}{4}
\providecommand{\BIBentryALTinterwordspacing}{\spaceskip=\fontdimen2\font plus
\BIBentryALTinterwordstretchfactor\fontdimen3\font minus
  \fontdimen4\font\relax}
\providecommand{\BIBforeignlanguage}[2]{{%
\expandafter\ifx\csname l@#1\endcsname\relax
\typeout{** WARNING: IEEEtran.bst: No hyphenation pattern has been}%
\typeout{** loaded for the language `#1'. Using the pattern for}%
\typeout{** the default language instead.}%
\else
\language=\csname l@#1\endcsname
\fi
#2}}
\providecommand{\BIBdecl}{\relax}
\BIBdecl

\bibitem{Pan18}
J.~Pan, D.~Sun, H.~Pfister, and M.~H. Yang, ``Deblurring images via dark
  channel prior,'' \emph{IEEE Transactions on Pattern Analysis and Machine
  Intelligence}, vol.~40, no.~10, pp. 2315--2328, 2018.

\bibitem{Lai18}
W.~S. Lai, J.~B. Huang, N.~Ahuja, and M.~H. Yang, ``Fast and accurate image
  super-resolution with deep laplacian pyramid networks,'' \emph{IEEE
  Transactions on Pattern Analysis and Machine Intelligence}, vol.~41, no.~11,
  pp. 2599--2613, 2019.

\bibitem{Gu19}
S.~Gu, S.~Guo, W.~Zuo, Y.~Chen, R.~Timofte, L.~V. Gool, and L.~Zhang, ``Learned
  dynamic guidance for depth image reconstruction,'' \emph{IEEE Transactions on
  Pattern Analysis and Machine Intelligence}, vol.~42, no.~10, pp. 2437--2452,
  2020.

\bibitem{LiTIP19}
C.~Li, C.~Guo, W.~Ren, R.~Cong, J.~Hou, S.~Kwong, and D.~Tao, ``An underwater
  image enhancement benchmark dataset and beyond,'' \emph{IEEE Transactions on
  Image Processing}, vol.~29, pp. 4376--4389, 2019.

\bibitem{LiTMM19}
C.~Li, C.~Guo, J.~Guo, P.~Han, H.~Fu, and R.~Cong, ``{PDR-Net}:
  Perception-inspired single image dehazing network with refinement,''
  \emph{IEEE Transactions on Multimedia}, vol.~22, no.~3, pp. 704--716, 2019.

\bibitem{Wang2019}
R.~Wang, Q.~Zhang, C.-W. Fu, X.~Shen, W.-S. Zheng, and J.~Jia, ``Underexposed
  photo enhancement using deep illumination estimation,'' in \emph{CVPR}, 2019,
  pp. 6849--6857.

\bibitem{Chen2018}
C.~Wei, W.~Wang, W.~Yang, and J.~Liu, ``Deep retinex decomposition for
  low-light enhancement,'' in \emph{BMVC}, 2018.

\bibitem{Xu2020CVPR}
K.~Xu, X.~Yang, B.~Yin, and R.~W.~H. Lau, ``Learning to restore low-light
  images via decomposition-and-enhancement,'' in \emph{CVPR}, 2020, pp.
  2281--2290.

\bibitem{Jiang2019}
Y.~Jiang, X.~Gong, D.~Liu, Y.~Cheng, C.~Fang, X.~Shen, J.~Yang, P.~Zhou, and
  Z.~Wang, ``Enlighten{GAN}: Deep light enhancement without paired
  supervision,'' 2019, arXiv arXiv:1906.06972.

\bibitem{CycleGAN}
J.-Y. Zhu, T.~Park, P.~Isola, and A.~A. Efros, ``Unpaired image-to-image
  translation using cycle-consistent adversarial networks,'' in \emph{ICCV},
  2017, pp. 2223--2232.

\bibitem{Guo2020CVPR}
C.~Guo, C.~Li, J.~Guo, C.~C. Loy, J.~Hou, S.~Kwong, and R.~Cong,
  ``Zero-reference deep curve estimation for low-light image enhancement,'' in
  \emph{CVPR}, 2020, pp. 1780--1789.

\bibitem{Coltuc2006}
D.~Coltuc, P.~Bolon, and J.-M. Chassery, ``Exact histogram specification,''
  \emph{IEEE Transactions on Image Processing}, vol.~15, no.~5, pp. 1143--1152,
  2006.

\bibitem{Ibrahim2007}
H.~Ibrahim and N.~S.~P. Kong, ``Brightness preserving dynamic histogram
  equalization for image contrast enhancement,'' \emph{IEEE Transactions on
  Consumer Electronics}, vol.~53, no.~4, pp. 1752--1758, 2007.

\bibitem{Stark2000}
J.~A. Stark, ``Adaptive image contrast enhancement using generalizations of
  histogram equalization,'' \emph{IEEE Transactions on Image Processing},
  vol.~9, no.~5, pp. 889--896, 2000.

\bibitem{Lee2013}
C.~Lee, C.~Lee, and C.-S. Kim, ``Contrast enhancement based on layered
  difference representation of 2d histograms,'' \emph{IEEE Transactions on
  Image Processing}, vol.~22, no.~12, pp. 5372--5384, 2013.

\bibitem{Land1977}
E.~H. Land, ``The retinex theory of color vision,'' \emph{Scientific American},
  vol. 237, no.~6, pp. 108--128, 1977.

\bibitem{Wang2013}
S.~Wang, J.~Zheng, H.-M. Hu, and B.~Li, ``Naturalness preserved enhancement
  algorithm for non-uniform illumination images,'' \emph{IEEE Transactions on
  Image Processing}, vol.~22, no.~9, pp. 3538--3548, 2013.

\bibitem{Fu2016}
X.~Fu, D.~Zeng, Y.~Huang, X.-P. Zhang, and X.~Ding, ``A weighted variational
  model for simultaneous reflectance and illumination estimation,'' in
  \emph{CVPR}, 2016, pp. 2782--2790.

\bibitem{Guo2017}
X.~Guo, Y.~Li, and H.~Ling, ``{LIME}: Low-light image enhancement via
  illumination map estimation,'' \emph{IEEE Transactions on Image Processing},
  vol.~26, no.~2, pp. 982--993, 2017.

\bibitem{Li2018}
M.~Li, J.~Liu, W.~Yang, X.~Sun, and Z.~Guo, ``Structure-revealing low-light
  image enhancement via robust retinex model,'' \emph{IEEE Transactions on
  Image Processing}, vol.~27, no.~6, pp. 2828--2841, 2018.

\bibitem{Yuan2012}
L.~Yuan and J.~Sun, ``Automatic exposure correction of consumer photographs,''
  in \emph{ECCV}, 2012, pp. 771--785.

\bibitem{Lore2017}
K.~G. Lore, A.~Akintayo, and S.~Sarkar, ``{LLNet}: A deep autoencoder approach
  to natural low-light image enhancement,'' \emph{Pattern Recognition},
  vol.~61, pp. 650--662, 2017.

\bibitem{MBLLEN}
F.~Lv, F.~Lu, J.~Wu, and C.~Lim, ``{MBLLEN}: Low-light image/video enhancement
  using cnns,'' in \emph{BMVC}, 2018.

\bibitem{Adobe5K}
V.~Bychkovsky, S.~Paris, E.~Chan, and F.~Durand, ``Learning photographic global
  tonal adjustment with a database of input/output image pairs,'' in
  \emph{CVPR}, 2011, pp. 97--104.

\bibitem{Chenchen2018}
C.~Chen, Q.~Chen, J.~Xu, and K.~Vladlen, ``Learning to see in the dark,'' in
  \emph{CVPR}, 2018, pp. 3291--3300.

\bibitem{Chenchen2019}
C.~Chen, Q.~Chen, M.~N. Do, and V.~Koltun, ``Seeing motion in the dark,'' in
  \emph{ICCV}, 2019, pp. 3185--3194.

\bibitem{Ren2019}
W.~Ren, S.~Liu, L.~Ma, Q.~Xu, X.~Xu, X.~Cao, J.~Du, and M.-H. Yang, ``Low-light
  image enhancement via a deep hybrid network,'' \emph{IEEE Transactions on
  Image Processing}, vol.~28, no.~9, pp. 4364--4375, 2019.

\bibitem{Zhang2019ACM}
Y.~Zhang, J.~Zhang, and X.~Guo, ``Kindling the darkness: A practical low-light
  image enhancer,'' in \emph{ACMMM}, 2019, pp. 1632--1640.

\bibitem{Yang2020CVPR}
W.~Yang, S.~Wang, Y.~Fang, Y.~Wang, and J.~Liu, ``From fidelity to perceptual
  quality: A semi-supervised approach for low-light image enhancement,'' in
  \emph{CVPR}, 2020, pp. 3063--3072.

\bibitem{Exposure2007}
T.~Mertens, J.~Kautz, and F.~V. Reeth, ``Exposure fusion,'' in \emph{PCCGA},
  2007.

\bibitem{Exposure2009}
Mertens, J.~Kautz, and F.~V. Reeth, ``Exposure fusion: A simple and practrical
  alterrnative to high dynamic range photography,'' \emph{Computer Graphics
  Forum}, vol.~28, no.~1, pp. 161--171, 2009.

\bibitem{Buchsbaum1980}
G.~Buchsbaum, ``A spatial processor model for object colour perception,''
  \emph{J. Franklin Institute}, vol. 310, no.~1, pp. 1--26, 1980.

\bibitem{depthwiseCon}
F.~Chollet, ``Xception: Deep learning with depthwise separable convolutions,''
  in \emph{CVPR}, 2017, pp. 1251--1258.

\bibitem{ChenECCV18}
L.~C. Chen, Y.~Zhu, G.~Papandreou, F.~Schroff, and H.~Adam, ``Encoder-decoder
  with atrous separable convolution for semantic image segmentation,'' in
  \emph{ECCV}, 2018, pp. 801--818.

\bibitem{HuCVPR18}
H.~Hu, J.~Gu, Z.~Zhang, J.~Dai, and Y.~Wei, ``Relation networks for object
  detection,'' in \emph{CVPR}, 2018, pp. 3588--3597.

\bibitem{LiuCVPR19}
C.~Liu, L.~C. Chen, F.~Schroff, H.~Adam, W.~Hua, A.~Yuille, and L.~F. Fei,
  ``Auto-deeplab:hierarchical neural architecture search for semantic image
  segmentation,'' in \emph{CVPR}, 2019, pp. 82--92.

\bibitem{Chencvpr2018}
Y.~Chen, Y.~Wang, M.~Kao, and Y.~Chuang, ``Deep photo enhancer: Unpaired
  learning for image enhancement from photographs with gans,'' in \emph{CVPR},
  2018, pp. 6306--6314.

\bibitem{Cai2018}
J.~Cai, S.~Gu, and L.~Zhang, ``Learning a deep single image contrast enhancer
  from multi-exposure image,'' \emph{IEEE Transactions on Image Processing},
  vol.~27, no.~4, pp. 2049--2026, 2018.

\bibitem{LightenNet}
C.~Li, J.~Guo, F.~Porikli, and Y.~Pang, ``{LightenNet}: a convolutional neural
  network for weakly illuminated image enhancement,'' \emph{Pattern Recognition
  Letters}, vol. 104, pp. 15--22, 2018.

\bibitem{Ma2015}
K.~Ma, K.~Zeng, and Z.~Wang, ``Perceptual quality assessment for multi-exposure
  image fusion,'' \emph{IEEE Transactions on Image Processing}, vol.~24,
  no.~11, pp. 3345--3356, 2015.

\bibitem{Lee2012}
C.~Lee, C.~Lee, and C.-S. Kim, ``Contrast enhancement based on layered
  difference representation,'' in \emph{ICIP}, 2012, pp. 965--968.

\bibitem{2019arXiv190404474Y}
Y.~Yuan, W.~Yang, W.~Ren, J.~Liu, W.~J. Scheirer, and W.~Zhangyang, ``{UG+
  Track 2}: A collective benchmark effort for evaluating and advancing image
  understanding in poor visibility environments,'' 2019, arXiv
  arXiv:1904.04474.

\bibitem{SSIM}
Z.~Wang, A.~C. Bovik, H.~R. Sheikh, and E.~P. Simoncelli, ``Image quality
  assessment: From error visibility to structural similarity,'' \emph{IEEE
  Transactions on Image Processing}, vol.~13, no.~4, pp. 600--612, 2004.

\bibitem{PI}
Y.~Blau and T.~Michaeli, ``The perception-distortion tradeoff,'' in
  \emph{CVPR}, 2018, pp. 6228--6237.

\bibitem{Ma2017}
C.~Ma, C.-Y. Yang, X.~Yang, and M.-H. Yang, ``Learning a no-reference quality
  metric for single-image super-resolution,'' \emph{Computer Vision and Image
  Understanding}, vol. 158, pp. 1--16, 2017.

\bibitem{NIQE}
A.~Mittal, R.~Soundararajan, and A.~C. Bovik, ``Making a ``completely blind''
  image quality analyzer,'' \emph{IEEE Signal Processing Letters}, vol.~20,
  no.~3, pp. 209--212, 2013.

\bibitem{Qu2019}
Y.~Qu, Y.~Chen, J.~Huang, and Y.~Xie, ``Enhanced pix2pix dehazing network,'' in
  \emph{CVPR}, 2019, pp. 8160--8168.

\bibitem{DSFD}
J.~Li, Y.~Wang, C.~Wang, Y.~Tai, J.~Qian, J.~Yang, C.~Wang, J.~Li, and
  F.~Huang, ``Dsfd: Dual shot face detector,'' in \emph{CVPR}, 2019, pp.
  5060--5069.

\bibitem{WIDER}
S.~Yang, P.~Luo, C.-C. Loy, and X.~Tang, ``{WIDER FACE}: A face detection
  benchmark,'' in \emph{CVPR}, 2016, pp. 5525--5533.

\end{thebibliography}
}
\begin{IEEEbiography}[{\includegraphics[width=1in,height=1.25in,clip,keepaspectratio]{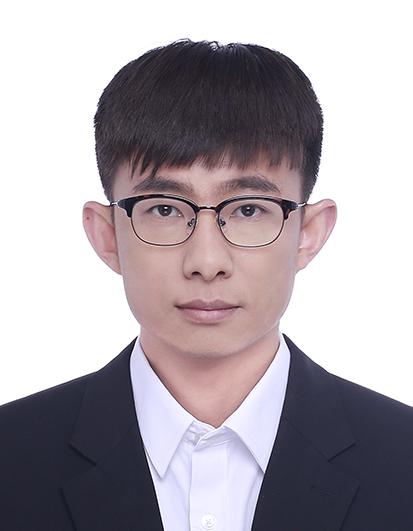}}]{Chongyi Li}  received the Ph.D. degree from the School of Electrical and Information Engineering, Tianjin University, Tianjin, China, in June 2018. From 2016 to 2017, he was a joint-training Ph.D. Student with Australian National University, Australia, under the supervision of Prof. Fatih Porikli. He was a postdoctoral fellow with the Department of Computer Science, City University of Hong Kong, working with Chair Prof. Sam Kwong. He is currently a research fellow with the School of Computer Science and Engineering, Nanyang Technological University (NTU), Singapore. His current research focuses on image processing, computer vision, and deep learning, particularly in the domains of image restoration and enhancement.  
\end{IEEEbiography}

\begin{IEEEbiography}[{\includegraphics[width=1in,height=1.25in,clip,keepaspectratio]{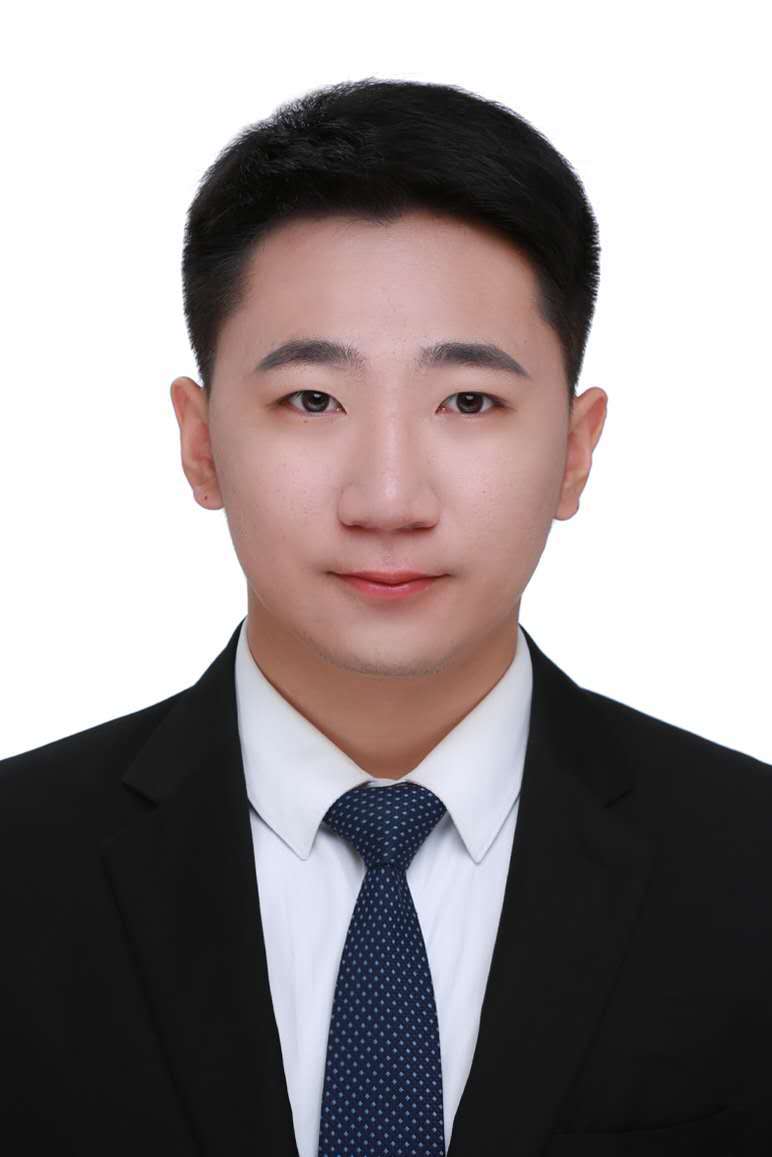}}]{Chunle Guo}  received his PhD degree from Tianjin University in China under the supervision of Prof. Jichang Guo. He conducted the Ph.D. research as a Visiting Student with the School of Electronic Engineering and Computer Science, Queen Mary University of London (QMUL), UK.  He continued his research as a Research Associate with Department of Computer Science, City University of Hong Kong (CityU), from 2018 to 2019. Now he is a postdoc research fellow working with Prof. Ming-Ming Cheng in Nankai University. His research interests lies in image processing, computer vision, and deep learning.
\end{IEEEbiography}

\begin{IEEEbiography}[{\includegraphics[width=1in,height=1.25in,clip,keepaspectratio]{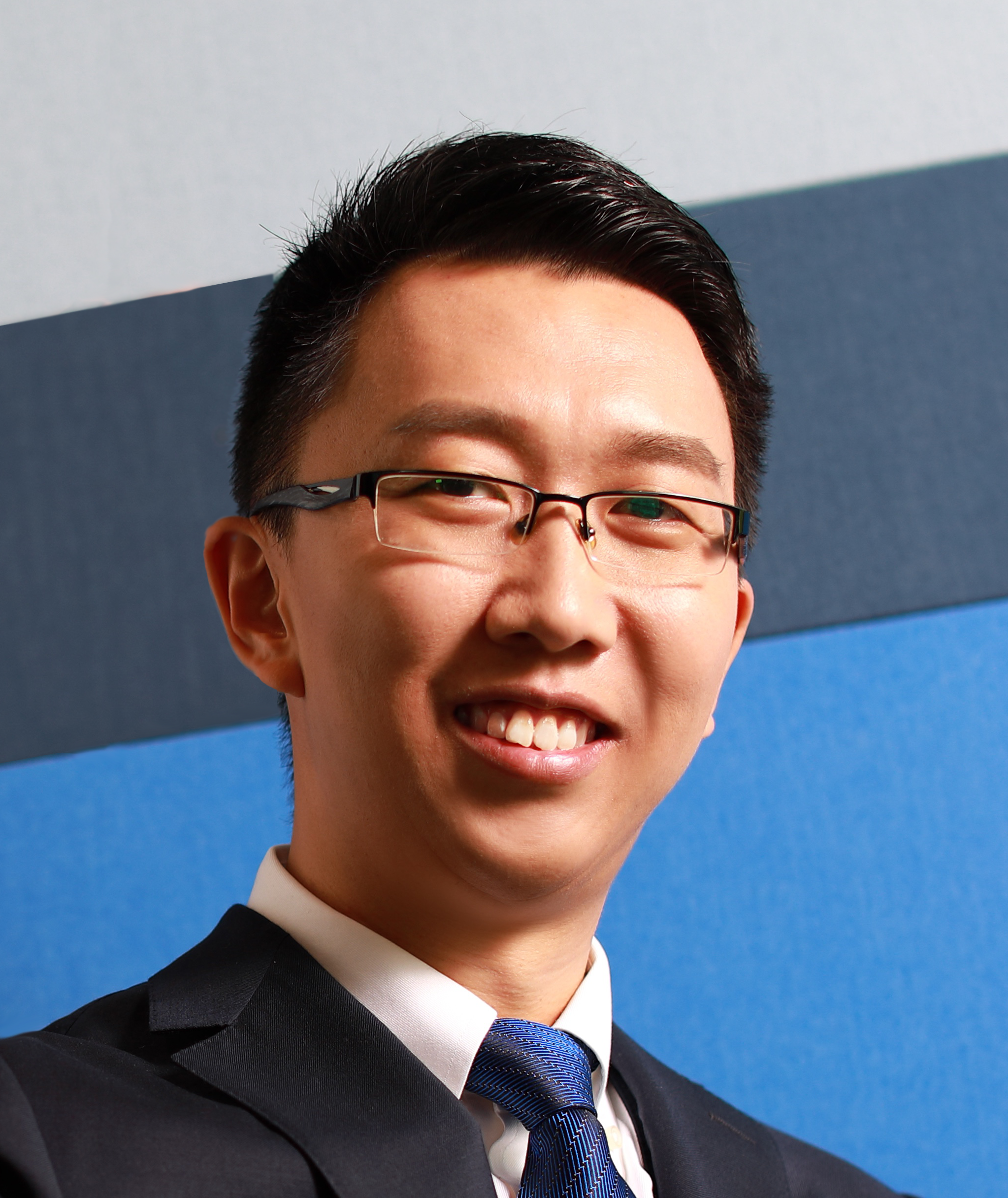}}]{Chen Change Loy} (Senior Member, IEEE) received the PhD degree in computer science from the Queen Mary University of London, in 2010. He is an associate professor with the School of Computer Science and Engineering, Nanyang Technological University. Prior to joining NTU, he served as a research assistant professor with the Department of Information Engineering, The Chinese University of Hong Kong, from 2013 to 2018. His research interests include computer vision and deep learning. He serves as an associate editor of the IEEE Transactions on Pattern Analysis and Machine Intelligence and the International Journal of Computer Vision. He also serves/served as an Area Chair of CVPR 2021, CVPR 2019, ECCV 2018, AAAI 2021 and BMVC 2018-2020. 
\end{IEEEbiography}

\end{document}